\documentclass[twoside]{article}

\usepackage[accepted]{aistats2023}
\usepackage{todonotes}
\usepackage[font=small,labelfont=bf]{caption}
\usepackage{xcolor}

\usepackage{amsmath,amsfonts,bm}

\def\eqref#1{equation~\ref{#1}}

\def\1{\bm{1}}

\DeclareMathAlphabet{\mathsfit}{\encodingdefault}{\sfdefault}{m}{sl}
\SetMathAlphabet{\mathsfit}{bold}{\encodingdefault}{\sfdefault}{bx}{n}

\definecolor{cadmiumgreen}{rgb}{0.0, 0.42, 0.24}
\usepackage{balance}
\usepackage{bm}
\usepackage{threeparttable}
\usepackage{booktabs} 
\usepackage{multirow}
\usepackage{subfigure}
\usepackage{enumitem}
\usepackage{balance}

\usepackage{enumitem}
\usepackage{hyperref}
\usepackage{url}
\usepackage{makecell}
\usepackage[utf8]{inputenc} 
\usepackage[T1]{fontenc}    
\usepackage{hyperref}       
\usepackage{url}            
\usepackage{booktabs}       
\usepackage{amsfonts}       
\usepackage{nicefrac}       
\usepackage{microtype}      
\usepackage{xcolor}         
\usepackage{amsmath}
\usepackage{amssymb}
\usepackage{bbding}
\usepackage{graphicx}
\usepackage{float}
\usepackage{graphicx}

\newtheorem{theorem}{Theorem}

\usepackage[round]{natbib}
\renewcommand{\bibname}{References}
\renewcommand{\bibsection}{\subsubsection*{\bibname}}
\begin{document}

\twocolumn[

\aistatstitle{Spectral Augmentations for Graph Contrastive Learning}

\aistatsauthor{Amur Ghose \And Yingxue Zhang \And Jianye Hao \And Mark Coates}

\aistatsaddress{Huawei \And Huawei \And Huawei, Tianjin University \And McGill} ]

\begin{abstract}
Contrastive learning has emerged as a premier method for learning representations with or without supervision. Recent studies have shown its utility in graph representation learning for pre-training. Despite successes, the understanding of how to design effective graph augmentations that can capture
structural properties common to many different types of downstream graphs remains incomplete. We propose a set of well-motivated graph transformation operations derived via graph spectral analysis to provide a bank of candidates when constructing augmentations for a graph contrastive objective, enabling contrastive learning to capture useful structural representation from pre-training graph datasets.
We first present a spectral graph cropping augmentation that involves filtering nodes by applying thresholds to the eigenvalues of the leading Laplacian eigenvectors. 
Our second novel augmentation reorders the graph frequency components in a structural Laplacian-derived position graph embedding. 
Further, we introduce a method that leads to improved views of local subgraphs by performing alignment via global random walk embeddings. 
Our experimental results indicate consistent improvements in out-of-domain graph data transfer compared to state-of-the-art graph contrastive learning methods, shedding light on how to design a graph learner that is able to learn structural properties common to diverse graph types.

\end{abstract}

%

\section{Introduction}

Representation learning is of perennial importance, with contrastive learning being a recent prominent technique. Taking images as an example, under this framework, a set of transformations is applied to image samples, without changing the represented object or its label. Candidate transformations include cropping, resizing, Gaussian blur, and color distortion. These transformations are termed \textit{augmentations}~\citep{chen2020simclr,grill2020BYOL}. A pair of augmentations from the same sample are termed \textit{positive pairs}. During training, their representations are pulled together~\citep{khosla2020SCL}. In parallel, the representations from \textit{negative pairs},  consisting of augmentations from different samples, are pushed apart. The contrastive objective encourages representations that are invariant to distortions but capture useful features. This constructs general representations, even without labels, that are usable downstream.

Recently, self-supervision has been employed to support the training process for graph neural networks (GNNs). Several approaches (e.g., Deep Graph Infomax (DGI)~\citep{velickovic2019DGI}, InfoGCL~\citep{xu2021infogcl}) rely on mutual information maximization or information bottlenecking between pairs of positive views. Other GNN pre-training strategies construct objectives or views that rely heavily on domain-specific features~\citep{hu2019strategies_gnn,hu2020gpt-gnn}. This inhibits their ability to generalize to other application domains. Some recent graph contrastive learning strategies such as GCC~\citep{qiu2020gcc} and GraphCL~\citep{you2020graphcl} can more readily transfer knowledge to out-of-domain graph domains, because they derive embeddings based solely on local graph structure, avoiding possibly unshared attributes entirely. However, these approaches employ heuristic augmentations such as random walk with restart and edge-drop, which are not designed to preserve graph properties and might lead to unexpected changes in structural semantics~\citep{AFGRL}. 
There is a lack of diverse and effective graph transformation operations to generate augmentations. We aim to fill this gap with a set of well-motivated graph transformation operations derived via graph spectral analysis to provide a bank of candidates when constructing augmentations for a graph contrastive objective. This allows the graph encoder to learn
structural properties that are common for graph data spanning multiple graphs and domains.
\begin{table*}
\centering
\caption{Properties of different approaches to graph contrastive (unsupervised) learning. ${^{\star}}$ indicates that the method was not originally designed for pre-training, but can be trivially adapted to it. \textbf{See Appendix 1 for a more complete description with relevant references.}}
\label{tab:approach-comparison-init}
{\resizebox{\textwidth}{!}{
\begin{tabular}{l|ccccc}
\toprule
Approaches& \makecell{Goal is pre-training \\ or transfer} & \makecell{No  requirement \\ for features} & \makecell{Domain \\ transfer} & \makecell{Shareable graph \\ encoder } \\  
  \midrule
Category 1  (DGI, InfoGraph, MVGRL, DGCL, InfoGCL, AFGRL)  & \XSolidBrush & \XSolidBrush & \XSolidBrush & \XSolidBrush\\

Category 2 (GPT-GNN, Strategies for pre-training GNNs)  & \CheckmarkBold & \XSolidBrush & \XSolidBrush & \CheckmarkBold \\
Category 3  (Deepwalk, LINE, node2vec) & \XSolidBrush & \CheckmarkBold &  \CheckmarkBold & \XSolidBrush \\

Category 4 (struc2vec, graph2vec, DGK, Graphwave, InfiniteWalk)  & \XSolidBrush & \CheckmarkBold &  \CheckmarkBold & \XSolidBrush \\

Category 5 (GraphCL, CuCo${^{\star}}$, GCC, BYOV, GRACE${^{\star}}$, GCA${^{\star}}$, Ours)& \CheckmarkBold & \CheckmarkBold & \CheckmarkBold & \CheckmarkBold \\

\bottomrule
\end{tabular}}}
\label{taxonomy}
\end{table*}

\textbf{Contributions}. We introduce three novel methods: \textit{(i) spectral graph cropping}, \textit{(ii) graph frequency component reordering}, both being graph data augmentations, and a post-processing step termed \textit{(iii) local-global embedding alignment}. We also propose a strategy to select from candidate augmentations, termed \textit{post augmentation filtering}. \textit{First}, we define a graph transformation that removes nodes based on the graph Laplacian eigenvectors. This generalizes the image crop augmentation. \textit{Second}, we introduce an augmentation that reorders graph frequency components in a structural Laplacian-derived position embedding. We motivate this by showing its equivalence to seeking alternative \textit{diffusion} matrices instead of the Laplacian for factorization. This resembles image color channel manipulation. 
\textit{Third}, we introduce the approach of aligning local structural positional embeddings with a global embedding view to better capture structural properties that are common for graph data. Taken together, we improve state-of-the-art methods for contrastive learning on graphs for out-of-domain graph data transfer. We term our overall suite of augmentations \textbf{SGCL (Spectral Graph Contrastive Learning)}.

\section{Related Work} \label{sec:related}

\textbf{Graph contrastive methods.} Table~\ref{taxonomy} divides existing work into five categories. \textit{Category 1} methods rely on mutual information maximization or bottlenecking. \textit{Category 2} methods require that pre-train and downstream task graphs come from the same domain. \textit{Category 3} includes random walk based embedding methods and \textit{Category 4} includes structural similarity-based methods.  These methods do not provide shareable parameters~\citep{you2020graphcl}. \textit{Category 5 (our setting)}: These methods explicitly target pre-training or transfer. Two of the more closely related approaches are Graph Contrastive Coding (GCC)~\citep{qiu2020gcc} and GraphCL~\citep{you2020graphcl}. In GCC, the core augmentation is random walk with return~\citep{tong2006fastrw} and Laplacian positional encoding is used to improve out-of-domain generalization. GraphCL~\citep{you2020graphcl} expands this augmentation suite by including node dropping, edge perturbations, and attribute masking. Other methods in Category 5 construct adaptive/learnable contrastive views~\citep{GCA,Cuco,BYOV_graph,AFGRL}. Please see Appendix 1 for more detailed discussion.

\textbf{Graph structural augmentations.} We focus on the most general, adaptable and transferable \textit{structure-only} scenario --- learning a GNN encoder using a large scale pre-training dataset with solely structural data and no attributes or labels. While not all methods in category $5$ address this setting, they can be adapted to run in such conditions by removing domain or attribute-reliant steps. The graph augmentation strategy plays a key role in the success of graph contrastive learning~\citep{qiu2020gcc, you2020graphcl, li2021dgcl, sun2019infograph,hassani2020MVGRL, xu2021infogcl} and is a natural target as our area of focus. Commonly-used graph augmentations include: 1) attribute dropping or masking~\citep{you2020graphcl,hu2020gpt-gnn}; 2) random edge/node dropping~\citep{li2021dgcl,xu2021infogcl,Grace,GCA}; 3) graph diffusion~\citep{hassani2020MVGRL} and 4) random walks around a center node~\citep{tong2006fastrw,qiu2020gcc}. Additionally, there is an augmentation called GraphCrop~\citep{Graphcrop}, which uses a node-centric strategy to crop a contiguous subgraph from the original graph while maintaining its connectivity; this is different from the spectral graph cropping we propose. Existing structure augmentation strategies are not tailored to any special graph properties and might unexpectedly change the semantics~\citep{AFGRL}.

\textbf{Positioning our work.} Encoding human-interpretable structural patterns such as degree, triangle count, and graph motifs, is key to successful architectures such as GIN~\citep{xu2019GIN} or DiffPool~\citep{ying2018hierarchical} and these patterns control the quality of out-of distribution transfer~\citep{yehudai2021local} for graph tasks, which naturally relates to the pre-train framework where the downstream dataset may differ in distribution from the pre-train corpus. We seek a GNN which learns to capture structural properties common to diverse types of downstream graphs.

These commonly used structural patterns (e.g., degree, triangle count) are handcrafted.
It is preferable to learn these features instead of defining them by fiat. Our goal is to create an unsupervised method that learns functions of the graph structure alone, which can freely transfer downstream to any task. The use of spectral features to learn these structural embeddings is a natural choice; spectral features such as the second eigenvalue or the spectral gap relate strongly to purely structural features such as the number of clusters in a graph, the number of connected components, and the d-regularity~\citep{spielman2007spectral}. Methods based on spectral eigendecomposition such as Laplacian embeddings are ubiquitous, and even random-walk based embeddings such as LINE~\citep{tang2015line} are simply eigendecompositions of transformed adjacency matrices. Instead of handcrafting degree-like features, we strive to construct a learning process that allows the GNN to learn, in an unsupervised fashion, useful structural motifs. By founding the process on the spectrum of the graph, learning can  move freely between the combinatorial, discrete domain of the nodes and the algebraic domain of embeddings.

Such structural features are required for the \textbf{structure-only} case, where we have large, unlabeled, pre-train graphs, and no guarantee that any attributes are shared with the downstream task. This is the most challenging setting in graph pre-training. In such a setting, it is only structural patterns that can be learned from the corpus and potentially transferred and employed in the downstream phase. 

\begin{figure*}
    \centering
    \includegraphics[height=3.7 cm]{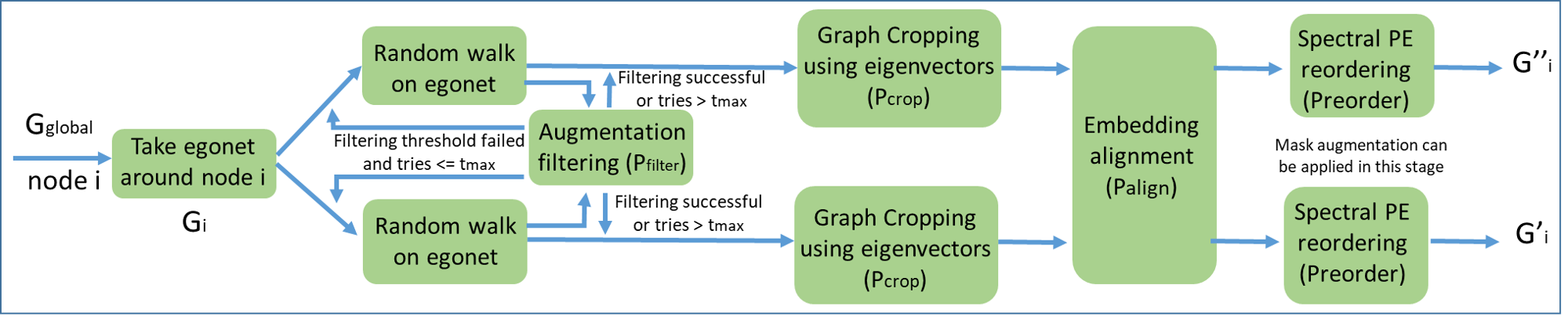}    \caption{Overall framework of SGCL. A box denotes an augmentation that may change the graph with probability denoted or leave it unchanged. Final two augmentations are mutually exclusive. $t_{max}$ denotes the maximum number of tries permitted to the filtering step.}
    \label{fig:overview_figure}
\end{figure*}
\section{Graph Contrastive Learning} \label{sec:subgraphgeneration}

We consider a setting where we have a set of graphs $\mathcal{G} = \{G_t\}$ available for pre-training using contrastive learning. If we are addressing graph-level downstream tasks, then we work directly with the $G_t$. However, if the task is focused on nodes (e.g., node classification), then we associate with each node $i \in G_t$ a subgraph $G_i$, constructed as the $r$-ego subnetwork around $i$ in $G_{t}$, defined as 
\begin{equation}
G_i \triangleq G_{t}[\{ v \in G_{t} : d(i,v) \leq r \}] \label{ego} \,,
\end{equation}
where $d(u,v)$ is the shortest path distance between nodes $i$ and $v$ and $G[S]$ denotes the subgraph induced from $G$ by the subset of vertices $S$. 
During pre-training there are no labels, but in a fine-tuning phase when labels may be available, a subgraph $G_i$ inherits any label associated with node $i$. Thus, node classification is treated as graph classification, finding the label of $G_i.$ This processing step allows us to treat node and graph classification tasks in a common framework.

Our goal is to construct an encoder parametrized by $\theta$, denoted $\mathcal{E}_{\theta}$, such that for a set of instances $G_i \in \mathcal{G}$, the output $\mathcal{E}_{\theta}(G_i)$ captures the essential information about $G_i$ required for downstream tasks. We employ instance discrimination as a contrastive learning objective and minimize~\citep{gutmann2010noise,gutmann2012noise,hjelm2018learning}:
\begin{equation}\label{eqn:contralosseqn}
-\log \frac { \exp \langle \mathcal{E}_{\theta} (G'+),  \mathcal{E}_{\theta'} (G+) \rangle  } { \langle \mathcal{E}_{\theta} (G'+),  \mathcal{E}_{\theta'} (G+) \rangle + \sum_{j=1}^{r} \langle \mathcal{E}_{\theta} (G'+),  \mathcal{E}_{\theta'} (G_j-) \rangle  }\,.
\end{equation} 
Here, $G+,G'+$ may be any augmented version of $G$, and one of them can be $G$ itself. There is an additional sum in the denominator, denoting the number of negative instances.

For the encoder, we construct {\em structure positional embeddings} generalizable to unseen graphs. Let $G_i$ have $N$ nodes, adjacency matrix $\boldsymbol{A_i}$, diagonal degree matrix $\boldsymbol{D_i}$. The normalized Laplacian of $G_i$ is $\boldsymbol{L_i}$, which is eigendecomposed:
\begin{equation}
\boldsymbol{L_i} = \boldsymbol{I} - \boldsymbol{D_i}^{-1/2} \boldsymbol{A_i} \boldsymbol{D_i}^{-1/2}, \quad  \boldsymbol{U_i} \boldsymbol{\Lambda_i} \boldsymbol{U_i}^T = \boldsymbol{L_i}\,.   
\end{equation}
With the $\boldsymbol{\Lambda_i}$ (eigenvalues) sorted in ascending order of magnitude, the first $k$ columns of $\boldsymbol{U_i}$ yield the $k$-dimensional positional embedding, $\boldsymbol{X_i}$, of shape $N \times k$. The pair $(G_i, \boldsymbol{X_i})$ then serves as input to a GNN graph encoder (in our case GIN~\citep{xu2019GIN}), which creates a corresponding hidden vector $\boldsymbol{H_i}$ of shape $N \times h$, where $h$ is the dimensionality of the final GNN layer. Each row corresponds to a vertex $v \in G_i$. A readout function~\citep{gilmer2017mpnn,xu2019GIN}, which can be a simple permutation invariant function such as summation, or a more complex
graph-level pooling function, takes the hidden states over $v \in G_i$ and creates an $h$-dimensional graph representation $\boldsymbol{r}_i$. A view of $G_i$ can be created by conducting an independent random walk (with return) from node $i$, and collecting all the nodes visited in the walk to form $G'_i$. 
The random walk captures the local structure around $i$ in $G_i$ while perturbing it, and is inherently \textit{structural}. A random walk originating from another node $j \in G_{j}, j \neq i$ leads to a negative example $G_j-$.

\section{Spectral Graph Contrastive Augmentation Framework}

In this work, we introduce two novel graph data augmentation strategies: \textit{graph cropping} and \textit{reordering of graph frequency components}. We also propose two important quality-enhancing mechanisms. The first, which we call \textit{augmentation filtering}, selects among candidate augmentations based on their representation similarity. The second, called \textit{local-global embedding view alignment},  aligns the representations of the nodes that are shared between augmentations. We add the masking attribute augmentation~\citep{hu2019strategies_gnn} which randomly replaces embeddings with zeros to form our overall flow of operations for augmentation construction, as depicted in Figure~\ref{fig:overview_figure}. The first two mandatory steps are ego-net formation and random walk. Subsequent steps may occur (with probabilities as $p_{filter},p_{crop},p_{align},p_{mask},p_{reorder}$) or may not. Two of the steps --- mask and reorder --- are mutually exclusive. For more detail, see Appendix $4.4$. In the remainder of the section, we provide a detailed description of the core novel elements in the augmentation construction procedure: (i) spectral {\bf crop}ping; (ii) frequency component {\bf reorder}ing; (iii) {\bf similar} filtering; and (iv) embedding {\bf align}ment. We aim to be as general as possible and graphs are a general class of data - images, for instance, may be represented as grid graphs. Our general graph augmentations such as ``cropping" reduce to successful augmentations in the image domain, lending them credence, as a general method should excel in all sub-classes it contains.

\begin{figure}[ht]
    \centering
    \includegraphics[height=3.7 cm]{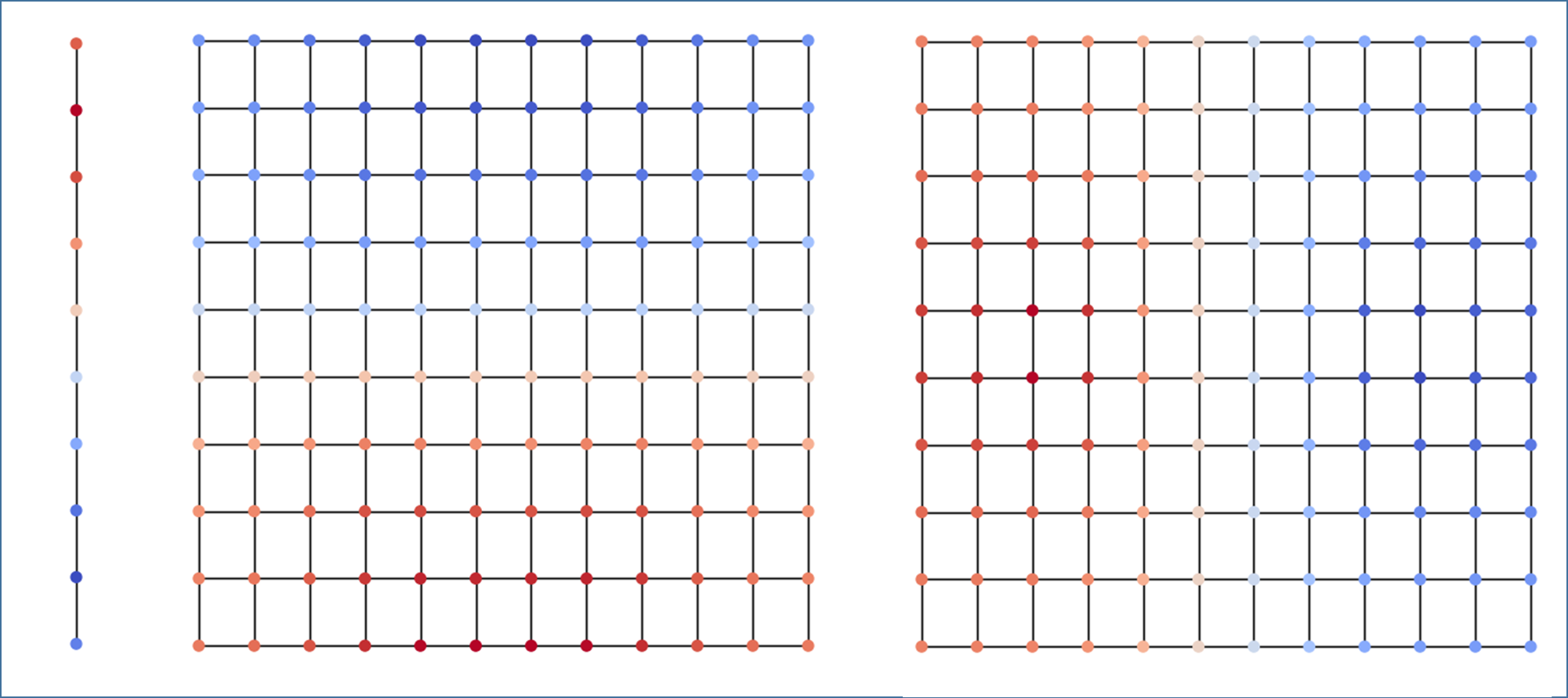}

    \caption{Fiedler eigenvector induced values on a line graph via $\lambda_2$, which change monotonically along the graph, translate into eigenvector values for the grid graph with $\lambda_2,\lambda_3$.}
    \label{fig:fiedler_gridgraph}
\end{figure}

\label{cropdetails}
\subsection{Graph cropping using eigenvectors.} The image cropping augmentation is extremely effective~\citep{chen2020simclr,grill2020BYOL}. It trims pixels along the $(x,y)$ axes. There is no obvious way to extend this operation to general (non-grid) graphs. We now introduce a graph cropping augmentation that removes nodes using the eigenvectors corresponding to the two smallest non-zero eigenvalues of the graph Laplacian $\boldsymbol{L}_i$. When eigenvalues are non-decreasingly sorted, the second eigenvector (corresponding to the lowest nonzero eigenvalue) provides a well-known method to partition the graph --- the {\em Fiedler cut}. We use the eigenvectors corresponding to the first two nonzero eigenvalues, $\lambda_2$ and $\lambda_3$. Let $\boldsymbol{x}(v)$ denote the value assigned to node $v$ in the second eigenvector, and similarly $\boldsymbol{y}(v)$ with the third eigenvector corresponding to $\lambda_3$.
We define the spectral {\bf crop} augmentation as : 
$G_i{[x_{\min},x_{\max},y_{\min},y_{\max}]}$ (a cropped view) being the set of vertices $v \in G_i$ satisfying $x_{\min} \leq \boldsymbol{x}(v) \leq x_{\max}$
and $y_{\min} \leq \boldsymbol{y}(v) \leq y_{\max}$.

\paragraph{Link to image cropping: } We claim that the proposed graph cropping generalizes image cropping. Let us view the values of the eigenvector corresponding to $\lambda_2$ on a {\em line graph} (Figure~\ref{fig:fiedler_gridgraph}). If we set a threshold $t$, and retain only the nodes with eigenvector values below (above) the threshold, we recover a contiguous horizontal segment of the graph. Thus, the eigenvector for $\lambda_2$ corresponds to variation along an axis~\citep{ortega2018gsp,chung1997spectralgraph,davies2000discrete}, much like the $x$ or $y$ axis in an image.

\label{prodgraphdefn} We consider now a {\em product graph}. A product of two graphs $A,B$ with vertex sets $(v_A,v_B)$ and edge sets $(e_A,e_B)$ is a graph $A.B$ where each $v \in A.B$ can be identified with an ordered pair $(i,j), i \in v_A, j \in v_B$. Two nodes corresponding to $(i,j),(i',j')$ in $A.B$ have an edge between them if and only if either $i'=i, (j,j') \in v_B$ or $(i,i') \in v_A, j=j'$. The product of two line graphs of length $M,N$ respectively is representable as a planar rectangular grid of lengths $M,N$.

Denote by $P_n$ the path-graph on $n$ vertices, which has $n{-}1$ edges of form $(i,i+1)$ for $i=1,\dots,n{-}1$. This corresponds to the line graph. Denote by $G_{a,b}$ the rectangular grid graph formed by the product $P_a.P_b$. Structurally, this graph represents an image with dimensions $a \times b$. The eigenvectors of the (un-normalized) Laplacian of $P_n$, for $n > k \geq 0$, are  of the form: 
$\boldsymbol{x_k}(u) = \cos (\pi k u / n - \pi k / 2n)$, with eigenvalues $2 - 2\cos(\pi k/n)$. Clearly, $k=0$ yields the constant eigenvector. The first nonzero eigenvalue corresponds to $k=1$, where the eigenvector completes one ``period" (with respect to the cosine's argument) over the path, and it is this pattern that is shown in Figure~\ref{fig:fiedler_gridgraph}.

The following properties are well-known for the spectrum of product graphs~\citep{brouwer2011spectra}. Each eigenvalue is of the form $\lambda_i + \lambda_j$, where $\lambda_i$ is from the spectrum of $P_a$ and $\lambda_j$ from $P_b$. Further, the corresponding eigenvector $\boldsymbol{v_{i,j}}$ satisfies $\boldsymbol{v_{i,j}}(u,v) = \boldsymbol{x_i}(u)\boldsymbol{y_j}(v)$, where $\boldsymbol{x_i},\boldsymbol{y_j}$ denote the eigenvectors from the respective path graphs.
This means, for the spectra of $G_{a,b}$, that the lowest eigenvalue of the Laplacian corresponds to the constant eigenvector, and the second lowest eigenvalue corresponds to the constant eigenvector along one axis (path) and $\cos(\pi u / n - \pi/2n)$ along another. The variation is along the larger axis, i.e., along $a$, because the $2 - 2\cos(\pi k/n)$ term is smaller. This implies that for $G_{a,b}, 2b > a > b$, $\lambda_2, \lambda_3$ correspond to eigenvectors that recover axes in the grid graph (Figure~\ref{fig:fiedler_gridgraph}).

\subsection{Frequency-based positional embedding reordering} 

Images have multi-channel data, derived from the RGB encoding.  The channels correspond to different frequencies of the visible spectrum. The successful color
reordering augmentation for images~\citep{chen2020simclr} thus corresponds to a permutation
of frequency components. This motivates us to introduce a novel
augmentation that is derived by reordering the graph frequency
components in a structural position embedding. A structural
position embedding can be obtained by factorization of the graph
Laplacian. The Laplacian eigendecomposition corresponds to a
frequency-based decomposition of signals defined on the
graph~\citep{von2007spe,chung1997spectralgraph}. We thus consider
augmentations that permute, i.e., {\em reorder}, the columns of the
structural positional embedding $\boldsymbol{X_i}$.

However, \textit{arbitrary} permutations do not lead to good augmentations. In deriving a position embedding, the normalized Laplacian $\boldsymbol{L_i}$ is not the only valid choice of matrix to factorize. \citet{qiu2018netfm} show that popular random walk
embedding methods arise from the
eigendecompositions of:
\begin{equation} \log (\sum_{j=1}^{r}
  (\boldsymbol{I}-\boldsymbol{L_i})^r) \boldsymbol{D_i}^{-1}\, = \log\Big(\boldsymbol{U_i} \big(\sum_{j=1}^r
  (\boldsymbol{I}-\boldsymbol{\Lambda_i})^r\big) \boldsymbol{U_i^T}\Big)\boldsymbol{D_i}^{-1}.
\label{eq:hodec}
\end{equation}
We have excluded negative sampling and graph volume
terms for clarity. We observe that $\sum_{j=1}^{r} (\boldsymbol{I}-\boldsymbol{L_i})^r$
replaces $(\boldsymbol{I}-\boldsymbol{L_i})$ in the
spectral decomposition. Just as the adjacency matrix
$\boldsymbol{A_i}$ encodes the first order proximity (edges),
$\boldsymbol{A_i^2}$ encodes second order connectivity,
$\boldsymbol{A_i^3}$ third order and so on. Using larger values of $r$ in~\eqref{eq:hodec} thus
integrates higher order information in the embedding. The sought-after eigenvectors in
$\boldsymbol{X}$ are the columns in $\boldsymbol{U}$ corresponding to
the top $k$ values of $\sum_{j=1}^{r} (1-\lambda)^j$. There is no
need to repeat the eigendecomposition to obtain a new embedding. The
higher-order embedding is obtained by {\em reordering} the
eigenvectors (in descending order of
$\sum_{j=1}^{r} (1-\lambda_w)^j$).

This motivates our proposed reordering
augmentation and identifies suitable permutation matrices. Rather than permute all of the eigenvectors in the eigendecomposition, for
computational efficiency, we first extract the $k$
eigenvectors with the highest corresponding eigenvalues in the first order positional embedding derived using $(\boldsymbol{I}-\boldsymbol{L_i})$. The reordering augmentation only permutes those $k$ eigenvectors. The augmentation thus forms $\boldsymbol{X_iP_r}$ where $\boldsymbol{P_r}$ is a permutation matrix of shape $k \times k$. The
permutation matrix $\boldsymbol{P_r}$ sorts
eigenvectors with respect to the values
$\sum_{j=1}^{r} (1-\lambda_w)^j$. We randomize the permutation matrix generation step by
sampling an integer uniformly in the range $[1,r_{max}]$ to serve as $r$ and apply the permutation to produce the view $G'_i$. 

\subsection{Embedding alignment} 

In this subsection and the next, we present two quality enhancing mechanisms that are incorporated in our spectral augmentation
generation process and 
lead to superior  augmentations. Both use auxiliary global structure information. 

Consider two vertices $v$ and $v'$ in the same graph $G_t$. Methods such as Node2vec~\citep{grover2016node2vec}, LINE~\citep{tang2015line}, \&
DeepWalk~\citep{perozzi2014deepwalk} operate on $G_t$ outputting an embedding matrix $\mathbf{E}_{t}$. The row corresponding to vertex $v$ provides a node embedding $\boldsymbol{e_v}$.

Node embedding alignment allows comparing
embeddings between disconnected graphs
$G_1,G_2$ utilizing the structural connections in each
graph~\citep{singh2007pairwise,chen2020conealign,heimann2018regal,grave2019unsupervised}.
 Consider two views $G'_i,G''_i$ and a node $v_i$ such that $v_i \in G'_i, v_i \in G''_i$. Given the embeddings $\boldsymbol{X'_i,X''_i}$ for
$G'_i,G''_i$, ignoring permutation terms, alignment seeks to find an
orthogonal matrix $\boldsymbol{Q}$ satisfying
$\boldsymbol{X''_i Q} \approx \boldsymbol{X'_i}$. If the embedding is computed via eigendecomposition of $\boldsymbol{L'_i, L''_i}$, the final structural
node embeddings (rows corresponding to $v_i$ in $\boldsymbol{X'_i,X''_i}$) for $v_i$ may differ. To correct this, we align the
structural features $\boldsymbol{X'_i,X''_i}$, using the global
matrix $\mathbf{E}_{t}$ as a bridge.

Specifically, let $\boldsymbol{N_{G'_i}}$ be the sub-matrix of $\mathbf{E}_{t}$ obtained by collecting all rows $j$ such that
$v_j \in G_i$. Define $\boldsymbol{N_{G''_i}}$ similarly. We find an orthogonal matrix $\boldsymbol{Q^*} = \min_{\boldsymbol{Q}} || \boldsymbol{X'_i Q - N_{G'_i}} ||^2$. The solution is $\boldsymbol{AC^T}$, where
$\boldsymbol{ABC^T}$ is the singular value decomposition (SVD) of $\boldsymbol{(X'_i)^T N_{G'_i}}$~\citep{heimann2018regal,chen2020conealign}. Similarly, we compute $\boldsymbol{Q^{**}}$ for $G''_i$. We consider the resulting
matrices $\boldsymbol{X'_i Q^*} \approx \boldsymbol{N_{G'_i}} $ and
$ \boldsymbol{X''_i Q^{**}} \approx \boldsymbol{N_{G''_i}}$.
Since $\boldsymbol{N_{G'_i}, N_{G''_i}}$ are both derived from $\mathbf{E}_{t}$, the rows (embeddings) corresponding to a common node are the same.  We can thus derive improved augmentations by reducing the undesirable disparity induced by misalignment and replacing $\boldsymbol{X'_i, X''_i}$ with their aligned counterparts $\boldsymbol{X'_i Q^*}, \boldsymbol{X''_i Q^{**}}$, terming this as \textbf{align}.

\subsection{Augmentation filter.} Consider two views $G'_i, G''_i$ resulting from random walks from a node $a_i$ of which $G_i$ is the ego-network in $G_{t}$. Let $\boldsymbol{E_{G'_i}}= \sum_{v_z \in G'_i} \mathbf{e_z}$. We can measure the similarity of the views as $\langle \boldsymbol{E_{G'_i}, E_{G''_i}} \rangle$. To enforce \textbf{similar filtering} 
of views, we accept the views if they are similar to avoid potential noisy augmentations:
$\frac {\langle \boldsymbol{E_{G'}, E_{G''}} \rangle}{
 \boldsymbol{||E_{G'}||||E_{G''}||} } > 1 - c$, for some constant $0 \leq c \leq 1\,.$ (For choice of $c$, see appendix 4.3.) We couple this filtering step with the random walk to accept candidates (Figure~\ref{fig:overview_figure}). Please note that applying similarity filtering empirically works much better than the other possible alternative, diverse filtering. We present the ablation study in appendix section 4.5.


\begin{table*}[ht]
    \caption{Datasets for pre-training, sorted by number of vertices. Bolded dataset indicates use for ablation.}
    \centering
    \small
    \begin{tabular}{l|r|r|r|r|r|r}
        \toprule
        Dataset & \textbf{DBLP~(SNAP)} & Academia & DBLP~(NetRep) & IMDB      & Facebook   & LiveJournal \\\hline
        Nodes   & 317,080  & 137,969     & 540,486       & 896,305   & 3,097,165  & 4,843,953   \\ \hline
        Edges   & 2,099,732 & 739,384    & 30,491,458    & 7,564,894 & 47,334,788 & 85,691,368  \\\bottomrule
    \end{tabular}
    \label{tbl:pretrain_dataset}
    \normalsize
\end{table*}

\subsection{Theoretical analysis.} 

We conduct a theoretical analysis of the spectral crop augmentation. In the Appendix, we extend this to a variant of the similar filtering operation. We
investigate a simple case of the two-component stochastic block model (SBM) with $2N$ nodes divided equally between classes $0,1$. These results are also extensible to certain multi-component SBMs.  Let the edge probabilities be $p$ for edges between nodes of class $0$, $q$ for edges between nodes of class $1$, and $z$ for edges between nodes of different classes. We assume that $p>q>z>0$.

Denote by $G$ a random graph from this SBM. We define the class, $Y(G)$, to be the majority of the classes of its nodes, with $Y(v)$ being the class of a node $v$. Let $E_{d,v}(G)$ denote the ego-network of $v$ up to distance $d$ in $G$. Let $C_{\epsilon}(v)$ be the cropped local neighbourhood around node $v$ defined as $\{ v' : ||\lambda(v')-\lambda(v)|| \leq \epsilon \}$ where $\lambda(v) = [\lambda_2(v), \lambda_3(v)]$, with $\lambda_j$ as the $j$-th eigenvector (sorted in ascending order by eigenvalue) of the Laplacian of $G$.  In the Appendix, we prove the following result:

\begin{theorem}
Let node $v$ be chosen uniformly at random from $G$, a $2N$-node graph generated according to the SBM described above.  With probability $\geq 1 - f(N)$ for a function $f(N) \rightarrow 0$ as $N \rightarrow \infty$, $\exists \epsilon \in \mathbb{R+}, k_{max} \in \mathbb{N}$ such that :
\begin{equation} 
\forall k \in \mathbb{N} \leq k_{max}, Y(E_{k,v}(G)) = Y(C_{\epsilon}(v)) = Y(v)
\end{equation}
\end{theorem}
This theorem states that for the SBM, both a view generated by the ego-network
and a view generated by the crop augmentation acquire, with high probability as the number of nodes grows, graph class labels that coincide with the class of the centre node. This supports  the validity of the crop augmentation --- it constructs a valid ``positive'' view. 

We further analyze global structural embeddings and similar/diverse filtering, and specify $f(N)$, in Appendix $10$.


The proof of Theorem 1 relies on the Davis-Kahan theorem. Let $\boldsymbol{A},\boldsymbol{H} \in \mathbb{R}^{N \times N}, \boldsymbol{A} = \boldsymbol{A^T}, \boldsymbol{H} = \boldsymbol{H^T}$ with $\mu_1 \geq \mu_2 \geq \dots \mu_N$ the eigenvalues of $\boldsymbol{A}$, $\boldsymbol{v}_1,\boldsymbol{v}_2,\dots, \boldsymbol{v}_N$ the corresponding eigenvectors of $\boldsymbol{A}$, and $\boldsymbol{v}'_1,\boldsymbol{v}'_2,\dots \boldsymbol{v}'_N$ those of $\boldsymbol{A}+\boldsymbol{H}$. By the Davis-Kahan theorem~\citep{demmel1997applied} (Theorem $5.4$), if the angle between $\boldsymbol{v}_i,\boldsymbol{v}'_i$ is $\theta_i$, then, with $\|\boldsymbol{H}\|_{op}$ as the max eigenvalue by magnitude of $\boldsymbol{H}$
\begin{equation}
\sin (2 \theta_i) \leq \frac { 2 \|\boldsymbol{H}\|_{op} } { N \times \min_{j \neq i} |\mu_i - \mu_j|  } \ 
\end{equation}

\label{eigenreln} In our setting, we consider $\boldsymbol{A}+\boldsymbol{H}$ to be the adjacency matrix of the observed graph, which is corrupted by some noise $\boldsymbol{H}$ applied to a ``true'' adjacency matrix $\boldsymbol{A}$. The angle $\theta_i$ measures how this noise $\boldsymbol{H}$ impacts the eigenvectors, which are used in forming Laplacian embeddings and also in the cropping step. Consider $\theta_2$, the angular error in the second eigenvector. For a normalized adjacency matrix, such that $\mu_1 = 1$, this error scales as $\frac{1}{\min (\mu_2 - \mu_3, \mu_1 - \mu_2)}$. We can anticipate that the error is larger as 
$\mu_2$ becomes larger $(\mu_1-\mu_2$ falls) or smaller ($\mu_2-\mu_3$ falls).
The error affects the quality of the crop augmentation and the quality of generated embeddings. In Section~\ref{sec:ablation}, we explore the effectiveness of the augmentations as we split datasets by their spectral properties (by an estimate of $\mu_2$). As expected, we observe that the crop augmentation is less effective for graphs with large or small (estimated) $\mu_2$.


\section{Experiments}
\vspace{-0.4cm}

{\bf Datasets.} The datasets for pretraining are summarized in Table~\ref{tbl:pretrain_dataset}. They are relatively large, with the largest graph having $\sim$4.8 million nodes and $\sim$85 million edges. Key statistics of the downstream datasets are summarized in the individual result tables. Our primary node-level datasets are US-Airport~\citep{ribeiro2017struc2vec} and H-index~\citep{zhang2019oag} while our graph datasets derive from~\citep{yanardag2015dgk} as collated in~\citep{qiu2020gcc}. Node-level tasks are at all times converted to graph-level tasks by forming an ego-graph around each node, as described in Section~\ref{sec:subgraphgeneration}. We conduct similarity search tasks over the academic graphs of data mining conferences following~\citep{zhang2019oag}. Full dataset details are in Appendix section $4$.

\textbf{Training scheme.} We use two representative contrastive learning training schemes for the graph encoder via minibatch-level contrasting (E2E) and MoCo~\citep{he2020momentum_moco} (Momentum-Contrasting). In all experiment tables, we present results where the encoder only trains on pre-train graphs and never sees target domain graphs. In Appendix $4$, we provide an additional setting where we fully fine-tune all parameters with the target domain graph after pre-training. We construct all graph encoders (ours and other baselines) as a $5$ layer GIN~\citep{xu2019GIN} for fair comparison.

{\bf Competing baselines.} As noted in our categorization of existing methods in Table~\ref{taxonomy}, the closest analogues to our approach are GraphCL~\citep{you2020graphcl} and GCC~\citep{qiu2020gcc} which serve as our key benchmarks. Additionally, although they are not designed for pre-training, we integrated the augmentation strategies from MVGRL~\citep{hassani2020MVGRL}, Grace~\citep{Grace}, Cuco~\citep{Cuco}, and Bringing Your Own View (BYOV)~\citep{BYOV_graph} to work with the pre-train setup and datasets we use. We include additional baselines that are specifically tailored for each downstream task and require unsupervised pre-training on target domain graphs instead of our pre-train graphs. We include Struc2vec~\citep{ribeiro2017struc2vec},  ProNE~\citep{zhang2019proNE}, and GraphWave~\citep{donnat2018graphwave} as baselines for the node classification task. For the graph classification task, we include Deep Graph Kernel (DGK)~\citep{yanardag2015dgk}, graph2vec~\citep{narayanan2017graph2vec}, and InfoGraph~\citep{sun2019infograph} as baselines. For the top-k similarity search method, two specialized methods are included: Panther~\citep{zhang2015panther} and RolX~\citep{henderson2012rolx}. All results for GCC are copied from~\citep{qiu2020gcc}. For GraphCL~\citep{you2020graphcl}, we re-implemented the described augmentations to work with the pre-training set up and datasets we use. We also add two strong recent benchmarks, namely InfoGCL~\citep{xu2021infogcl} and GCA~\citep{you2021graph}. Some strong baselines such as G-MIXUP~\citep{han2022g} were excluded because they require labels during the pre-training phase.


\begin{figure}[ht]
    \centering
    \includegraphics[width=8 cm]{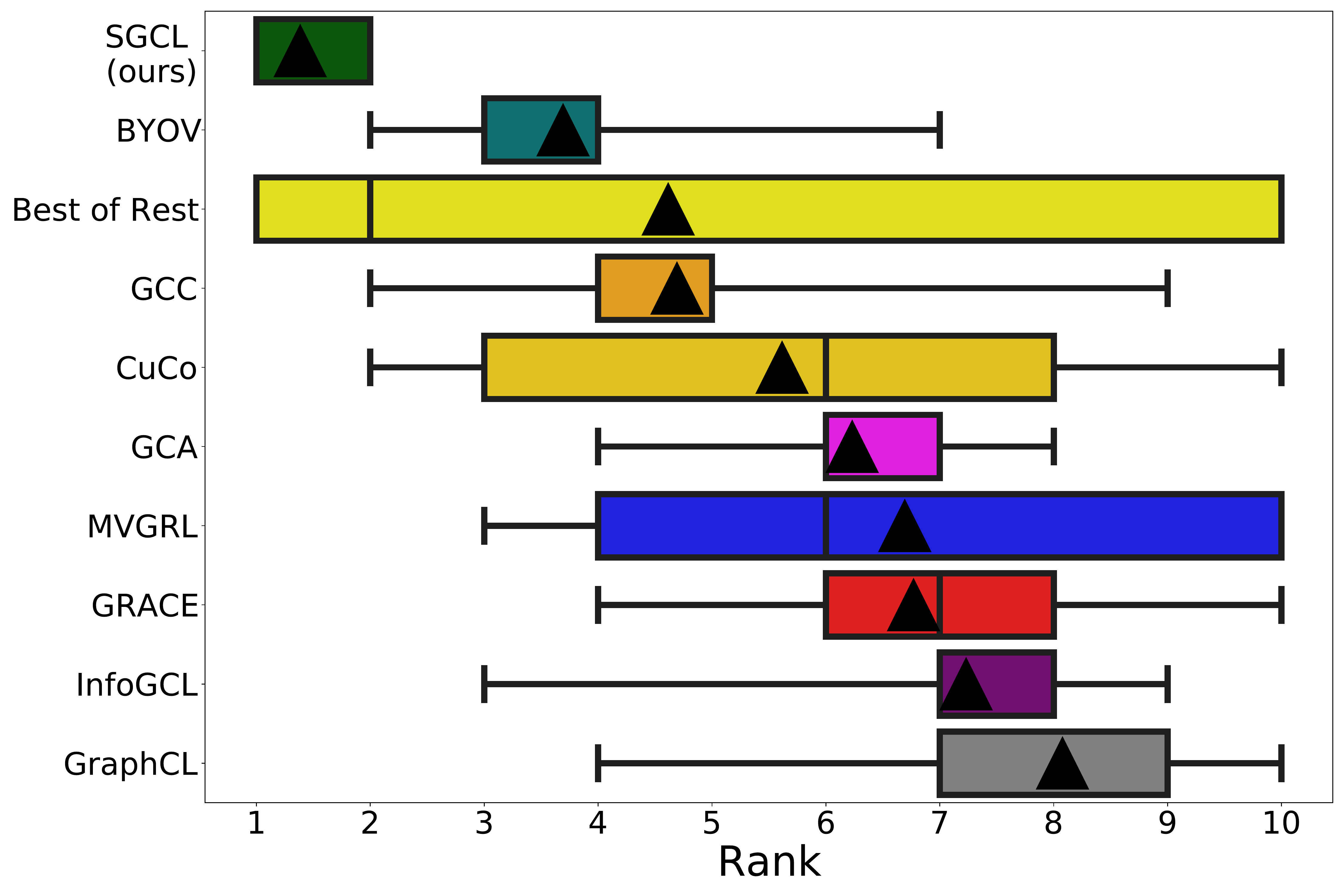}
    \caption{A rank test across 10 datasets including node and graph level classification and similarity search.}
    \label{Rankplot}
\end{figure}

\begin{figure}[ht]
    \centering
    \includegraphics[width=5 cm]{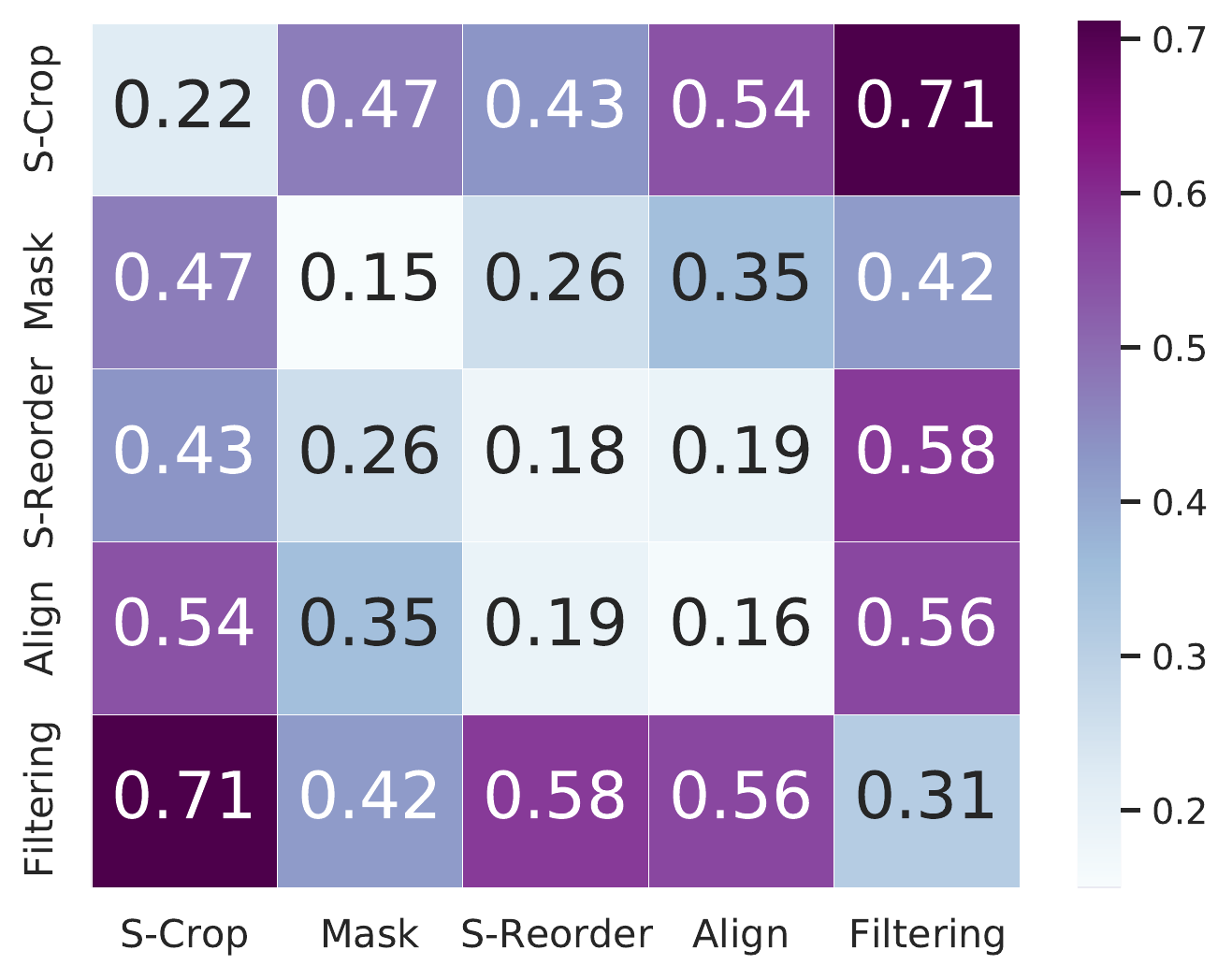}
    \caption{Heatmap indicating the effectiveness of each augmentation, pairwise, on reddit-binary. Numbers are percentage improvement relative to the SOTA method GCC~\citep{qiu2020gcc}.}
    \label{fig:Heatmapaugments}
\end{figure}






{\bf Performance metrics.} After pre-training, we train a regularized logistic regression model (node classification) or SVM classifier (graph classification) from the scikit-learn package on the obtained representations using the target graph data, and evaluate using $k=10$ fold splits of the dataset labels. Following ~\citep{qiu2020gcc}, we use F-1 score (out of $100$) as the metric for node classification tasks, accuracy percentages for graph classification, and HITS@10 (top $10$ accuracy) at top $k=20, 40$ for similarity search.

\textbf{Experimental procedure.} We carefully ensure our reported results are reproducible and accurate. We run our model 80 times with different random seeds; the seed controls the random sampler for the augmentation generation and the initialization of neural network weights. We conduct three statistical tests to compare our method with the second best baseline under both E2E and MoCo training schemes: Wilcoxon signed-rank~\citep{woolson2007wilcoxon}, Whitney-Mann~\citep{mcknight2010mann}, and the t-test. Statistical significance is declared if the p-values for all tests are less than $10^{-6}$. Appendix $4$ details hyperparameters, experimental choices and statistical methodologies. Standard deviations, statistical test results, and confidence intervals are provided in Appendix $5$, and additional experimental results for the CIFAR-10, MNIST, and OGB datasets are presented in Appendix $6$. 
\vspace{-0.3cm}

 \begin{table}[ht]
\caption{Runtime comparison (seconds per mini-batch)} \label{tb:efficiency}
\footnotesize
\begin{tabular}{@{}ccllll@{}}
\toprule
               \thead{MVGRL\\PPR} & \thead{MVGRL\\heat} & \thead{GraphCL} & \thead{GRACE}   & \thead{GCC} & \thead{SGCL\\(Ours)} \\ \midrule
$2.34$      & $0.164$      & $0.079$                     & $0.074$ & $0.063$                 & $0.084$                         \\ \bottomrule
\end{tabular}  
\vspace{-0.3cm}
\end{table}

\subsection{Runtime and scaling considerations}
Table~\ref{tb:efficiency} reports running time per mini-batch (with batch size $16$) for baselines and our proposed suite of augmentations. We observe a significant increase in computation cost for MVGRL~\citep{hassani2020MVGRL}, introduced by the graph diffusion operation~\citep{page1999pagerank,kondor2002diffusion}. The operation is more costly than the Personalized Page Rank (PPR)~\citep{page1999pagerank} based transition matrix since it requires an inversion of the adjacency matrix. Other baselines, as well as our method, are on the same scale, with GCC being the most efficient. Additional time analysis is present in Appendix section $11$.

\begin{figure}[ht]
    \centering
    \includegraphics[width=5 cm]{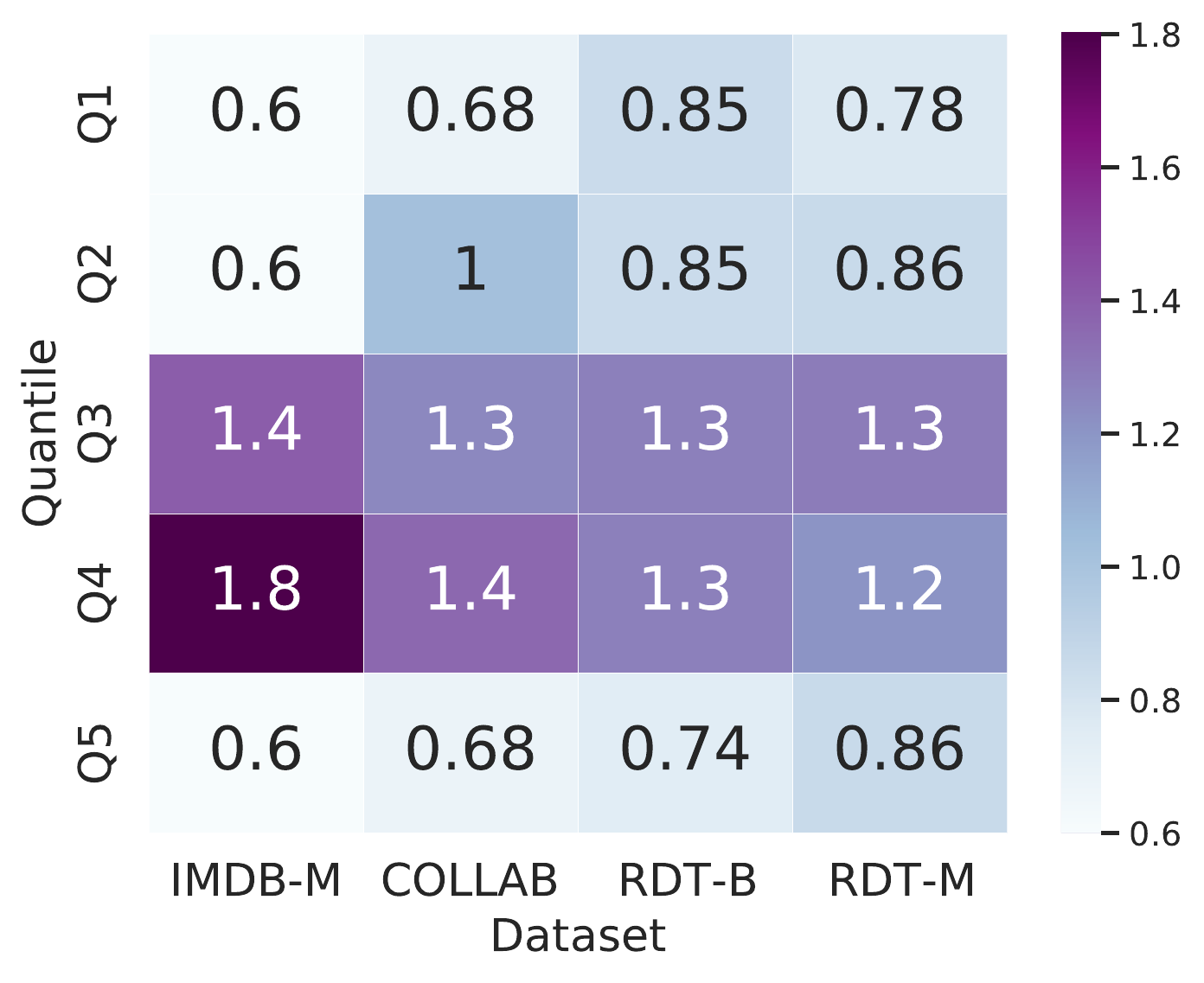}
    \caption{Heatmap indicating the effectiveness of SGCL on quintiles split by the second eigenvalue. Numbers are percentage improvement relative to the SOTA method GCC~\citep{qiu2020gcc}}
    \label{fig:Heatmaplambda2}
\end{figure}

\begin{table*}
\caption{Graph (left) and node (right) classification results  when the pre-trained graph encoder transfers to an out-of-domain graph. "-" indicates the model cannot produce reasonable results after 24 hours of training, as explained in~\citep{qiu2020gcc}.  Bold indicates best result; asterisk indicates statistical significant difference from next best. Appendix $5$ provides standard deviations and confidence intervals. }
\resizebox{1.4\columnwidth}{!}{{\begin{tabular}{@{}c|cc|cc|cc|cc|cc@{}}
\toprule
 \multicolumn{11}{c}{Graph Classification}   \\ \midrule
Datasets & \multicolumn{2}{c|}{IMDB-B} & \multicolumn{2}{c|}{IMDB-M} & \multicolumn{2}{c|}{COLLAB} & \multicolumn{2}{c|}{RDT-B} & \multicolumn{2}{c}{RDT-M} \\ \midrule
\# graphs & \multicolumn{2}{c|}{1,000} & \multicolumn{2}{c|}{1,500} & \multicolumn{2}{c|}{5,000} & \multicolumn{2}{c|}{2,000} & \multicolumn{2}{c}{5,000} \\
\# classes & \multicolumn{2}{c|}{2} & \multicolumn{2}{c|}{3} & \multicolumn{2}{c|}{3} & \multicolumn{2}{c|}{2} & \multicolumn{2}{c}{5} \\
Avg. \# nodes & \multicolumn{2}{c|}{19.8} & \multicolumn{2}{c|}{13.0} & \multicolumn{2}{c|}{74.5} & \multicolumn{2}{c|}{429.6} & \multicolumn{2}{c}{508.5} \\ \midrule
DGK & \multicolumn{2}{c|}{67.0} & \multicolumn{2}{c|}{44.6} & \multicolumn{2}{c|}{73.1} & \multicolumn{2}{c|}{78.0} & \multicolumn{2}{c}{41.3} \\
graph2vec & \multicolumn{2}{c|}{71.1} & \multicolumn{2}{c|}{\textbf{50.4}} & \multicolumn{2}{c|}{--} & \multicolumn{2}{c|}{75.8} & \multicolumn{2}{c}{47.9} \\
InfoGraph & \multicolumn{2}{c|}{73.0} & \multicolumn{2}{c|}{49.7} & \multicolumn{2}{c|}{--} & \multicolumn{2}{c|}{82.5} & \multicolumn{2}{c}{53.5} \\ \midrule
Training mode & MoCo & E2E & MoCo & E2E & MoCo & E2E & MoCo & E2E & MoCo & E2E \\
GCC & 72.0 & 71.7 & 49.4 & 49.3 & 78.9 & 74.7 & 89.8 & 87.5 & 53.7 & 52.6 \\
GraphCL & $72.2 $  & $70.9 $ &  $49.3  $  & $ 47.9 $  & $77.2$  & $ 74.1 $  & $88.7 $ &  $87.2 $    & $52.9 $ & $51.8 $  \\
GRACE & $71.7 $  & $71.5 $ &  $49.2  $  & $ 48.8 $  & $78.3 $  & $ 74.5 $  & $89.2 $ &  $87.0 $    & $53.4 $ & $52.0 $  \\
CuCo & 71.8 & 71.3  & 48.7  & 48.5  & 78.5  & 74.2  & 89.3  & 87.8  & 52.5  & 51.6  \\
BYOV & 72.3 & 72.0  & 48.5  & 49.2  & 78.4  & 75.1  & 89.5  & 87.9  & 53.6  & 53.0  \\

MVGRL & 72.3  & 72.2  & 49.2  & 49.4  & 78.6  & 75.0  & 89.6  & 87.4  & 53.4  & 52.8  \\ 
InfoGCL & 72.0  & 71.0  & 48.8  & 48.2  & 77.8  & 74.6  & 89.1  & 87.3  & 52.7  & 52.2  \\ 

GCA & 72.2  & 71.9  & 49.0  & 48.7  & 78.4  & 74.4  & 88.9  & 87.5  & 53.2  & 52.4  \\ 
SGCL & $\textbf{73.4*}$ & $\textbf{73.0} $ & $50.0  $  & $ 49.8 $  & $\textbf{79.7*} $ & $ \textbf{75.6}$   & $\textbf{90.6*} $ & $\textbf{88.4}$  & $\textbf{54.2*} $ & $\textbf{53.8} $  \\\bottomrule
\end{tabular}}} 
\hspace{0.5cm}
\resizebox{0.7\columnwidth}{!}{{\begin{tabular}{@{}c|cc|cc@{}}
\toprule
 \multicolumn{5}{c}{Node Classification}   \\ \midrule
Datasets & \multicolumn{2}{c|}{US-Airport} & \multicolumn{2}{c}{H-index} \\ \midrule
$|V|$ & \multicolumn{2}{c|}{1,190} & \multicolumn{2}{c}{5,000} \\
$|E|$ & \multicolumn{2}{c|}{13,599} & \multicolumn{2}{c}{44,020} \\\midrule
ProNE & \multicolumn{2}{c|}{62.3} & \multicolumn{2}{c}{69.1} \\
GraphWave & \multicolumn{2}{c|}{60.2} & \multicolumn{2}{c}{70.3} \\
Struc2vec & \multicolumn{2}{c|}{\textbf{66.2}} & \multicolumn{2}{c}{$\geq 1$ Day} \\ \midrule
Training mode & MoCo & E2E & MoCo & E2E \\\midrule
GCC & 65.6 & 64.8 & 75.2 & 78.3 \\ \hline \vspace{0.1cm}
GraphCL & $62.8  $  &  $63.5  $   &$74.3  $ & $76.5  $ \\
GRACE & $62.6 $  &  $63.3  $   &$74.5  $ & $77.0  $ \\
MVGRL &  $65.2  $  &  $64.5  $   & $75.1  $ & $78.1 $ \\  
CuCo &  $64.9  $  &  $64.3 $   & $75.3  $ & $78.2  $ \\
BYOV &  ${65.3}  $ &   $64.7 $  & $76.0  $   & ${78.1} $ \\
InfoGCL &  ${63.2}  $ &  $ 64.1 $  & $75.4 $   & ${77.6} $ \\
GCA &  ${64.5}  $ &   $64.3 $  & $ 75.8  $   & ${78.0} $ \\
SGCL & ${65.9}  $ &   $65.3 $  & $\textbf{76.7}  $   & $\textbf{78.9*}$ \\ \bottomrule
\end{tabular}}}

\end{table*}
\label{tab:nlandgl}
\vspace{-0.3cm}
\subsection{Results and Discussion}
\vspace{-0.3cm}

We report the performance of our model as well as other baselines on node classification and on graph classification in Table~\ref{tab:nlandgl}. Top-$k$ similarity search results are provided in Appendix 4.6. We report both the average performance across 80 trials and confidence intervals (Appendix $5$) for our proposed design, GraphCL~\citep{you2020graphcl}, MVGRL~\citep{hassani2020MVGRL}, Grace~\citep{Grace}, Cuco~\citep{Cuco} and BYOV~\citep{BYOV_graph}. Confidence intervals indicate a span between the $5$-th and $95$-th percentiles, estimated by bootstrapping over splits and random seeds. For the other baselines, we copy the results reported in~\citep{qiu2020gcc}.

Our design achieves robust improvement on both node and graph classification tasks over other baselines for the domain transfer setting. We emphasize that the graph encoders for all the baselines from \textit{Category 5} in Table~\ref{tab:approach-comparison-init} are not trained on the target source dataset, whereas other baselines use this as training data (in an unsupervised fashion). Although this handicaps the domain transfer-based methods, our proposed method performs competitively or even significantly better compared to classic unsupervised learning approaches including ProNE~\citep{zhang2019proNE}, GraphWave~\citep{donnat2018graphwave} and Struc2vec~\citep{ribeiro2017struc2vec} for node-level classification tasks and DGK, graph2vec and InfoGraph for graph level classification.
We observe similar improvements relative to baselines for both the E2E and MoCo training schemes. These improvements are also evident for the similarity search task. 
The performance gains are also present when the encoder is fully fine-tuned on graphs from the downstream task, but due to space limitations, we present the results in Appendix $4$.



\textbf{Effectiveness of individual augmentations, processing/selection steps, and pairwise compositions.} \label{sec:ablation} We show evaluation results (average over 80 trials) for both individual
augmentations or filtering/selection steps and their pairwise compositions in Figure~\ref{fig:Heatmapaugments}. For a clear demonstration, we select Reddit-binary as the downstream task and the smallest pre-train DBLP (SNAP) dataset. Using more pre-train datasets should result in further performance improvements. The full ablation study results are presented in Appendix $4$. As noted previously~\citep{you2020graphcl}, combining augmentations often improves the outcome. We report improvement relative to the SOTA method GCC~\citep{qiu2020gcc}. Performance gains are observed for all augmentations. On average across 7 datasets, spectral crop emerges as the best augmentation of those we proposed. Appendix 4.5 reports the results of ablations against random variants of the crop and reorder augmentations; the specific procedures we propose lead to a substantial performance improvement.


\textbf{Performance variations due to spectral properties.} We split the test graphs into quintiles based on their $\lambda_2$ values to explore whether the test graph spectrum impacts the performance of the proposed augmentation process. Figure \ref{fig:Heatmaplambda2} displays the improvements obtained for each quintile. As suggested by our theoretical analysis in Section~\ref{eigenreln}, we see a marked elevation for the middle quintiles of $\lambda_2$. These results support the conjecture that small or large values of $\lambda_2$ (an approximation of $\mu_2$ in Section~\ref{eigenreln}) adversely affect the embedding quality and the crop augmentation.

\vspace{-0.3cm}
\section{Conclusion}

We introduce \textbf{SGCL}, a comprehensive suite of spectral augmentation methods suited to pre-training graph neural networks contrastively over large scale pre-train datasets. The proposed methods do not require labels or attributes, being reliant only on structure, and thus are applicable to a wide variety of settings.  We show that our designed augmentations can aid the pre-training procedure to capture generalizable structural properties that are agnostic to downstream tasks. Our designs are not ad hoc, but are well motivated through spectral analysis of the graph and its connections to augmentations and other techniques in the domains of vision and network embedding analysis. The proposed augmentations make the graph encoder --- trained by either E2E or MoCo --- able to adapt to new datasets without fine-tuning. The suite outperforms the previous state-of-the-art methods with statistical significance. The observed improvements persist across multiple datasets for the three tasks of node classification, graph classification and similarity search.

\clearpage

\bibliography{newattempt}
\bibliographystyle{abbrvnat}

%

%

\onecolumn
\aistatstitle{Supplementary Materials for SGCL}

\section*{List of all contents}

\begin{itemize}
    \item Sections $1$ to $3$ position our work better and provide references.
    \item Section $4$ has implementation details, ablations, statistical tests, and all other results not in the main text. Section $5$ adds confidence intervals and some more results and confidence intervals.
    \item Section $6$ has some results for MNIST and CIFAR-10 to illustrate linkages between our method and image cropping.
    \item Section $7$ discusses limitations, societal impact and reproducibility.
    \item Section $8$ has additional ablations based on spectral properties.
    \item Section $9$ has proofs of some mathematical results used in the paper for various augmentations.
    \item Section $10$ has proofs on the stochastic block model.
    \item Sections $11,12,13$ add small details that help explain how the entire suite works. Section $11$ has time complexity graphs, section $12$ has a case of negative transfer which pops up in some experiments and looks anomalous otherwise, and section $13$ visualizes why we need to align two graphs in the first place. 
\end{itemize}

\clearpage

\section{Position of our work}\label{sec:appendix_position}
We emphasize that there are three distinct cases of consideration in the field of contrastive learning in graphs. We hope to make a clear separation between them to help readers understand our model design and the choice of baselines as well as the datasets we conduct experiments on. A summarizing of the position of our work vs. prior works is presented in Table~\ref{tab:appendix_position}.

\begin{table}[h]
\centering
\caption{Detailed analysis of our work vs. existing work}
{\resizebox{\textwidth}{!}{
\begin{tabular}{@{}cccccc@{}}
\toprule
 & \begin{tabular}[c]{@{}c@{}}Pre-train with \\ Freeze encoder\\ (Out of domain)\end{tabular} & \begin{tabular}[c]{@{}c@{}}Pre-train with\\ encoder fine-tune\\ (Out of domain)\end{tabular} & \begin{tabular}[c]{@{}c@{}}Pre-train with \\ Freeze encoder\\ (Same domain)\end{tabular}& \begin{tabular}[c]{@{}c@{}}Pre-train with\\ encoder fine-tune\\ (Same domain)\end{tabular}& \begin{tabular}[c]{@{}c@{}}Unsupervised\\ learning\end{tabular} \\ \midrule
\begin{tabular}[c]{@{}c@{}}Training \\ Data\end{tabular} & $\{\textbf{X}_{LP},\textbf{A}\}_\textrm{pretrain}$ & \begin{tabular}[c]{@{}c@{}}$\{\textbf{X}_{LP},\textbf{A}\}_\textrm{pretrain}$\\ $\{\textbf{X}_{LP},\textbf{A}, \textbf{Y}\}_\textrm{downstream}$\end{tabular}  &$\{\textbf{X},\textbf{A}\}_\textrm{pretrain}$ & \begin{tabular}[c]{@{}c@{}}$\{\textbf{X},\textbf{A}\}_\textrm{pretrain}$\\ $\{\textbf{X},\textbf{A}, \textbf{Y}\}_\textrm{downstream}$\end{tabular}  & $\{\textbf{X},\textbf{A}\}_\textrm{downstream}$ \\
 \begin{tabular}[c]{@{}c@{}}Methods in \\ the category\end{tabular} & Ours,GCC &  Ours,GCC  & \begin{tabular}[c]{@{}c@{}}GPT-GNN, GraphCL, MICRO-Graph\\ JOAO, BYOV, GraphMAE\end{tabular}  & \begin{tabular}[c]{@{}c@{}}GPT-GNN, GraphCL, MICRO-Graph\\ JOAO, BYOV, GraphMAE \end{tabular} & \begin{tabular}[c]{@{}c@{}}InfoGCL, MVGRL\\ GraphCL, GCA \end{tabular} \\
 
 
\bottomrule
\end{tabular}}}
\label{tab:appendix_position}
\end{table}

First, we have out of domain pre-training, where the encoder only sees a large pre-training corpus in the training phase that may share no node attributes at all with the downstream task. A representative method is GCC~\citep{qiu2020gcc}. For example, this pre-training dataset can be a large citation network or social network (as the pre-train corpus used in our paper), while the downstream task may be on a completely different domain such as molecules. Since no node attributes are shared between the two domains, the initial node attributes have to rely solely on structure, e.g., the adjacency or Laplacian matrices and embeddings derived from their eigendecomposition . Importantly, in this case, the encoder is never trained on any labeled or unlabeled instances for the downstream graph related tasks before doing the inference. It allows the model to obtain the results on the downstream task very fast (since only the model inference step is applied to obtain the node representations for the downstream tasks). We call this setting pre-training with frozen encoder (out of domain). This is the most difficult graph contrastive learning (GCL) related task. In our paper, we strictly follow this setup. The downstream task performance can be further improved if the downstream training instances (data, but also possibly with labels) are shown to the GNN encoder. We call this setting pre-training with fine-tuning (out of domain).

Second, we have the domain specific pre-training method where the encoder sees a large pre-training corpus which shares similar features or the same feature tokenization method as the downstream task. The representative methods that fall under this category include GPT-GNN~\cite{hu2020gpt-gnn}, GraphCL~\citep{you2020graphcl}, MICRO-Graph~\citep{micro-graph}, JOAO~\citep{you2021graph}, BYOV~\citep{BYOV_graph}, and GraphMAE~\citep{hou2022graphmae}. The typical experiment design for this setting is to pre-train the GNN encoder on a large-scale bioinformatics dataset, and then fine-tune and evaluate on smaller datasets of the same category. Since the feature space is properly aligned between the pre-train dataset and the downstream datasets, the node attributes usually are fully exploited during the pre-training stage. Similarly, the downstream task performance can be further improved if the downstream tasks (data and/or their labels) are shown to the GNN encoder. We call these two setting pre-training with frozen encoder (same domain) and pre-training with encoder fine-tuning (same domain).

Third, in the unsupervised learning setting for GCL, there is no large pre-training corpus that is distinct from the downstream task data. Rather, the large training set of the downstream task is the sole material for contrastive pre-training. Note that in this case, if there are multiple unrelated downstream tasks, e.g., a citation network and also a molecule task, a separate pre-training procedure must be conducted for each task and a separate network must be trained. The representative methods that fall under this category include InfoGCL~\citep{xu2021infogcl}, MVGRL~\citep{hassani2020MVGRL}, GraphCL~\citep{you2020graphcl}, GCA~\citep{GCA}. Generally speaking, for tasks that rely heavily on the node attributes (such as citations, and molecule graphs), such unsupervised methods, when the training set data (adjacency matrix and node attributes) is available for the unsupervised training phase, can potentially outperform the out of domain pre-trained frozen encoder case. But this is natural, and expected, because in the out-of-domain pre-training with a frozen encoder setting the pre-trained network never even sees the source domain. It can never take advantage of node attributes because the pre-train datasets do not share the same feature space as the downstream task. It can only rely on the potential transferable structural features. But this is also not its purpose - its purpose is to act like a general large language model (LLM) or a Foundation Model like GPT-3. Such a model is not necessarily an expert in every area and can be outperformed by, for instance, specific question-answer-specialized language models for answering questions, but it performs relatively well zero-shot in most tasks without needing any pre-training. This is why in our main paper, we did not compare with the commonly used small-scale datasets (Cora, Citesser) for the unsupervised learning tasks.

Previous papers in this field such as GraphMAE~\citep{hou2022graphmae} often include the frozen, pre-trained models under the unsupervised category in the experiments, which is not completely accurate or fair. In fact, this category is relatively understudied and introduces unique challenges for the out-of-domain transfer setting. Its importance, and relative lack of study, is precisely why it deserves attention - it is a step toward out-of-domain generalization on graphs and avoids expensive pre-training in every domain. In the following table, we provide a novel way to categorize the existing graph contrastive learning work and we hope it provides better insight to the readers in terms of the position of our work.

Please note that even though not directly applicable for the pre-train (out of domain) mode, for the existing methods under the category of pre-train (same domain) and unsupervised learning, we are able to make modifications to allow them to be applied in the out of domain settings. The main changes are 1) we use the pre-train out of domain corpus $\{\textbf{X}_{LP},\textbf{A}\}_\textrm{pretrain}$ to train the GNN encoder instead of $\{\textbf{X},\textbf{A}\}_\textrm{downstream}$; 2) since the feature space between the pre-train domains and the target domain are not aligned, we use  $\textbf{X}_{LP}$ instead of the original feature $\textbf{X}$. We conduct the above modification to some of the existing unsupervised learning based methods such as MVGRL~\citep{hassani2020MVGRL} and GraphCL~\citep{you2020graphcl}. This is also why we cannot, without running the out of domain pre-training experiment setups, directly report the experimental performance in the previous papers (largely unsupervised, except GCC), and why some of our numbers do not always agree with those reported in the original paper, e.g., MVGRL on IMDB-BINARY. The entire training process is completely different with a new pre-training corpus, and the same numbers are not guaranteed to occur.

\section{Reasons to pursue augmentations based on graph spectra}

First, we want to address the question of what is meant by the term "universal topological properties". Our method is inherently focused on transferring the pre-trained GNN encoder to any domain of graphs, including those that share no node attributes with the pre-train data corpus. This means that the only properties the encoder can use when building its representation are transferable structural clues. We use the word topological to denote this structure-only learning. The word universal denotes the idea of being able to easily transfer from pre-train graphs to any downstream graphs. It is a common practice to augment node descriptors with structural features~\citep{errica2020fair}, especially for graph classification tasks. DiffPool~\citep{ying2018hierarchical} adds the degree and clustering coefficient to each node feature vector. GIN~\citep{xu2019GIN} adds a one-hot representation of node degrees. In short, a universal topological property is some property such as the human-defined property of "degree" that we hope the GNN will learn in an unsupervised fashion. Just as degree - a very useful attribute to know for any graph for many downstream tasks - is derivable from the adjacency matrix by taking a row sum, we hope the GNN will learn a sequence of operations that distill some concept that is even more meaningful than the degree and other basic graph statistics.

Since structural clues are the only ones that can be transferable between pre-train graphs and the downstream graphs, the next part to answer is why spectral methods, and why should we use the spectral-inspired augmentations to achieve the out-of-domain generalization goal. We elaborate as follows.

For multiple decades, researchers have demonstrated the success of graph spectral signals with respect to preserving the unique structural characteristics of graphs (see~\citep{torres2020glee} and references therein). Graph spectral analysis has also been the subject of extensive theoretical study and it has been established that the graph spectral information is important to characterize the graph properties. For example, graph spectral values (such as the Fiedler eigenvalue) related directly to fundamental properties such as graph partitioning properties~\citep{kwok2013improved,lee2014multiway} and graph connectivity~\citep{chung1997spectral,kahale1995eigenvalues,fiedler1973algebraic}. Spectral analyses of the Laplacian matrix have well-established applications in graph theory, network science, graph mining, and dimensionality reduction for graphs~\citep{torres2020glee}. They have also been used for important tasks such as clustering~\citep{belkin2001laplacian,von2007tutorial} and sparsification~\citep{spielman2008graph}. Moreover, many network embedding methods such as LINE~\citep{tang2015line} and DeepWalk reduce to factorizing a matrix derived from the Laplacian, as addressed in NetMF~\citep{qiu2018netfm}. These graph spectral clues allow us to extract transferable structural features and structural commonality across graphs from different domains. All of these considerations motivate us to use spectral-inspired augmentations for graph contrastive learning to fully exploit the potential universal topological properties across graphs from different domains.

\section{Related Work Extension}\label{sec:related_work}
\subsection{Representation learning on graphs}

A significant body of research focuses on using graph neural networks to encode both the underlying graph describing relationships between nodes as well as the attributes for each  node~\citep{kipf2017gcn,gilmer2017mpnn,graphsage,velivckovic2017gat,xu2019GIN}. The core idea for GNNs is to perform feature mapping and recursive neighborhood aggregation based on the local neighborhood using shared aggregation functions. The neighborhood feature mapping and aggregation steps can be parameterized by learnable weights, which together constitute the graph encoder.  

There has been a rich vein of literature that discusses how to design an effective graph encoding function that can leverage both node attributes and  structure information to learn representations~\citep{kipf2017gcn,graphsage,velivckovic2017gat,zhang2018gaan,wu2019SGCN, chen2020GCN2, xu2019GIN}. In particular, we highlight the Graph Isomorphism Network (GIN)~\citep{xu2019GIN}, which is an architecture that is provably one of the most expressive among the class of GNNs and is as powerful as the Weisfeiler Lehman graph isomorphism test.
Graph encoding is a crucial component of GNN pre-training and self-supervised learning methods. However, most  existing graph encoders are based on message passing and the transformation of the initial node attributes. Such encoders can only capture vertex similarity based on features or node proximity, and are thus restricted to being domain-specific, incapable of achieving transfer to unseen or out-of-distribution graphs. In this work, to circumvent this issue, we employ structural positional encoding to construct the initial node attributes. 
By focusing on each node's local subgraph level representation, we can extract universal topological properties that apply across multiple graphs. This endows the resultant graph encoder with the potential to achieve out-of-domain graph data transfer.


\subsection{Data augmentations for contrastive learning}
\paragraph{Augmentations for image data.} Representation learning is of perennial importance in machine learning with contrastive learning being a recent prominent technique. In the field of representation learning for image data, under this framework, there has been an active research theme in terms of defining a set of transformations applied to image samples, which do not change the semantics of the image. Candidate transformations include cropping, resizing, Gaussian blur, rotation, and color distortion. Recent experimental studies~\citep{chen2020simclr,grill2020BYOL} have highlighted that the combination of random crop and color distortion can lead to significantly improved performance for image contrastive learning. Inspired by this observation, we seek analogous graph augmentations.

\paragraph{Augmentations for graph data.} The unique nature of graph data means that the augmentation strategy plays a key role in the success of graph contrastive learning~\citep{qiu2020gcc, you2020graphcl, li2021dgcl, sun2019infograph,hassani2020MVGRL, xu2021infogcl}. Commonly-used graph augmentations include: 1) \textit{attribute dropping or masking}~\citep{you2020graphcl,hu2020gpt-gnn}: these graph feature augmentations rely heavily on domain knowledge and this prevents learning a domain invariant encoder that can transfer to out-of-domain downstream tasks; 2) \textit{random edge/node dropping}~\citep{li2021dgcl,xu2021infogcl,Grace,GCA}: these augmentations are based on heuristics and they are not tailored to preserve any special graph properties; 3) \textit{graph diffusion}~\citep{hassani2020MVGRL}: this operation offers a novel way to generate positive samples, but it has a large additional computation cost~\citep{hassani2020MVGRL, page1999pagerank,kondor2002diffusion}. The graph diffusion operation is more costly than calculation and application of the Personalized Page Rank (PPR)~\citep{page1999pagerank} based transition matrix since it requires an inversion of the adjacency matrix;  and 4) \textit{random walks around a center node}~\citep{tong2006fastrw,qiu2020gcc}: this augmentation creates two independent random walks from each vertex that explore its ego network and these form multiple views (subgraphs) of each node.
Additionally, there is an augmentation called GraphCrop~\citep{Graphcrop}, which uses a node-centric strategy to crop a contiguous subgraph from the original graph while maintaining its connectivity; this is different from the spectral graph cropping we propose. Existing structure augmentation strategies are not tailored to any special graph properties and might unexpectedly change the semantics~\citep{AFGRL}.

\subsection{Pre-training, self-supervision, unsupervised \&\ contrastive graph representation learning}

Though not identical, pre-training, self-supervised learning, and contrastive learning approaches in the graph learning domain use many of the same underlying methods. A simple technique such as attribute masking can, for example, be used in pre-training as a surrogate task of predicting the masked attribute, while in the contrastive learning scenario, the masking is treated as an augmentation. We categorize the existing work into the following 5 categories.

\begin{table}
\centering
\caption{Properties of different approaches to graph contrastive (unsupervised) learning. ${^{\star}}$ indicates that the method was not originally designed for pre-training, but can be trivially adapted to it. \textbf{A fuller description with relevant references is added in the appendix.}}
\label{tab:approach-comparison}
{\resizebox{\textwidth}{!}{
\begin{tabular}{l|ccccc}
\toprule
Approaches& \makecell{Goal is pre-training \\ or transfer} & \makecell{No  requirement \\ for features} & \makecell{Domain \\ transfer} & \makecell{Shareable graph \\ encoder } \\  
  \midrule
Category 1  (DGI, InfoGraph, MVGRL, DGCL, InfoGCL, AFGRL)  & \XSolidBrush & \XSolidBrush & \XSolidBrush & \XSolidBrush\\

Category 2 (GPT-GNN, Strategies for pre-training GNNs)  & \CheckmarkBold & \XSolidBrush & \XSolidBrush & \CheckmarkBold \\
Category 3  (Deepwalk, LINE, node2vec) & \XSolidBrush & \CheckmarkBold &  \CheckmarkBold & \XSolidBrush \\

Category 4 (struc2vec, graph2vec, DGK, Graphwave, InfiniteWalk)  & \XSolidBrush & \CheckmarkBold &  \CheckmarkBold & \XSolidBrush \\

Category 5 (GraphCL, CuCo${^{\star}}$, GCC, BYOV, GRACE${^{\star}}$, GCA${^{\star}}$, Ours)& \CheckmarkBold & \CheckmarkBold & \CheckmarkBold & \CheckmarkBold \\

\bottomrule
\end{tabular}}}
\label{taxonomy_app}
\end{table}

\paragraph{Category 1.} One of the early works in the contrastive graph learning direction is Deep Graph Infomax (DGI)~\citep{velickovic2019DGI}. Though not formally identified as contrastive learning, the method aims to maximize the mutual information between the patch-level and high-level summaries of a graph, which may be thought of as two views. Infomax is a similar method that uses a GIN (Graph Isomorphism Network) and avoids the costly negative sampling by using batch-wise generation~\citep{sun2019infograph}. MVGRL~\citep{hassani2020MVGRL} tackles the case of multiple views, i.e., positive pairs per instance, similar to Contrastive Multiview coding for images. DGCL~\citep{li2021dgcl} adopts a disentanglement approach, ensuring that the representation can be factored into components that capture distinct aspects of the graph. InfoGCL~\citep{xu2021infogcl} learns representations using the Information Bottleneck (IB) to ensure that the views minimize overlapping information while preserving as much label-relevant information as possible. None of the methods in this category is capable of capturing universal topological properties that extend across multiple graphs from different domains.

\paragraph{Category 2.} Predicting masked edges/attributes in chemical and biological contexts has emerged as a successful pre-train task.  GPT-GNN~\citep{hu2020gpt-gnn} performs generative pre-training successively over the graph structure and relevant attributes. In~\citep{hu2019strategies_gnn}, Hu et al.\ propose several strategies (attribute masking, context structure prediction) to pre-train GNNs with joint node-level and graph-level contrastive objectives. This allows the model to better encode domain-specific knowledge. However, the predictive task for these methods relies heavily on the features and domain knowledge. As a result, the methods are not easily applied to general graph learning problems.


\paragraph{Category 3} Random-walk-based embedding methods like Deepwalk~\citep{perozzi2014deepwalk}, LINE~\citep{tang2015line}, and Node2vec~\citep{grover2016node2vec} are widely used to learn network embeddings in an unsupervised way. The main purpose is to encode the similarity by measuring the proximity between nodes. The embeddings are derived from the skip-gram encoding method, Word2vec, in Natural Language Processing (NLP). However, the proximity similarity information can only be applied within the same graph. Transfer to unseen graphs is challenging since the embeddings learned on different graphs are not naturally aligned.


\paragraph{Category 4} To aid transferring learned representations, another approach of unsupervised learning attempts encoding structural similarities. Two nodes can be structurally similar while belonging to two different graphs.  Handcrafted domain knowledge based representative structural patterns  are proposed in~\citep{yanardag2015dgk, ribeiro2017struc2vec,narayanan2017graph2vec}. Spectral graph theory provides the foundation for modelling structural similarity in~\citep{qiu2018netfm, donnat2018graphwave,chanpuriya2020infinitewalk}. 

\paragraph{Category 5}
In the domain of explicitly contrastive graph learning, we consider Graph Contrastive Coding (GCC)~\citep{qiu2020gcc} as the closest approach to our work. In GCC, the core augmentation used is random walk with return~\citep{tong2006fastrw}. This forms multiple views (subgraphs) of each node. GraphCL~\citep{you2020graphcl} expands this augmentation suite to add node dropping, edge perturbations, and attribute masking. Additionally, although they are not designed for pre-training, we integrated the augmentation strategies from MVGRL~\citep{hassani2020MVGRL}, Grace~\citep{Grace}, Cuco~\citep{Cuco}, and Bringing Your Own View (BYOV)~\citep{BYOV_graph} to work with the pre-train setup in category 5.





\section{Implementation details, additional results, and ablations}\label{sec:appendix_ci}

\subsection{Codebase references}

In general, we follow the code base of GCC~\citep{qiu2020gcc}, provided at : \href{https://github.com/THUDM/GCC}{https://github.com/THUDM/GCC}. We use it as a base for our own implementation (provided along with supplement). Please refer to it in general with this section. For results using struc2vec~\citep{ribeiro2017struc2vec}, ProNE~\citep{zhang2019proNE}, Panther~\citep{zhang2015panther}, RolX~\citep{henderson2012rolx} and graphwave~\citep{donnat2018graphwave} we report the results directly from~\citep{qiu2020gcc} wherever applicable.

\subsection{Dataset details}

We provide the important details of the pre-training datasets in the main paper, so here we describe the downstream datasets. We obtain US-airport from the core repository of GCC~\citep{qiu2020gcc} which itself obtains it from ~\citep{ribeiro2017struc2vec}. H-index is obtained from GCC as well via OAG~\citep{zhang2019oag}. COLLAB, REDDIT-BINARY, REDDIT-MULTI5K, IMDB-BINARY, IMDB-MULTI all originally derive from the graph kernel benchmarks~\citep{morris2020tudataset}, provided at : \href{https://chrsmrrs.github.io/datasets/}{https://chrsmrrs.github.io/datasets/}. Finally, the top-k similarity datasets namely KDD-ICDM,SIGIR-CIKM,and SIGMOD-ICDE, are obtained from the GCC repository~\citep{qiu2020gcc}; these were obtained from the original source, Panther~\citep{zhang2015panther}.

\subsection{Hyperparameters and statistical experimental methodology}

\textbf{Hyperparameters}: Training occurs over $75,000$ steps with a linear ramping-on (over the first $10 \%$) and linear decay (over the last $10 \%$) using the ADAM optimizer, with an initial learning rate of $0.005$, $\beta_1=0.9, \beta_2=0.999, \epsilon = 10^{-8}$. The random walk return probability is $0.8$. The $E2E$ dictionary size $K=1023$, for MoCo $16384$. The batch size is $1024$ for $E2E$ and $32$ for MoCo. The dropout is set to $0.5$ with a degree embedding of dimension $16$ and positional embedding of dimension $64$. These hyperparameters are retained from GCC and do not require grid search. The hyperparameter $c$ for alignment is chosen by grid search from $6$ values, namely $0.2,0.25,0.3,0.35,0.4,0.45$.

\textbf{Runtime}: Using DBLP as the test bed, we observed $33.27$ seconds per epoch for baseline GCC, which was only increased to at most $41.08$ seconds in the settings with the most augmentations. Note that epoch time is largely CPU controlled in our experience and may vary from server to server. However, we found that the ratios between different methods were far more stable. The main paper reports these values on a per graph basis.

\paragraph{Confidence intervals and statistical methodology} : To construct confidence bounds around our results, we carry out the following procedure. We introduce some randomness through seeds. The seed is employed twice: once during training the encoder, and again while fine-tuning the encoder on downstream tasks on the datasets. We carry out training with $8$ random seeds, resulting in $8$ encoders. From each of these encoders, the representation of the graphs in the downstream datasets is extracted.

Next, we train a SVC (for graph datasets) or a logistic regression module (for node datasets), both with a regularization co-efficient over $10$ stratified K-fold splits. Before testing the model obtained from the train fraction of any split, we sample uniformly with replacement from the test set of the split of size $T$ until we draw $T$ samples. These instances are graphs (ego-graphs for the case of node datasets and distinct graphs for the case of graph datasets). After this we report the testing result. This is a bootstrapping procedure that leads to a random re-weighing of test samples. This entire process - i.e., generating a new $10$-fold split, training, bootstrapping, testing - is repeated $10$ times per encoder.

This leads to a total of $800$ data points per encoder, allowing fine-grained confidence intervals. However, we found that performance varied too strongly as a function of the splits, leading us to average over the splits instead. Therefore, each determination carries an effective sample size of $80$. Upon this, the Whitney-Mann, Wilcoxon signed rank, and t-tests are carried out to determine p-values, $5$ to $95$ percentile confidence bounds, and standard deviations.

\setcounter{table}{0}
\setcounter{figure}{0}

\subsection{Augmentation details and sequences} \label{augflow}

\paragraph{Masking:} As mentioned in the main paper, we follow previous work~\citep{hu2019strategies_gnn} and add simple masking of $\boldsymbol{X_i}$. The masking involves setting some columns to zero. Since we consider smaller eigenvalues of
$\boldsymbol{L_i}$ to be more important, we draw an integer $z$
uniformly in the range $[0,M]$ and mask out $z$ eigenvectors
corresponding to the top $z$ eigenvalues of $\boldsymbol{L_i}$.

\paragraph{Sequence of augmentations:} We have discussed the creation of views of $G'_i$ from graph instances $G_i$. However, in our case, the goal is to create two positive views $G'_i,G''_i$ per mini-batch for an instance $G_i$. Let us now clarify the sequence of augmentations we employ. It should be understood that for any $G_i$, the negative view is any augmented or un-augmented view of $G_j, j \neq i$.

\begin{itemize}
\item First, we create $G'_i, G''_i$ using random walk with return on $G_i$. We use the random walk hyperparameters identified in~\citep{qiu2020gcc}.
\item With probability $p_{filter}$, we then test $G'_i,G''_i$ using similarity or diversity thresholds $1{-}c$ and repeat the first step if the test fails. If $G'_i,G''_i$ do not pass in $t_{max}$ tries, we proceed to the next step.
\item We randomly crop both $G'_i,G''_i$ independently with probabilities over different crops $c_1,c_2,\dots$ (including no crop). In total, we allow five outcomes, i.e. $c_1, c_2, c_3, c_4, c_5 = 1 - \sum_{i=1}^{4} c_i$. The last outcome is the case of no cropping. We keep $c_1 = c_2$, $c_3 = c_4$. These correspond to different types of crops, explained below. We can term $c_1 + c_2 + c_3 + c_4$ as $p_{crop}$.
\item With probability $p_{align}$, we replace $\mathbf{X'_i, X''_i}$ with $\mathbf{X'_i Q^*, X''_i Q^{**}}$ or keep $\mathbf{X'_i, X''_i}$ unchanged with probability $1 - p_{align}$.

\item We apply one of the mask and reorder augmentations on both $G'_i, G''_i$ independently to form the final positive pairs. That is, for $G'_i$, we mask it with $p_{mask}$, or reorder it with $p_{reorder}$, or keep it unchanged with $1-p_{mask} - p_{reorder}$. The same process is then done, independently, for $G''_i$.
\end{itemize}

For a graph $G$, the $x_{min}, x_{max}, y_{min}, y_{max}$ values for cropping are chosen as follows. We calculate the values taken by the second eigenvector over $G$ and rank them, and set $x_{min}$ on the basis of the rank. That is, $x_{min} = R_{0.2}$ would correspond to $x_{min}$ being set as the $20$-th percentile value over $G$. This is done instead of absolute thresholds to sidestep the changes of the second eigenvector over the different $G$s. The corresponding different types of crop are, written as $R$ values $[x_{min},x_{max},y_{min},y_{max}]$ tuples :

\begin{itemize}
\item $[R_{0.2},R_{0.8},R_{0.2},R_{0.8}]$ with $c_1 = 0.1$
\item$[R_{0.1},R_{0.9},R_{0.1},R_{0.9}]$ with $c_2 = 0.1$
\item $[R_{0},R_{0.8},R_{0},R_{0.8}]$ with $c_3 = 0.05$
\item$[R_{0.2},R_{1.0},R_{0.2},R_{1.0}]$ with $c_4 = 0.05$
\item No crop, with $c_5 = 0.7$
\end{itemize}

Note that in terms of alignment augmentations, our arguments regarding $\mathbf{X'_i, X''_i}$ being transformed to $\mathbf{X'_i Q^*, X''_i Q^{**}}$ as it respects the inner product carry over if we instead use $\mathbf{X'_i, X''_i Q^{**}  (Q^*)^T}$ or $\mathbf{X'_i Q^* (Q^{**})^T, X''_i }$ as the augmented views.

\textbf{Order of augmentations.} We have chosen the augmentations to proceed in this order due to the following reasons. 

\begin{itemize}
\item Of all the candidates for the first augmentation in our sequence, the random walk is supreme as it cuts down on the size of the ego-net for the future steps and the Laplacian eigendecomposition's complexity as well. Doing away with it greatly increases the runtime for any other step preceding it.

\item Filtering is best done as early as possible to reject candidates on its basis before expensive augmentation steps have already been performed. Hence, we place it second.

\item Cropping precedes align, mask and ordering as these change the attribute vectors, and cropping uses the second eigenvector which is part of the embedding itself.

\item Alignment precedes mask and reorder, as alignment on shifted embeddings post-mask or post-reorder no longer follows from our arguments of its necessity.

\item Mask and reorder are mutually exclusive as reordering a masked matrix does not obey the diffusion matrix argument we make for reordering as an augmentation. While masking a diffused matrix is logically allowed, we did not experiment on this case thoroughly and did not find any encouraging preliminary empirical results for this case.
\end{itemize}

We do not claim our order is the best of all possible permutations. Nevertheless, it can be seen the choice is not entirely ad hoc.

\subsection{Ablation tables} \label{sec:appendix_abl}

\textbf{The necessity of our spectral crop and reorder frequency components.} We first report in Table \ref{tbl:ablation} the results of ablations that involve replacing the proposed crop and reorder augmentations with random analogues. The results validate the necessity of following our eigenspectrum-designed approaches for cropping and reordering. We explore replacing the proposed crop with a random crop (randomly selecting a subgraph by excluding nodes). For reordering, we compare to a random permutation of spectral positional encoding. We observe a consistent drop in performance across all datasets when we replace either augmentation with its random counterpart. This indicates that our spectral augmentation designs can make a significant difference in terms of capturing more effective universal topological properties.

\begin{table}[h]
\centering
\caption{Ablation study for random versions of crop and permute vs. our spectral cropping and spectral positional encoding permute, E2E only. Bolding means best, asterisk for significance. Statistical analysis and standard deviations in Appendix C.}
\centering
{\resizebox{\textwidth}{!}{
\begin{tabular}{l|l|l|l|l|l|l|l}
\hline
Augment                    & US-Airport & H-index & IMDB-B & IMDB-M & COLLAB & RDT-B & RDT-M \\ \hline
GCC  & 64.8          & {78.3}  & 71.7          & 49.3          & 74.7          & 87.5          & 52.6          \\ \hline
SGCL - Random Permute  & 63.5       & 76.1    & 71.4   & 47.8   & 74.2   & 86.8  & 52.2  \\ \hline
SGCL - Random Crop     & 64.5       & 78.5    & 71.8   & 49.4   & 74.4   & 87.8  & 52.1  \\ \hline

SGCL  & \textbf{65.3*}           & \textbf{78.9} & \textbf{73.0*}    &\textbf{49.8}        & \textbf{75.6*}          & \textbf{88.4*}         & \textbf{53.8*}  \\\hline

\end{tabular}}}
\label{tbl:ablation}
\end{table}

In this section we now report the results for all other datasets when subjected to pairwise augmentations. We do want to re-iterate that since we have ten downstream datasets and five large-scale pre-train datasets, we select Reddit-binary as the downstream task and the smallest pre-train DBLP (SNAP) dataset as a demonstration. Using more pre-train datasets should result in further performance improvements, but due to computation time constraints, we focus on the simpler setting. We present these results (raw percentage gains over GCC) in tables \ref{tab:aug_abalation},\ref{tbl:rdt-m},\ref{tbl:imdb},\ref{tbl:hindex},\ref{tbl:usa},\ref{tbl:collab},\ref{tbl:imdb-m}. Statistically \textcolor{cadmiumgreen}{positive} and \textcolor{red}{negative} cases are marked accordingly.

\begin{table}
\centering
\caption{Pairwise effect of augmentations and post-processing methods, with E2E, frozen setting, on Reddit-binary, showing raw percentage gains over GCC~\citep{qiu2020gcc}. Further ablations appear in Appendix B. Marked are statistically significant \textcolor{cadmiumgreen}{positives} and \textcolor{red}{negatives}.}
\begin{tabular}{l|l|l|l|l|l|l}
\hline
Dataset & S-Crop & Mask & S-Reorder & Align & Similar & Diverse \\ \hline
S-Crop    & 0.22 & \textcolor{cadmiumgreen}{0.47} & 0.43 & \textcolor{cadmiumgreen}{0.54}  & \textcolor{cadmiumgreen}{0.71}    & -0.38   \\ \hline
Mask    &      & 0.15 & 0.26 & 0.35  & 0.42    & -0.15   \\ \hline
S-Reorder    &      &      & 0.18 & 0.19  & \textcolor{cadmiumgreen}{0.58}    & 0.45    \\ \hline
Align   &      &      &      & 0.16  & \textcolor{cadmiumgreen}{0.56}    & -0.26   \\ \hline
Similar &      &      &      &       & 0.31    & N/A     \\ \hline
Diverse &      &      &      &       &         & \textcolor{red}{-0.72}   \\ \hline
\end{tabular}
\label{tab:aug_abalation}

\end{table}

\begin{table}[h]
\centering
\begin{tabular}{|l|l|l|l|l|l|l|}
\hline
Dataset & Crop & Mask  & Reorder  & Align & Similar & Diverse \\ \hline
Crop    & \textcolor{cadmiumgreen}{0.62} & 0.58  & 0.55  & 0.43  & \textcolor{cadmiumgreen}{0.85}    & 0.32    \\ \hline
Mask    &      & -0.24 & -0.56 & \textcolor{red}{-1.04} & 0.52    & -0.17   \\ \hline
Reorder    &      &       & -0.14 & -0.49 & 0.38    & -0.26   \\ \hline
Align   &      &       &       & -0.47 & \textcolor{cadmiumgreen}{0.64}    & -0.56   \\ \hline
Similar &      &       &       &       & 0.45    & N/A     \\ \hline
Diverse &      &       &       &       &         & \textcolor{red}{-0.59}   \\ \hline
\end{tabular}

\caption{Ablation results on Reddit-5K}
\label{tbl:rdt-m}
\end{table}

\begin{table}[h]
\centering
\begin{tabular}{|l|l|l|l|l|l|l|}
\hline
Dataset & Crop & Mask  & Reorder & Align & Similar & Diverse \\ \hline
Crop    & \textcolor{cadmiumgreen}{0.83} & \textcolor{cadmiumgreen}{0.92}  & \textcolor{cadmiumgreen}{1.16}    & \textcolor{cadmiumgreen}{0.98}  & \textcolor{cadmiumgreen}{0.87}    & 0.62    \\ \hline
Mask    &      & -0.32 & 0.58    & 0.22  & -0.35   & \textcolor{red}{-0.78}   \\ \hline
Reorder &      &       & \textcolor{cadmiumgreen}{0.75}    & \textcolor{cadmiumgreen}{0.82}  & 0.56    & 0.43    \\ \hline
Align   &      &       &         & 0.18  & -0.19   & -0.22   \\ \hline
Similar &      &       &         &       & -0.08   & N/A     \\ \hline
Diverse &      &       &         &       &         & -0.15   \\ \hline
\end{tabular}
\caption{Ablation results on IMDB-Binary}
\label{tbl:imdb}
\end{table}

\begin{table}[h]
\centering
\begin{tabular}{|l|l|l|l|l|l|l|}
\hline
Dataset & Crop & Mask & Reorder & Align & Similar & Diverse \\ \hline
Crop    & 0.24 & 0.42 & \textcolor{cadmiumgreen}{0.56}    & 0.47  & 0.38    & 0.31    \\ \hline
Mask    &      & 0.31 & \textcolor{cadmiumgreen}{0.52}    & 0.43  & 0.25    & 0.38    \\ \hline
Reorder &      &      & 0.48    & 0.38  & 0.21    & 0.34    \\ \hline
Align   &      &      &         & 0.28  & 0.20    & 0.36    \\ \hline
Similar &      &      &         &       & -0.11   & N/A     \\ \hline
Diverse &      &      &         &       &         & 0.06    \\ \hline
\end{tabular}
\caption{Ablation results on h-index dataset}
\label{tbl:hindex}
\end{table}

\begin{table}[h]
\centering
\begin{tabular}{|l|l|l|l|l|l|l|}
\hline
Dataset & Crop  & Mask & Reorder & Align & Similar & Diverse \\ \hline
Crop    & -0.08 & 0.27 & -0.18   & -0.43 & 0.28    & 0.22    \\ \hline
Mask    &       & 0.31 & 0.26    & 0.16  & 0.43    & 0.38    \\ \hline
Reorder &       &      & -0.15   & -0.23 & -0.07   & 0.03    \\ \hline
Align   &       &      &         & -0.12 & 0.14    & 0.05    \\ \hline
Similar &       &      &         &       & 0.25    & N/A     \\ \hline
Diverse &       &      &         &       &         & 0.14    \\ \hline
\end{tabular}
\caption{Ablation results on US-Airport dataset}
\label{tbl:usa}
\end{table}

\begin{table}[h]
\centering
\begin{tabular}{|l|l|l|l|l|l|l|}
\hline
Dataset & Crop & Mask & Reorder & Align & Similar & Diverse \\ \hline
Crop    & 0.23 & 0.42 & 0.32    & 0.34  & \textcolor{cadmiumgreen}{0.71}    & 0.54    \\ \hline
Mask    &      & 0.18 & 0.45    & 0.32  & \textcolor{cadmiumgreen}{0.62}    & 0.42    \\ \hline
Reorder &      &      & 0.22    & 0.45  & 0.55    & \textcolor{cadmiumgreen}{0.59}    \\ \hline
Align   &      &      &         & 0.15  & \textcolor{cadmiumgreen}{0.68}    & 0.52    \\ \hline
Similar &      &      &         &       & 0.48    & N/A     \\ \hline
Diverse &      &      &         &       &         & 0.34    \\ \hline
\end{tabular}
\caption{Ablation results on COLLAB dataset}
\label{tbl:collab}
\end{table}

\begin{table}[h]
\centering
\begin{tabular}{|l|l|l|l|l|l|l|}
\hline
Dataset & Crop & Mask  & Reorder & Align & Similar & Diverse \\ \hline
Crop    & 0.35 & 0.25  & 0.48    & 0.21  & \textcolor{cadmiumgreen}{0.57}    & \textcolor{cadmiumgreen}{0.66}    \\ \hline
Mask    &      & -0.07 & 0.22    & -0.19 & 0.22    & 0.47    \\ \hline
Reorder &      &       & 0.28    & 0.17  & 0.38    & \textcolor{cadmiumgreen}{0.82}    \\ \hline
Align   &      &       &         & -0.11 & 0.16    & 0.54    \\ \hline
Similar &      &       &         &       & 0.19    & N/A     \\ \hline
Diverse &      &       &         &       &         & \textcolor{cadmiumgreen}{0.62}    \\ \hline
\end{tabular}
\caption{Ablation results on IMDB-Multi dataset}
\label{tbl:imdb-m}
\end{table}

\clearpage

\subsection{Additional results with MoCo and Fine-tuning} \label{sec:appendix_full_train}

Pre-trained encoders fine-tuned using ADAM with $3$ epochs warmup and $3$ epochs ramp-down with a learning rate of $0.005$ are used for the fine-tuned case. These results appear in tables \ref{tab:nl_fine_tune}, \ref{tab:gl_fine_tune}  and \ref{tbl:ssfinetune_ci}. We present results for E2E and MoCo~\citep{he2020momentum_moco} in both the frozen and fine-tuned setting.

\begin{table}[h]
    \caption{Node classification. Results obtained by pretraining along with fine-tuning on the downstream dataset labels for both E2E and MoCo, with frozen results also re-provided from the main paper. Statistical details are discussed in Appendix C. The methods that appear above the ``Frozen" category are compared relative to frozen methods. They require no fine-tuning and are more similar to frozen methods, however they are distinct in that they are not pre-training heavy but rather extract the structure directly like Laplacian methods, making them distinct but strong baselines with the least requirement in terms of pre-training or data-specific work. Thus, there forms a continuum from these methods to fully fine-tuned methods with frozen methods lying in an intermediate position.}
    \label{tab:nl_fine_tune}

    \centering
\begin{tabular}{@{}c|cc|cc@{}}
\toprule
 & \multicolumn{2}{c|}{US-Airport} & \multicolumn{2}{c}{H-index} \\ \midrule
$|V|$ & \multicolumn{2}{c|}{1,190} & \multicolumn{2}{c}{5,000} \\
$|E|$ & \multicolumn{2}{c|}{13,599} & \multicolumn{2}{c}{44,020} \\\midrule
\multicolumn{5}{c}{Frozen-like methods}  \\ \hline
ProNE & \multicolumn{2}{c|}{62.3} & \multicolumn{2}{c}{69.1} \\
GraphWave & \multicolumn{2}{c|}{60.2} & \multicolumn{2}{c}{70.3} \\
Struc2vec & \multicolumn{2}{c|}{\textbf{66.2}} & \multicolumn{2}{c}{$\geq 1$ Day} \\ \midrule
\multicolumn{5}{c}{Frozen}  \\ \hline
Training mode & MoCo & E2E & MoCo & E2E \\\midrule 
GCC & 65.6 & 64.8 & 75.2 & 78.3 \\ \hline 
GraphCL & $62.8  $  &  $63.5  $   &$74.3  $ & $76.5  $ \\
GRACE & $62.6 $  &  $63.3  $   &$74.5  $ & $77.0  $ \\
MVGRL &  $65.2  $  &  $64.5  $   & $75.1  $ & $78.1 $ \\  
CuCo &  $64.9  $  &  $64.3 $   & $75.3  $ & $78.2  $ \\
BYOV &  ${65.3}  $ &   $64.7 $  & $76.0  $   & ${78.1} $ \\
InfoGCL &  ${63.2}  $ &  $ 64.1 $  & $75.4 $   & ${77.6} $ \\
GCA &  ${64.5}  $ &   $64.3 $  & $ 75.8  $   & ${78.0} $ \\
SGCL & ${65.9}  $ &   $65.3 $  & $\textbf{76.7}  $   & $\textbf{78.9*}$ \\ \bottomrule
\midrule
\multicolumn{5}{c}{Fine-tuned}  \\ \hline
Training mode & MoCo & E2E & MoCo & E2E \\\midrule 
GCC & 67.2 & 68.3 & 80.6 &  80.5 \\ \hline 
GraphCL & $64.3  $  &  $66.4  $   &$79.0  $ & $78.8  $ \\
GRACE & $64.0 $  &  $65.9  $   &$78.2  $ & $78.5  $ \\
MVGRL &  $66.5  $  &  $67.9  $   & $79.6  $ & $79.9 $ \\  
CuCo &  $66.1  $  &  $67.7 $   & $79.4  $ & $80.1  $ \\
BYOV &  ${67.0}  $ &   $67.8 $  & $80.3  $   & $ 80.2 $ \\
InfoGCL &  ${65.5}  $ &   $67.2 $  & $ 79.8  $   & $ 79.5 $ \\
GCA &  $ 66.8  $ &   $67.6 $  & $80.0  $   & $ 79.9 $ \\
SGCL & $\textbf{67.5}  $ &   $\textbf{68.6} $  & $\textbf{80.8}  $   & $\textbf{80.7}$ \\ \bottomrule
\end{tabular}
    \normalsize

\end{table}

\begin{table}[h]
\centering
\caption{Graph classification results when the pre-trained graph encoder transfers to an out-of-domain graph or is fine tuned. "-" indicates the model is unable to produce reasonable results given 24 hours of training time, as explained in~\citep{qiu2020gcc}.  Bolding indicates best result, asterisk indicates statistical significance. Standard deviations, confidence intervals etc. in Appendix C. The methods that appear above the ``Frozen" category are compared relative to frozen methods. They require no fine-tuning and are more similar to frozen methods, however they are distinct in that they are not pre-training heavy but rather extract the structure directly like Laplacian methods, making them distinct but strong baselines with the least requirement in terms of pre-training or data-specific work. Thus, there forms a continuum from these methods to fully fine-tuned methods with frozen methods lying in an intermediate position.} \label{tab:gl_fine_tune}
\resizebox{\columnwidth}{!}{
\begin{tabular}{@{}c|cc|cc|cc|cc|cc@{}}
\toprule
Datasets & \multicolumn{2}{c|}{IMDB-B} & \multicolumn{2}{c|}{IMDB-M} & \multicolumn{2}{c|}{COLLAB} & \multicolumn{2}{c|}{RDT-B} & \multicolumn{2}{c}{RDT-M} \\ \midrule
\# graphs & \multicolumn{2}{c|}{1,000} & \multicolumn{2}{c|}{1,500} & \multicolumn{2}{c|}{5,000} & \multicolumn{2}{c|}{2,000} & \multicolumn{2}{c}{5,000} \\
\# classes & \multicolumn{2}{c|}{2} & \multicolumn{2}{c|}{3} & \multicolumn{2}{c|}{3} & \multicolumn{2}{c|}{2} & \multicolumn{2}{c}{5} \\
Avg. \# nodes & \multicolumn{2}{c|}{19.8} & \multicolumn{2}{c|}{13.0} & \multicolumn{2}{c|}{74.5} & \multicolumn{2}{c|}{429.6} & \multicolumn{2}{c}{508.5} \\ \midrule
\multicolumn{11}{c}{Frozen-like methods}  \\ \hline
DGK & \multicolumn{2}{c|}{67.0} & \multicolumn{2}{c|}{44.6} & \multicolumn{2}{c|}{73.1} & \multicolumn{2}{c|}{78.0} & \multicolumn{2}{c}{41.3} \\
graph2vec & \multicolumn{2}{c|}{71.1} & \multicolumn{2}{c|}{\textbf{50.4}} & \multicolumn{2}{c|}{--} & \multicolumn{2}{c|}{75.8} & \multicolumn{2}{c}{47.9} \\
InfoGraph & \multicolumn{2}{c|}{73.0} & \multicolumn{2}{c|}{49.7} & \multicolumn{2}{c|}{--} & \multicolumn{2}{c|}{82.5} & \multicolumn{2}{c}{53.5} \\ \midrule
\multicolumn{11}{c}{Frozen}  \\ \hline
Training mode & MoCo & E2E & MoCo & E2E & MoCo & E2E & MoCo & E2E & MoCo & E2E \\
GCC & 72.0 & 71.7 & 49.4 & 49.3 & 78.9 & 74.7 & 89.8 & 87.5 & 53.7 & 52.6 \\
GraphCL & $72.2 $  & $70.9 $ &  $49.3  $  & $ 47.9 $  & $77.2$  & $ 74.1 $  & $88.7 $ &  $87.2 $    & $52.9 $ & $51.8 $  \\
GRACE & $71.7 $  & $71.5 $ &  $49.2  $  & $ 48.8 $  & $78.3 $  & $ 74.5 $  & $89.2 $ &  $87.0 $    & $53.4 $ & $52.0 $  \\
CuCo & 71.8 & 71.3  & 48.7  & 48.5  & 78.5  & 74.2  & 89.3  & 87.8  & 52.5  & 51.6  \\
BYOV & 72.3 & 72.0  & 48.5  & 49.2  & 78.4  & 75.1  & 89.5  & 87.9  & 53.6  & 53.0  \\

MVGRL & 72.3  & 72.2  & 49.2  & 49.4  & 78.6  & 75.0  & 89.6  & 87.4  & 53.4  & 52.8  \\ 

InfoGCL & 72.0  & 71.0  & 48.8  & 48.2  & 77.8  & 74.6  & 89.1  & 87.3  & 52.7  & 52.2  \\ 

GCA & 72.2  & 71.9  & 49.0  & 48.7  & 78.4  & 74.4  & 88.9  & 87.5  & 53.2  & 52.4  \\ 

SGCL & $\textbf{73.4*}$ & $\textbf{73.0} $ & $50.0  $  & $ 49.8 $  & $\textbf{79.7*} $ & $ \textbf{75.6}$   & $\textbf{90.6*} $ & $\textbf{88.4}$  & $\textbf{54.2*} $ & $\textbf{53.8} $  \\\bottomrule
\multicolumn{11}{c}{Fine-tuned}  \\ \hline
Training mode & MoCo & E2E & MoCo & E2E & MoCo & E2E & MoCo & E2E & MoCo & E2E \\
DGCNN & \multicolumn{2}{c|}{70.0} & \multicolumn{2}{c|}{47.8} & \multicolumn{2}{c|}{73.7} & \multicolumn{2}{c|}{-} & \multicolumn{2}{c}{-}  \\ 
GIN & \multicolumn{2}{c|}{\textbf{75.6*}} & \multicolumn{2}{c|}{\textbf{51.5*}} & \multicolumn{2}{c|}{\textbf{80.2}} & \multicolumn{2}{c|}{\textbf{89.4*}} & \multicolumn{2}{c}{\textbf{54.5}}  \\ 
GCC(Random) & \multicolumn{2}{c|}{75.6} & \multicolumn{2}{c|}{50.9} & \multicolumn{2}{c|}{79.4} & \multicolumn{2}{c|}{87.8} & \multicolumn{2}{c}{52.1}  \\ \hline
GCC & 73.8 & 70.8 & 50.3 & 48.5 & 81.1 & 79.0 & 87.6 & 86.4 & 53.0 & 47.4 \\
GraphCL & $73.5 $  & $71.1 $ &  $49.8  $  & $ 47.9 $  & $80.6$  & $ 78.6 $  & $87.1 $ &  $86.7 $    & $51.9 $ & $48.7 $  \\
GRACE & $73.0 $  & $71.3 $ &  $49.4  $  & $ 47.4 $  & $79.5 $  & $ 77.6 $  & $86.5 $ &  $86.7 $    & $51.5 $ & $48.3 $  \\
CuCo & 72.6 & 71.2  & 49.2  & 46.9  & 78.1  & 77.0  & 86.8  & 86.5  & 51.3  & 48.3  \\
BYOV & 73.5 & 72.4  & 50.1  & 49.6  & 81.2  & 79.3  & 88.2  & 87.0  & 53.9  & 50.2  \\

MVGRL & 72.3  & 72.2  & 49.2  & 49.4  & 78.6  & 77.3  & 87.9  & 86.8  & 53.4  & 49.8  \\ 

InfoGCL & 73.6  & 71.5  & 50.0  & 48.4  & 80.2  & 78.8  & 87.5  & 86.3  & 52.4  & 49.2  \\

GCA & 73.1  & 71.7  & 49.5 & 47.9  & 79.8  & 78.2  & 88.0  & 86.5  & 52.1  & 48.6  \\

SGCL & ${74.2}$ & ${72.8} $ & ${50.6}  $  & $ 50.1 $  & $\textbf{81.5*} $ & $ {79.8}$   & ${88.5} $ & ${87.4}$  & ${54.4} $ & ${50.8} $  \\\bottomrule
\end{tabular}}
\end{table}

\begin{table}[h]

    \caption{Top-$k$ similarity search~($k=20, 40$), frozen cases only with $4$ structural methods (Random, RolX, Panther, GraphWave) that are also similar to frozen methods in runtime requirements (see previous tables).  Bolding indicates best result, asterisk indicates statistical significance. Standard deviations, confidence intervals etc. in Appendix C.}
    \centering
{%
    \begin{tabular}{l | r r |r r| r r@{~}}
        \toprule
                            & \multicolumn{2}{c|}{KDD-ICDM} & \multicolumn{2}{c|}{SIGIR-CIKM} & \multicolumn{2}{c}{SIGMOD-ICDE}                                                       \\\midrule
        $|V|$               & 2,867                         & 2,607                           & 2,851                           & 3,548           & 2,616           & 2,559           \\
        $|E|$               & 7,637                         & 4,774                           & 6,354                           & 7,076           & 8,304           & 6,668           \\
        \#~ground truth      & \multicolumn{2}{r|}{697}      & \multicolumn{2}{r|}{874}        & \multicolumn{2}{r}{898}                                                               \\\midrule
        $k$                 & 20                            & 40                              & 20                              & 40              & 20              & 40              \\            \midrule
        \midrule
        \multicolumn{7}{c}{Frozen-like methods} \\ \hline
            Random              & 0.0198                        & 0.0566                          & 0.0223                          & 0.0447          & 0.0221          & 0.0521          \\
        RolX                & 0.0779                        & 0.1288                          & 0.0548                          & 0.0984          & 0.0776          & 0.1309          \\
        Panther++           & 0.0892                        & 0.1558                          & \textbf{0.0782}                 & 0.1185          & 0.0921          & 0.1320          \\
        GraphWave           & 0.0846                        & \textbf{0.1693}                 & 0.0549                          & 0.0995          & {0.0947} & {0.1470} \\\midrule
    \midrule
        \multicolumn{7}{c}{E2E} \\ \hline
          GCC  & 0.1047               & 0.1564                          & 0.0549                          & 0.1247 & 0.0835          & 0.1336          \\
          \hline

        GraphCL  &  0.0986         & 0.1574                           &          0.0583                & {0.1209}   & 0.0796          & {0.1205}         \\

        GRACE  &  0.1021                            &     0.1558                 & 0.0568  & 0.1226          & 0.0864 & 0.1262      \\

        MVGRL   &  0.0982                             &     0.1483                 & 0.0514  &      0.1174    & 0.0774 & 0.1159      \\

        CuCo   &    0.1063                          &   0.1543                   & 0.0568  &    0.1274      & 0.0924 & 0.1374       \\

        BYOV   &    0.1068                          &   0.1585                   & 0.0592  & 0.1268          & 0.0824 & 0.1318       \\

        InfoGCL   &    0.0972                          &   0.1550                   & 0.0595  & 0.1217          & 0.0802 & 0.1237  \\
        GCA   &    0.1007                          &   0.1563                   & 0.0559  & 0.1197          & 0.0849 & 0.1244  \\

        \midrule
SGCL  &       \textbf{0.1105*}         & 0.1642                           &          0.0658               & \textbf{0.1363*}   & \textbf{0.1076*}          & \textbf{0.1561*}       \\

\midrule
        \multicolumn{7}{c}{MoCo} \\ \hline
        GCC  & 0.0904             & 0.1521                          & 0.0652                          & 0.1178 & 0.0846         &  0.1425         \\

        \midrule
  GraphCL  &       0.0835         & 0.1507                         &          0.0629              & 0.1165   &  0.0872         &  0.1434     \\

        GRACE  &  0.0852                            &     0.1516                 & 0.0616  & 0.1172          & 0.0917 & 0.1469      \\

        MVGRL   &  0.0826                             &     0.1458                 & 0.0559  &      0.1116    & 0.0851 & 0.1387      \\

        CuCo   &    0.0864                          &   0.1512                   & 0.0624  &    0.1216      & 0.0877 & 0.1414      \\

        BYOV   &    0.0926                          &   0.1553                   & 0.0642  & 0.1228          & 0.0859 & 0.1468
        \\
        InfoGCL   &    0.0848                          &   0.1536                   & 0.0619  & 0.1183          & 0.0884 & 0.1425  \\
        GCA   &    0.0843                          &   0.1507                   & 0.0607  & 0.1192          & 0.0865 & 0.1426  \\

        \midrule
SGCL  &       \textbf{0.0978*}          & 0.1627                            &          0.0765            &  \textbf{0.1306*}   & \textbf{0.1049*}         & \textbf{0.1583*}      \\

        \bottomrule
    \end{tabular}}
        \label{tab:similarity}

    \label{tbl:ssfinetune_ci}

\end{table}

\clearpage

\subsection{Datasets and benchmark code}

We obtain the datasets from the following sources :

\begin{itemize}
    \item \href{https://github.com/leoribeiro/struc2vec/tree/master/graph}{https://github.com/leoribeiro/struc2vec/tree/master/graph}
    \item \href{https://www.openacademic.ai/oag/}{https://www.openacademic.ai/oag/}
    \item \href{https://ls11-www.cs.tu-dortmund.de/staff/morris/graphkerneldatasets} {https://ls11-www.cs.tu-dortmund.de/staff/morris/graphkerneldatasets}
\end{itemize}

And the relevant benchmarks from :

\begin{itemize}
    \item GCC : \href{https://github.com/THUDM/GCC}{https://github.com/THUDM/GCC}
    \item GraphCL : \href{https://github.com/Shen-Lab/GraphCL}{https://github.com/Shen-Lab/GraphCL}
    \item MVGRL : \href{https://github.com/kavehhassani/mvgrl}{https://github.com/kavehhassani/mvgrl}
    \item BYOV : \href{https://github.com/Shen-Lab/GraphCL_Automated}{https://github.com/Shen-Lab/GraphCL\_Automated}
    \item CuCo : \href{https://github.com/BUPT-GAMMA/CuCo}{https://github.com/BUPT-GAMMA/CuCo}
    \item GRACE : \href{https://github.com/CRIPAC-DIG/GRACE}{https://github.com/CRIPAC-DIG/GRACE}
\end{itemize}

\clearpage

\section{Hardware details and statistical confidence intervals of  results}

\textbf{Hardware and software}: We tested all code on Python $3.7$ with PyTorch $1.3.1$, CUDA $10.1$, scikit-learn $0.20.3$. The Tesla $V100$ (one per model per run) served as the GPU.

We compute statistical confidence bounds only for the methods whose results we do not copy over from the GCC paper.

\begin{table}[h]
    \caption{Node classification. Results indicate the upper confidence bound (95 percentile) and the lower (5th percentile) and the standard deviation in brackets.}
    \label{tbl:nodefinetune_ci}

    \centering
\begin{tabular}{@{}c|cc|cc@{}}
\toprule
 & \multicolumn{2}{c|}{US-Airport} & \multicolumn{2}{c}{H-index} \\ \midrule
$|V|$ & \multicolumn{2}{c|}{1,190} & \multicolumn{2}{c}{5,000} \\
$|E|$ & \multicolumn{2}{c|}{13,599} & \multicolumn{2}{c}{44,020} \\\midrule

Training mode & MoCo & E2E & MoCo & E2E \\\midrule 

GraphCL & $\substack{63.7 \\ 62.2} (0.4)  $  &  $\substack{64.2 \\ 62.8} (0.3)  $   &$\substack{73.9 \\ 75.2} (0.4) $ & $\substack{77.8 \\ 75.4} (0.9)  $ \\
GRACE & $\substack{63.5 \\ 61.9} (0.4)  $  &  $\substack{63.9 \\ 62.7} (0.3)  $   &$\substack{74.1 \\ 74.9} (0.3) $ & $\substack{77.9 \\ 76.0} (0.6)  $ \\
MVGRL &  $\substack{65.5 \\ 64.9} (0.2) $  &  $\substack{64.7 \\ 64.3} (0.1) $   & $ \substack{75.5 \\ 74.8} (0.2) $ & $\substack{78.4 \\ 77.7} (0.2) $ \\  
CuCo &  $\substack{65.2 \\ 64.5} (0.2) $  &  $\substack{64.6 \\ 63.9} (0.2) $   & $ \substack{75.9 \\ 74.8} (0.3) $ & $\substack{78.5 \\ 77.8} (0.2)  $ \\
BYOV &  $\substack{65.5 \\ 64.9} (0.2)  $ &   $\substack{65.2 \\ 64.0} (0.4)$  & $\substack{76.6 \\ 75.5} (0.3)  $   & $\substack{78.5 \\ 77.6} (0.3)$ \\
InfoGCL &  $\substack{63.6 \\ 62.7} (0.3)  $ &   $\substack{64.4 \\ 63.7} (0.2)  $  &  $\substack{75.8 \\ 74.9} (0.3)  $   &  $\substack{78.0 \\ 77.1} (0.3)  $ \\
GCA &  $\substack{64.9 \\ 64.0} (0.3)  $ &   $\substack{64.6 \\ 64.0} (0.2)$  & $\substack{76.2 \\ 75.4} (0.3)  $   & $\substack{78.3 \\ 77.6} (0.2)$ \\
SGCL & $\substack{66.2 \\ 65.7} (0.2) $ &   $\substack{65.6 \\ 65.1} (0.2) $  & $\substack{77.1 \\ 76.3} (0.3)  $   & $\substack{79.1 \\ 78.7}$ (0.2) \\ \bottomrule
\midrule
\multicolumn{5}{c}{Fine-tuned}  \\ \hline
Training mode & MoCo & E2E & MoCo & E2E \\\midrule 

GraphCL & $\substack{66.6 \\ 62.9} (1.1)  $  &  $\substack{67.5 \\ 65.6} (0.7)  $   &$\substack{78.2 \\ 79.8} (0.5) $ & $\substack{79.5 \\ 77.9} (0.5) $ \\
GRACE & $\substack{65.1 \\ 62.8} (0.8) $  &  $\substack{66.2 \\ 65.5} (0.3) $   &$\substack{78.6 \\ 77.5} (0.4) $ & $\substack{78.8 \\ 78.1} (0.2)$ \\
MVGRL &  $\substack{66.8 \\ 66.1} (0.2) $  &  $\substack{68.1 \\ 67.6} (0.2) $   & $ \substack{79.7 \\ 79.4} (0.1)$ & $\substack{80.2 \\ 79.5} (0.2)$ \\  
CuCo &  $ \substack{66.5 \\ 65.6} (0.3)  $  &  $  \substack{68.0 \\ 67.3} (0.3) $   & $ \substack{79.7 \\ 79.0} (0.2)  $ & $ \substack{80.2 \\ 79.8} (0.1)  $ \\
BYOV &  $ \substack{67.4 \\ 66.5} (0.3)  $ &   $ \substack{68.2 \\ 67.3} (0.3) $  & $ \substack{80.5 \\ 80.0} (0.2) $   & $  \substack{80.4 \\ 79.9} (0.2) $ \\
InfoGCL &  $\substack{65.9 \\ 65.0} (0.3)  $ &   $\substack{67.6 \\ 66.7} (0.3)  $  &  $\substack{79.5 \\ 80.0} (0.2)  $   &  $\substack{79.9 \\ 79.0} (0.3)  $ \\
GCA &  $\substack{67.2 \\ 66.3} (0.3)  $ &   $\substack{67.9 \\ 67.3} (0.2)$  & $\substack{80.2 \\ 79.7} (0.2)  $   & $\substack{80.2 \\ 79.5} (0.3)$ \\
SGCL & $\substack{67.6 \\ 67.4} (0.1) $ &   $\substack{68.8 \\ 68.5} (0.1) $  & $\substack{81.0 \\ 80.7} (0.1) $   & $\substack{80.8 \\ 80.6} (0.1)$ \\ \bottomrule
\end{tabular}
    \normalsize

\end{table}

\begin{table}[h]
\centering
\caption{Graph classification confidence bounds. Results indicate the upper confidence bound (95 percentile) and the lower (5th percentile) and the standard deviation in brackets. } \label{tbl:graphfinetune_ci}
\resizebox{\textwidth}{!}{
\begin{tabular}{@{}c|cc|cc|cc|cc|cc@{}}
\toprule
Datasets & \multicolumn{2}{c|}{IMDB-B} & \multicolumn{2}{c|}{IMDB-M} & \multicolumn{2}{c|}{COLLAB} & \multicolumn{2}{c|}{RDT-B} & \multicolumn{2}{c}{RDT-M} \\ \midrule
\# graphs & \multicolumn{2}{c|}{1,000} & \multicolumn{2}{c|}{1,500} & \multicolumn{2}{c|}{5,000} & \multicolumn{2}{c|}{2,000} & \multicolumn{2}{c}{5,000} \\
\# classes & \multicolumn{2}{c|}{2} & \multicolumn{2}{c|}{3} & \multicolumn{2}{c|}{3} & \multicolumn{2}{c|}{2} & \multicolumn{2}{c}{5} \\
Avg. \# nodes & \multicolumn{2}{c|}{19.8} & \multicolumn{2}{c|}{13.0} & \multicolumn{2}{c|}{74.5} & \multicolumn{2}{c|}{429.6} & \multicolumn{2}{c}{508.5} \\ \midrule
\multicolumn{11}{c}{Frozen}  \\ \hline
Training mode & MoCo & E2E & MoCo & E2E & MoCo & E2E & MoCo & E2E & MoCo & E2E \\

GraphCL & $\substack{72.8 \\ 71.5} (0.5) $  & $ \substack{72.1 \\ 69.5} (0.9)$ &  $ \substack{50.1 \\ 48.5} (0.2)  $  & $\substack{48.4 \\ 47.2} (0.4) $  & $\substack{77.5 \\ 76.8}$ (0.3) & $ \substack{74.7 \\ 73.5} (0.4) $  & $\substack{89.3 \\ 88.0} (0.5) $ &  $\substack{87.7 \\ 86.3} (0.5) $    & $\substack{53.5 \\ 52.1} (0.5) $ & $\substack{52.8 \\ 50.7} (0.7) $  \\
GRACE & $ \substack{72.2 \\ 71.1} (0.4)$  & $ \substack{71.9 \\ 71.0} (0.3)$ &  $ \substack{49.7 \\ 48.8} (0.3)$  & $ \substack{49.4 \\ 48.1} (0.5)$  & $ \substack{78.5 \\ 78.0} (0.2)$  & $   \substack{75.0 \\ 74.1} (0.3) $  & $\substack{89.5 \\ 89.0} (0.2) $ &  $\substack{87.4 \\ 86.5} (0.3)$    & $\substack{53.8 \\ 52.9} (0.3) $ & $\substack{52.7 \\ 51.2} (0.5) $  \\
CuCo & $\substack{72.3 \\ 71.1} (0.4)$ &  $\substack{71.9 \\ 70.7} (0.4)$  &   $\substack{49.2 \\ 48.3} (0.3)$& $\substack{48.9 \\ 48.0} (0.3)$ &  $\substack{78.9 \\ 78.0} (0.3)$  & $\substack{74.6 \\ 73.7} (0.3)$  &  $\substack{89.6 \\ 88.9} (0.2)$ & $\substack{88.1 \\ 87.6} (0.2)$  & $\substack{53.1 \\ 52.0} (0.4)$  & $\substack{52.1 \\ 51.0} (0.4)$  \\
BYOV & $\substack{72.6 \\ 72.0} (0.2)$ &  $\substack{72.3 \\ 71.6} (0.3)$ &  $\substack{49.1 \\ 48.0} (0.4)$ &  $\substack{49.6 \\ 48.6}(0.4)$  &  $\substack{79.1 \\ 77.7} (0.5)$ & $\substack{75.3 \\ 74.7} (0.2)$ &  $\substack{89.9 \\ 89.2} (0.2)$ &  $\substack{88.2 \\ 87.4} (0.2)$  & $\substack{53.9 \\ 53.2} (0.2)$  &  $\substack{53.3 \\ 52.6} (0.2)$ \\

MVGRL &$\substack{72.7 \\ 72.1} (0.2)$  &  $\substack{72.6 \\ 71.8} (0.3)$  &  $\substack{49.5 \\ 49.0} (0.2)$ &  $\substack{49.8 \\ 48.9} (0.3)$ & $\substack{78.8 \\ 78.3} (0.2)$  & $\substack{75.4 \\ 74.6} (0.3)$  & $\substack{89.9 \\ 89.2} (0.2)$  & $\substack{87.7 \\ 87.0} (0.2)$  & $\substack{53.6 \\ 53.1} (0.2)$  & $\substack{53.1 \\ 52.5} (0.2)$  \\

InfoGCL & $\substack{72.4 \\ 71.8} (0.2)$  &  $\substack{71.3 \\ 70.7} (0.2)$  &  $\substack{49.1 \\ 48.4} (0.2)$ &  $\substack{48.5 \\ 47.8} (0.2)$ & $\substack{78.1 \\ 77.5} (0.2)$  & $\substack{75.1 \\ 74.2} (0.3)$  & $\substack{89.5 \\ 88.7} (0.3)$  & $\substack{87.6 \\ 86.8} (0.3)$  & $\substack{53.0 \\ 52.3} (0.2)$  & $\substack{52.6 \\ 51.9} (0.2)$  \\

GCA &$\substack{72.6 \\ 72.0} (0.2)$  &  $\substack{72.2 \\ 71.6} (0.2)$  &  $\substack{49.3 \\ 48.7} (0.2)$ &  $\substack{49.0 \\ 48.3} (0.2)$ & $\substack{78.6 \\ 78.1} (0.2)$  & $\substack{74.7 \\ 74.1} (0.2)$  & $\substack{89.2 \\ 88.5} (0.3)$  & $\substack{87.7 \\ 87.2} (0.2)$  & $\substack{53.4 \\ 52.9} (0.2)$  & $\substack{52.7 \\ 52.0} (0.3)$  \\

SGCL & $\substack{73.8 \\ 73.1} (0.3)$ & $\substack{73.2 \\ 72.9} (0.1) $ & $\substack{50.2 \\ 49.9}  (0.1) $  & $ \substack{50.0 \\ 49.6} (0.2) $  & $\substack{79.9 \\ 79.6} (0.1) $ &  $\substack{75.9 \\ 75.4} (0.2)$  & $\substack{90.8 \\ 90.5} (0.1) $ & $\substack{88.7 \\ 88.2} (0.2)$  & $\substack{54.4 \\ 54.1} (0.1)$ & $\substack{54.1 \\ 53.5} (0.2)$  \\\bottomrule
\multicolumn{11}{c}{Fine-tuned}  \\ \hline
Training mode & MoCo & E2E & MoCo & E2E & MoCo & E2E & MoCo & E2E & MoCo & E2E \\

GraphCL & $\substack{74.1 \\ 73.0} (0.4) $  & $ \substack{71.6 \\ 70.7} (0.3) $ &  $ \substack{50.4 \\ 49.5} (0.3) $  & $\substack{48.8 \\ 47.1} (0.5)$  & $\substack{81.1 \\ 80.2} (0.3)$  & $ \substack{79.3 \\ 78.0} (0.4)$  & $\substack{87.7 \\ 86.4} (0.5) $ &  $\substack{87.3 \\ 86.4} (0.3) $    & $\substack{52.7 \\ 51.0}(0.5) $ & $\substack{49.5 \\ 47.8} (0.5) $  \\
GRACE & $\substack{73.6 \\ 72.5} (0.4)$  & $\substack{71.7 \\ 71.0} (0.2)$ &  $ \substack{49.8 \\ 49.1} (0.2) $  & $  \substack{47.7 \\ 47.2} (0.2)$  & $ \substack{79.8 \\ 79.1} (0.2)$  & $ \substack{77.9 \\ 77.2} (0.2)$  & $ \substack{86.9 \\ 86.0} (0.3)$ &  $ \substack{87.1 \\ 86.0} (0.4)$    & $ \substack{51.8 \\ 51.0} (0.3)$ & $ \substack{48.7 \\ 47.9} (0.3)$  \\
CuCo & $ \substack{73.1 \\ 72.2} (0.3)$ & $\substack{71.4 \\ 70.9} (0.2)$  & $\substack{49.5 \\ 48.8} (0.2)$ & $\substack{47.7 \\ 46.0} (0.5)$ &  $\substack{78.4 \\ 77.7} (0.2)$& $\substack{77.8 \\ 76.0} (0.6)$ & $\substack{87.5 \\ 86.1} (0.5)$ & $\substack{87.0 \\ 85.9} (0.4)$ & $\substack{51.6 \\ 50.9} (0.2)$ & $\substack{48.9 \\ 47.8} (0.4)$ \\
BYOV & $\substack{73.8 \\ 73.0} (0.3)$ &  $\substack{72.6 \\ 72.1} (0.2)$  &   $\substack{50.4 \\ 49.5} (0.3)$  &  $\substack{49.9 \\ 49.2} (0.2)$  &  $\substack{81.4 \\ 80.9} (0.2)$  &  $\substack{79.7 \\ 79.0} (0.2)$ &  $\substack{88.6 \\ 87.9} (0.2)$  & $\substack{87.4 \\ 86.5} (0.3)$  &  $\substack{54.2 \\ 53.5} (0.2)$  &  $\substack{50.5 \\ 49.8} (0.2)$  \\

MVGRL &$\substack{72.9 \\ 71.8} (0.4)$ & $\substack{72.6 \\ 71.5} (0.4)$  &  $\substack{49.6 \\ 48.5} (0.4)$  &  $\substack{49.9 \\ 48.8} (0.4)$  & $\substack{78.9 \\ 78.2} (0.2)$  & $\substack{77.8 \\ 76.9} (0.3)$  &$\substack{88.2 \\ 87.4} (0.3)$  & $\substack{87.2 \\ 86.3} (0.3)$  & $\substack{53.7 \\ 53.0} (0.2)$  & $\substack{50.2 \\ 49.3} (0.3)$ \\ 

InfoGCL &$\substack{73.8 \\ 73.5} (0.1)$ & $\substack{71.9 \\ 71.0} (0.3)$  &  $\substack{50.2 \\ 49.7} (0.2)$  &  $\substack{48.7 \\ 48.0} (0.2)$  & $\substack{80.5 \\ 79.8} (0.3)$  & $\substack{79.1 \\ 78.5} (0.2)$  &$\substack{87.9 \\ 87.0} (0.3)$  & $\substack{86.6 \\ 86.0} (0.2)$  & $\substack{52.7 \\ 52.0} (0.2)$  & $\substack{49.5 \\ 49.0} (0.2)$ \\ 

GCA &$\substack{73.3 \\ 72.8} (0.2)$ & $\substack{72.1 \\ 71.2} (0.3)$  &  $\substack{49.8 \\ 49.2} (0.2)$  &  $\substack{48.5 \\ 47.2} (0.4)$  & $\substack{80.2 \\ 79.4} (0.3)$  & $\substack{78.5 \\ 77.9} (0.2)$  &$\substack{88.2 \\ 87.7} (0.2)$  & $\substack{86.9 \\ 86.0} (0.3)$  & $\substack{52.5 \\ 51.6} (0.3)$  & $\substack{49.0 \\ 48.1} (0.3)$ \\ 

SGCL & $\substack{74.4 \\ 74.0} (0.1)$ & $ \substack{73.4 \\ 72.5} (0.3)$ & $\substack{50.7 \\ 50.6} (0.1) $  & $ \substack{50.2 \\ 49.9} (0.1)$  & $\substack{81.6 \\ 81.4} (0.1)$ & $ \substack{80.3 \\ 79.4}(0.2)$   & $\substack{88.7 \\ 88.4} (0.1) $ & $\substack{87.6 \\ 87.1} (0.2)$  & $\substack{54.6 \\ 54.3} (0.1)$ & $\substack{51.4 \\ 50.3} (0.4)$  \\\bottomrule
\end{tabular}}
\end{table}

\begin{table*}[h]

    \caption{Top-$k$ similarity search~($k=20, 40$), frozen cases only, 5-95 intervals with standard deviations in brackets.}
    \centering
{%
    \begin{tabular}{l | r r |r r| r r@{~}}
        \toprule
                            & \multicolumn{2}{c|}{KDD-ICDM} & \multicolumn{2}{c|}{SIGIR-CIKM} & \multicolumn{2}{c}{SIGMOD-ICDE}                                                       \\\midrule
        $|V|$               & 2,867                         & 2,607                           & 2,851                           & 3,548           & 2,616           & 2,559           \\
        $|E|$               & 7,637                         & 4,774                           & 6,354                           & 7,076           & 8,304           & 6,668           \\
        \#~ground truth      & \multicolumn{2}{r|}{697}      & \multicolumn{2}{r|}{874}        & \multicolumn{2}{r}{898}                                                               \\\midrule
        $k$                 & 20                            & 40                              & 20                              & 40              & 20              & 40              \\            \midrule

GraphCL  &      $\substack{0.1062 \\ 0.912}$  (0.0033)       & $\substack{0.1604 \\ 0.1546}$             (0.0016)                &          $\substack{0.0622 \\ 0.0552}$  (0.0024)                & $\substack{0.1259 \\ 0.1167}  (0.0031$)   & $\substack{0.0871 \\ 0.0714}  (0.0048)$        & $\substack{0.1294 \\ 0.1142}  (0.0042)$       \\

        GRACE  &   $\substack{0.1085 \\ 0.967}$  (0.0037)                       &       $\substack{0.1587 \\ 0.1529}$  (0.0022)             & $\substack{0.0612 \\ 0.0508}  (0.0035)$  &     $\substack{0.1317 \\ 0.1109}$  (0.0041)    &  $\substack{0.0955 \\ 0.078}$  (0.0057) & $\substack{0.1287 \\ 0.1252}$  (0.0012)     \\

        MVGRL   &    $\substack{0.1032 \\ 0.958}$  (0.0028)                         &       $\substack{0.1554 \\ 0.1418}$   (0.0041)            & $\substack{0.0582 \\ 0.0456}$  (0.0038)  &       $\substack{0.1285 \\ 0.1089}$  (0.0058)    & $\substack{0.0875 \\ 0.0687}$  (0.0062) &  $\substack{0.1194 \\ 0.1128}  (0.0025)$     \\

        CuCo   &   $\substack{0.1091 \\ 0.1027}$  (0.0026)                      &          $\substack{0.1591 \\ 0.1497}$  (0.0031)           & $\substack{0.0634 \\ 0.0502}$  (0.0040) &     $\substack{0.1342 \\ 0.1209}$  (0.0038)    & $\substack{0.1028 \\ 0.0857}$  (0.0052) &    $\substack{0.1508 \\ 0.1226}$  (0.0081)    \\

        BYOV   &           $\substack{0.1088 \\ 0.1015}$  (0.0019)                   &        $\substack{0.1612 \\ 0.1552}  (0.0028)$           & $\substack{0.0649 \\ 0.0556}  (0.0029)$  &        $\substack{0.1338 \\ 0.1219}  (0.0038)$   & $\substack{0.0978 \\ 0.0675}  (0.0102)$ &  $\substack{0.1417 \\ 0.1235}  (0.0068)$       \\

        InfoGCL   &           $\substack{0.1031 \\ 0.0928}$  (0.0038)                   &        $\substack{0.1592 \\ 0.1504}  (0.0032)$           & $\substack{0.0628 \\ 0.0572}  (0.0026)$  &        $\substack{0.1286 \\ 0.1167}  (0.0047)$   & $\substack{0.0847 \\ 0.0764}  (0.0035)$ &  $\substack{0.1292 \\ 0.1163}  (0.0043)$       \\

        GCA   &           $\substack{0.1056 \\ 0.0944}  (0.0034)$                   &        $\substack{0.1592 \\ 0.1524}  (0.0027)$           & $\substack{0.0587 \\ 0.0522}  (0.0025)$  &        $\substack{0.1254 \\ 0.1154}  (0.0035)$   & $\substack{0.0916 \\ 0.0782}  (0.0048)$ &  $\substack{0.1383 \\ 0.1122}  (0.0092)$       \\
        
        \midrule
SGCL  &       $\substack{0.1137 \\ 0.1073}$  (0.0024)         & $\substack{0.1675 \\ 0.1614}$        (0.0028)                     &          $\substack{0.0721 \\ 0.0587}$  (0.0042)              & $\substack{0.1426 \\ 0.1289}$  (0.0046)  &$\substack{0.1124 \\ 0.1046}$  (0.0028)         & $\substack{0.1615 \\ 0.1502}$  (0.0035)    \\

\midrule
        \multicolumn{7}{c}{MoCo} \\ \hline

GraphCL  &       $\substack{0.0877 \\ 0.0782}$  (0.0038)         & $\substack{0.1561 \\ 0.1467}$         (0.0031)                  &          $\substack{0.0656 \\ 0.0605}  (0.0018) $             & $\substack{0.1198 \\ 0.1127}$  (0.0021)  &  $\substack{0.0918 \\ 0.0836}$  (0.0021)         &  $\substack{0.1479 \\ 0.1358}$  (0.0029)     \\

        GRACE  &    $\substack{0.0895 \\ 0.0818}$  (0.0026)                          &         $\substack{0.1605 \\ 0.1421}$  (0.0051)            & $\substack{0.0702 \\ 0.0522}  (0.0046) $  &       $\substack{0.1258 \\ 0.1095}  (0.0042) $    & $\substack{0.0982 \\ 0.0845}  (0.0049) $ & $\substack{0.1552 \\ 0.1377}  (0.0058) $     \\

        MVGRL   &       $\substack{0.0874 \\ 0.0789}$  (0.0028)                          &       $\substack{0.1522 \\ 0.1395}$  (0.0037)              & $\substack{0.0622 \\ 0.0512}  (0.0036)$ &       $\substack{0.1286 \\ 0.1028}  (0.0074) $   & $\substack{0.0902 \\ 0.0812}  (0.0036) $ &  $\substack{0.1437 \\ 0.1330}  (0.0037) $     \\

        CuCo   &        $\substack{0.0895 \\ 0.0831}$  (0.0019)                      &        $\substack{0.1582 \\ 0.1459}$  (0.0052)             & $\substack{0.0691 \\ 0.0547}  (0.0045) $  &     $\substack{0.1295 \\ 0.1113}  (0.0052) $    & $\substack{0.0988 \\ 0.0792}  (0.0061) $ & $\substack{0.1532 \\ 0.1296}  (0.0072) $      \\

        BYOV   &          $\substack{0.0962 \\ 0.0891}$   (0.0019)                   &           $\substack{0.1596 \\ 0.1502}$  (0.0027)           & $\substack{0.0688 \\ 0.0592}  (0.0027) $  &       $\substack{0.1269 \\ 0.1194}  (0.0020) $   & $\substack{0.1022 \\ 0.0716}  (0.0082) $ & $\substack{0.1559 \\ 0.1347}  (0.0058) $      \\

        InfoGCL   &          $\substack{0.0895 \\ 0.0783}$   (0.0047)                   &           $\substack{0.1582 \\ 0.1473}$  (0.0038)           & $\substack{0.0652 \\ 0.0583}  (0.0023) $  &       $\substack{0.1224 \\ 0.1130}  (0.0027) $   & $\substack{0.0922 \\ 0.0835}  (0.0028) $ & $\substack{0.1463 \\ 0.1362}  (0.0037) $      \\
        
        GCA   &          $\substack{0.0885 \\ 0.0803}   (0.0024)$                   &           $\substack{0.1528 \\ 0.1475}  (0.0017)$           & $\substack{0.0648 \\ 0.0552}  (0.0031) $  &       $\substack{0.1246 \\ 0.1140}  (0.0029) $   & $\substack{0.0898 \\ 0.0826}  (0.0024) $ & $\substack{0.1455 \\ 0.1392}  (0.0018) $      \\
        
        \midrule
SGCL  &       $\substack{0.1058 \\ 0.0917}$  (0.0045)          & $\substack{0.1703 \\ 0.1568}$  (0.0041)                             &          $\substack{0.0816 \\ 0.0727}$  (0.0037)               &  $\substack{0.1347 \\ 0.1264}$  (0.0029)    & $\substack{0.1093 \\ 0.1028}$  (0.0028)         & $\substack{0.1628 \\ 0.1549}$  (0.0031)      \\

        \bottomrule
    \end{tabular}}
        \label{tab:ss_confintervals}

\end{table*}

\clearpage

\section{Auxiliary experiments on OGB datasets, MNIST and CIFAR-10}

\subsection{MNIST and CIFAR-10 motivation - implicit assumptions for the crop augment}

We now revisit the chain of reasoning that motivates the crop augmentation, enumerated below sequentially : 

\begin{itemize}
    \item Images are a very important, naturally occurring subclass of attributed grid graphs (products of line graphs). Indeed, for any grid graph, assigning the nodes a $3$-dimensional attribute corresponding to the RGB intensity assigns the colour part of the image. For the spatial aspect, every grid graph is enumerable in the indices of its constituent line graphs that it is a product of, i.e. we may denote a node of the grid graph as $v_{ij}$ where $1 \leq i \leq m, 1 \leq j \leq n$ for a grid graph that is the product of two line graphs with $m,n$ nodes. Associate $x_{i,j}$, $y_{i,j}$ with every such $v_{ij}$, with the condition that :
    
    $$ x_{i,j+1} - x_{i,j} = x_{i,j} - x_{i,j-1} $$
    $$ y_{i+1,j} - y_{i,j} = y_{i,j} - y_{i-1,j}$$
    
    Clearly, then, every image can be expressed as a grid graph, while the converse is not true. We assume that this generalization is meaningful - after all, an image of shape $(m,n)$ can equally be flattened and written as a $1$-dimensional sequence with its dimensions appended separately, yielding a length of $mn+2$ per channel. Every image can be expressed this way while not every $1d$ sequence of length $mn+2$ can be formed into an image, making this a generalization. \textbf{We need to demonstrate that the grid graph form of generalizing what an image is, turns out to be more meaningful via some metric than, for example, ad hoc flattening.} 
    
    \item The crop operation in images, when they are considered equivalent to grid graphs, is equivalent to a value-based thresholding on nodes, depending on the values induced on them using the first two eigenvectors corresponding to the first two nonzero eigenvalues of the Laplacian. This is indeed true, ignoring numerical errors in the eigendecomposition, when the dimensions $m,n$ with $m>n$ of the image are such that $2n > m$. However, the crop operation for images happens to be functional even when $2n > m$, which is not true for the eigenvector-based cropping we propose. 
    \item The crop augment is known to be - practically and empirically - a runaway success among the candidate augmentations in contrastive learning so far, when the representations to be learnt are to be evaluated for image classification.  
    
    \item Clearly, if the image is to be thought of as a graph, the corresponding expectation is that our proposed graph-level crop succeed for graph classification. Therefore, we investigate if value thresholding based on the two eigenvectors, which is strictly a generalization of the crop operation, is a similar success on graphs in general. 
\end{itemize}

What are the questionable steps taken above ? First, using the first two eigenvectors is one of infinitely many generalizations possible of the crop augmentation. We cannot investigate all such generalizations, but we can instead check if this particular generalization continues to hold when the domain (images) is perturbed. 

Secondly, to what extent is an image actually a grid graph ? Does such a generalization remove key aspects of the image ?

We can see that for the latter assumption, a start would be to consider the image classification tasks such as the ubiquitous tasks on MNIST and CIFAR-10, and turn them instead into graph classification tasks, after converting the images into grid graphs. If this process makes the task hopeless, the assumption is assuredly more questionable.

In fact, such benchmarking~\citep{dwivedi2020benchmarking} on MNIST and CIFAR-10 has already been carried out with Graph neural networks. The accuracy obtained is close to $100 \%$ for MNIST, and above $65 \%$ for CIFAR-10, which, while not exceptional, clearly shows that some reasonable information is retained relevant to the class labels by converting images to a grid graph.

Importantly, given such a grid graph, the nodes i.e. the pixels are initialized with their positions for such a graph classification task. We recall from our discussion of the spectra of grid graphs, that it is precisely the $(x,y)$ positions that will be recovered via the two relevant eigenvectors. 

If our generalization is correct, then we expect that at the point of generalization - i.e. in the original domain, the generalization and the specific operation it is generalizing (crop) will be identical operations. We now need to change the domain as slightly as possible to the level where the generalization remains valid, but the specific operation can no longer be performed.

This is easily achievable by replacing images (grid graphs) with their \textbf{subgraphs} and assuming we have no clue how these graphs came to be (an usual assumption made for graph datasets). Recall that the $(x,y)$ positions to grid graphs were assigned using the knowledge that they were images. However, if we do not know that they are images, we can only use their adjacency matrices. 

In the case of the complete grid graph, the adjacency matrix will be enough to recover the $(x,y)$ co-ordinates of each pixel. However, for a subgraph, the two eigenvectors induce different values that need not correlate to $(x,y)$ co-ordinates.

Recall that we have claimed that the values induced by these eigenvectors are useful for segmenting (selecting subgraphs from) graphs of arbitrary kinds for contrastive learning in the view creation process, using the images as an example. If they are useful for arbitrary graphs as our graph classification benchmarks indicate, they must be useful for slightly perturbed (transformed into subgraph) version of images. It should be understood that we are talking of usefulness solely in the sense of learning optimal representations for downstream classification tasks. If they cannot even succeed at this, then our reasoning is likely to be questionable.

Therefore, if the first two eigenvectors yield a positional encoding that is useful for the image classification task when the images are transformed into grid graphs and then made subgraphs of, the results will be consistent with our assumptions. Further, since the image has only meaningful co-ordinates upto $2$ axes, we expect no benefits for increasing the dimensionality of such spectral embeddings beyond $2$.

\begin{figure}[ht]
    \centering
    \includegraphics[width=\textwidth]{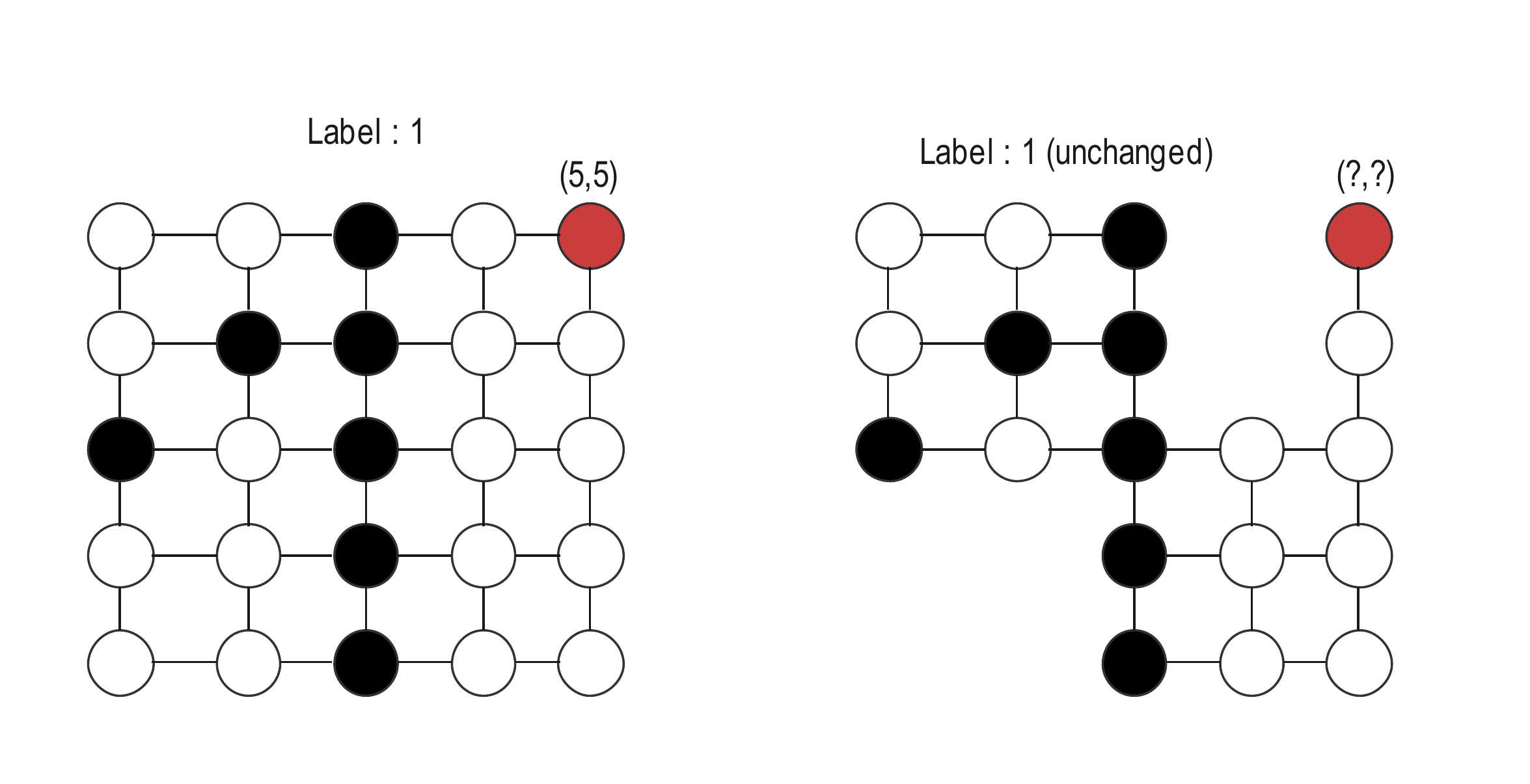}
    \caption{The layout of the subgraph classification experiment, as designed for MNIST. The aim of positional encoding is to give the node (red) a two-dimensional embedding that is as useful to find the label ($1$) as its initial $(x,y)$ co-ordinate pairing of $(5,5)$ with $1$-indexing. If the first two eigenvectors suffice, they are valid replacements for the axes and yield meaningful embeddings even when the graph is no longer a perfect grid, and this will be reflected in higher accuracy. We assume that for images, a meaningful embedding must at least capture some positional information and thus eigenvector embeddings, if they work, will be validated in generalizing axis-based co-ordinates.}
    \label{fig:subgraphclassification}
\end{figure}

\subsection*{Nature of testing}

We consider the following cases, on top of a previously investigated baseline scenario, where each image is converted to a grid graph and each pixel to a node, with edges between adjacent pixels, and the node attribute is $1$ or $3$ dimensional respectively for MNIST and CIFAR-10, to which $2$ dimensions are added via the Laplacian decomposition's eigenvectors corresponding to first two nonzero eigenvalues, bringing the problem into a graph classification problem, where a GCN ($5$-layer GIN of the same architecture for consistency) is used to process this grid graph, and the node-level representations pooled. 

The variants we investigate are :

\begin{itemize}
    \item Keeping every graph as-is
    \item Replacing each graph with a subgraph, which consists of the nodes visited by a random walk starting from the center of each graph, taking $512$ steps with a return probability of $0.1$
    \item Performing the subgraph step with a random graph crop on top of each subgraph, to simulate our augment scenario.
    \item Change the positional embedding to either be absent, have only the first dimension, or have $5$ dimensions.
\end{itemize}

In each of the following tables, namely tables~\ref{tb:mnist1},~\ref{tb:mnist2},~\ref{tb:mnist3},~\ref{tb:mnist4} for MNIST and ~\ref{tb:cif1},~\ref{tb:cif2},~\ref{tb:cif3},~\ref{tb:cif4} for CIFAR-10, rows signify train sets, and columns signify test sets in terms of the modifications performed on them. Overall, we see the same pattern. The random walk, or the subsequent cropping, do not significantly harm the accuracy. There are large gains from going from $0$-dimensional positional embeddings to $1$, smaller ones from $1$ to $2$ and beyond $2$, a significant drop at $5$. \textbf{This matches what we expect and justifies our assumptions}.

\begin{table}[ht]
\begin{tabular}{|l|l|l|l|}
\hline
\centering
                   & Original & Random Walk & Random Walk + Crop \\ \hline
Original           &    97.8      &  94.3           &  89.5                  \\ \hline
Random Walk        &    93.2    &      93.4       &       88.9             \\ \hline
Random Walk + Crop &    92.7      &     91.5        &    92.8                \\ \hline

\end{tabular}

\caption{MNIST, $2$-dimensional embedding}
\label{tb:mnist1}
\begin{tabular}{|l|l|l|l|}
\hline
\centering
                   & Original & Random Walk & Random Walk + Crop \\ \hline
Original           &   94.2       &     87.4        &  85.6                  \\ \hline
Random Walk        &    86.5      &       85.1      &       83.5             \\ \hline
Random Walk + Crop &     83.2     &       81.2      &      85.9              \\ \hline
\end{tabular}

\caption{MNIST, $1$-dimensional embedding}
\label{tb:mnist2}
\begin{tabular}{|l|l|l|l|}
\hline
                   & Original & Random Walk & Random Walk + Crop \\ \hline
Original           &    68.9      &   61.5          &  58.8                  \\ \hline
Random Walk        &    62.4      &       55.6      &        53.2            \\ \hline
Random Walk + Crop &    61.9      &      54.4       &         52.6           \\ \hline
\end{tabular}

\caption{MNIST, $0$-dimensional embedding}
\label{tb:mnist3}
\begin{tabular}{|l|l|l|l|}
\hline
\centering
                   & Original & Random Walk & Random Walk + Crop \\ \hline
Original           &    95.6      &     93.2       &   87.2                \\ \hline
Random Walk        &      91.5    &        84.2     &     84.5               \\ \hline
Random Walk + Crop &       91.9   &      86.5       &      83.8              \\ \hline
\end{tabular}

\caption{MNIST, $5$-dimensional embedding}
\label{tb:mnist4}
\end{table}

\begin{table}[ht]
\begin{tabular}{|l|l|l|l|}
\hline
\centering
                   & Original & Random Walk & Random Walk + Crop \\ \hline
Original           &    59.5      &     56.8        &  55.9                   \\ \hline
Random Walk        &    56.2      &     53.7        &    53.9                \\ \hline
Random Walk + Crop &    54.6      &     52.9        &       54.3             \\ \hline
\end{tabular}

\caption{CIFAR-10, $2$-dimensional embedding}
\label{tb:cif1}
\begin{tabular}{|l|l|l|l|}
\hline
\centering
                   & Original & Random Walk & Random Walk + Crop \\ \hline
Original           &    55.4      &     49.7        &  52.3                  \\ \hline
Random Walk        &    54.2     &      51.2       &   50.8                 \\ \hline
Random Walk + Crop &    53.2      &      47.8       &   53.2                 \\ \hline
\end{tabular}

\caption{CIFAR-10, $1$-dimensional embedding}
\label{tb:cif2}
\begin{tabular}{|l|l|l|l|}
\hline
\centering
                   & Original & Random Walk & Random Walk + Crop \\ \hline
Original           &    46.9      &     41.5        &  43.2                  \\ \hline
Random Walk        &     39.8     &     39.6        &   40.5                \\ \hline
Random Walk + Crop &     43.7     &     38.8        &     40.2               \\ \hline
\end{tabular}

\caption{CIFAR-10, $0$-dimensional embedding}
\label{tb:cif3}

\begin{tabular}{|l|l|l|l|}
\hline
\centering
                   & Original & Random Walk & Random Walk + Crop \\ \hline
Original           &    52.7      &    51.0         &   53.5                 \\ \hline
Random Walk        &  52.4        &     49.6        &   51.0                 \\ \hline
Random Walk + Crop &    51.3      &    50.5         &      50.2              \\ \hline
\end{tabular}

\caption{CIFAR-10, $5$-dimensional embedding}
\label{tb:cif4}
\end{table}

\clearpage

\subsection{Results on OGB datasets of Arxiv (accuracy) and molhiv (HIV) (ROC-AUC)}

We also tested our pre-trained models on datasets associated with the Open Graph Benchmark aka OGB~\citep{hu2020open}. Here, the entire test occurs in the fine-tuned setting. We observed some mild benefits associated with pre-training over the common sense GIN benchmark, even when both networks had the advantage of utilizing the structural embedding (Recall, of course, that only the structural embedding aspect can transfer between widely divergent datasets that share no attributes otherwise). These results are summarized in table~\ref{tab:ogb}.

\begin{table}[h]

\centering
\begin{tabular}{|l|l|l|}

\hline
                                                    & Arxiv & HIV \\ \hline
GIN - Attr + Struct                                 &  72.1     &  77.0    \\ \hline
GIN - Attr only                                     &  71.4      &  76.8   \\ \hline
GIN - Attr and Struct - GCC E2E Finetuned           &   72.3     &  77.2   \\ \hline
GIN - Attr and Struct - GCC MoCo Finetuned          &  72.3     &  77.5   \\ \hline
GIN - Attr and Struct - SGCL E2E Finetuned           &  72.6      &  77.4    \\ \hline
GIN - Attr and Struct - SGCL MoCo Finetuned     & 72.5       &  77.8   \\ \hline
\end{tabular}
\caption{Results on OGB datasets}
\label{tab:ogb}
\end{table}

\subsection{Citeseer and Cora}

We ran the frozen E2E transfer case for Citeseer and Cora datasets. When we transferred our structure-only models to these datasets and did not use any node attributes, we observed 50.8 (1.6) and 68.7 (2.1) percent accuracy on Citeseer and Cora respectively (standard deviations in brackets).

With node features included along with the frozen encoder, the performance rose to 71.5 (1.2) and 82.1 (1.6) respectively. The fact that these values (50.8 and 68.7) are significantly higher than a random guess (approx. 14.3 and 16.7) indicates that the structure-only encoder trained on a completely different pre-training corpus is still able to learn important topological characteristics.

\clearpage
\section{Limitations, societal impact and reproducibility}

\noindent\textbf{Limitations.} Our paper is not without limitations. Currently, the pre-train datasets we use in our paper are mostly inherited from established work focusing on GNN pre-training~\citep{qiu2020gcc,you2020graphcl}. Even though they are sufficiently large in terms of the scale of the graph, we believe our model can be further strengthened by more diverse graph datasets. As a very recent paper GraphWorld~\citep{palowitch2022graphworld} addressed, the commonly used datasets have graph statistics that lie in a limited and sparsely-populated region in terms of metrics such as the clustering coefficient, degree distribution, or Gini coefficient. Thus, to fully benefit from the power of pre-training techniques for graph data, it would be interesting and important to extend the use of pre-train datasets to graphs with diverse structural properties. 

Another limitation of our work is that the pre-training and transfer focuses exclusively on the graph structure information; this is a common approach for cross-domain training~\citep{qiu2020gcc}. We believe that there is value in further investigation into techniques that can process the node feature information as well as the structure information during the pre-train stage. This especially can be seen with the OGB datasets, which may share structural information between, for example, molecules and citation networks, while sharing no attribute related information.

\noindent\textbf{Potential societal impact.}
Graph neural network techniques have been commonly used for prediction tasks in social networks and recommender systems. Our techniques, as a variant of graph neural networks, can be used in those scenarios to further improve the model performance. However, having such an ability is a double-edged sword. On one hand, it can be beneficial to improve user experience. On the other hand, if these techniques are used purely for a profit-driven or political driven reason, they can aim to 1) monopolize user attention for as long as possible by seducing users into carrying out actions that will make them happy on very short timescales  and addictive to their product or even worse to 2) shape and influence public opinion on the key matters of the time. Thus, researchers and service providers should pay vigilant attention to ensure the research does end up being used positively for the social good.

\textbf{Reproducibility.} We have included the code and all hyperparameters, hardware details etc. to facilitate reproducibility.

\clearpage

\section{Additional ablations}

In this section, we present the variation of the model's success with changes in train dataset, degree, the standard deviation of the degree, and $\lambda_2$ (the second eigenvalue of the Laplacian).

\begin{table}[h]
\centering
\caption{Advantage of SGCL over GCC, with GCC performance in brackets, when the train set is restricted to particular graphs. The graphs are listed in descending order of their sizes.}
\begin{tabular}{|l|l|l|l|l|}
\hline
Advantage   & IMDB-M & COLLAB & RDT-B & RDT-M \\ \hline
All         & 0.5(49.3)    & 0.9(74.7)    & 0.9(87.5)   & 1.2(52.6)   \\ \hline
LiveJournal & 0.4 (48.8)    & 0.9 (74.2)    & 1.0 (86.9)   & 1.1 (51.8)   \\ \hline
Facebook    & 0.5 (49.0)    & 0.8 (74.0)    & 1.0 (86.7)   & 1.1 (51.6)   \\ \hline
IMDB        & 0.6 (48.2)    & 1.1 (73.8)    & 1.1 (86.5)   & 1.2 (51.5)   \\ \hline
DBLP        & 0.7 (49.1)    & 1.1 (74.3)    & 1.2 (86.4)   & 1.1 (51.9)   \\ \hline
Academia    & 0.6 (48.1)    & 1.0 (74.5)    & 1.1 (86.2)  & 1.1 (51.2)   \\ \hline
\end{tabular}
\end{table}

In the ablations against $\lambda_2$, degree, and standard deviation of the degree we see a pronounced U-curve where the middle quintiles perform best. This could be due to the hypothesized spectral gap effect that we derive. The results in degree statistics could well be due to the fact that such statistics in turn depend greatly on the $\lambda_2$ values, and cannot be considered truly independent findings.

\begin{table}[h]
\centering
\caption{Advantage of SGCL over GCC, with GCC performance in brackets, when the train set is restricted to particular graphs. The rows represent rank quintiles of $\lambda_2$.}
\begin{tabular}{|l|l|l|l|l|}

\hline
Advantage   & IMDB-M & COLLAB & RDT-B & RDT-M \\ \hline
Q1         & 0.3 (47.6)    & 0.6 (74.2)    & 0.8 (86.8)   & 0.9 (51.6)   \\ \hline
Q2       & 0.3 (48.7)    & 0.9 (74.0)    & 0.8 (86.7)    & 1.0 (51.5)   \\ \hline
Q3      & 0.7 (49.5)    & 1.1 (74.9)    & 1.2 (87.4)   & 1.5 (52.3)   \\ \hline
Q4        & 0.9 (49.4)    & 1.2 (75.0)   & 1.2 (88.0)   & 1.4 (53.0)   \\ \hline
Q5        & 0.3 (49.4)    & 0.6 (74.9)    & 0.7 (87.8)  & 1.0 (52.9)   \\ \hline
\end{tabular}
\end{table}

\begin{table}[h]
\centering
\caption{Advantage of SGCL over GCC, with GCC performance in brackets, when the train set is restricted to particular graphs. The rows represent rank quintiles of average degree.}
\begin{tabular}{|l|l|l|l|l|}

\hline
Advantage   & IMDB-M & COLLAB & RDT-B & RDT-M \\ \hline
Q1         & 0.4 (48.6)    & 0.7 (74.3)    & 0.8 (87.4)   & 1.0 (52.4)   \\ \hline
Q2       & 0.5 (49.2)    & 0.8 (74.5)    & 1.0 (87.6)   & 1.2 (52.5)  \\ \hline
Q3      & 0.6 (49.0)    & 1.0 (74.8)    & 0.9 (87.2)   & 1.3 (52.2)   \\ \hline
Q4        & 0.5 (49.4)    & 1.0 (74.6)   & 1.0 (87.3)   & 1.3 (52.4)   \\ \hline
Q5        & 0.6 (49.3)   & 1.1 (74.2)    & 1.0 (87.6)   & 1.3 (52.5)  \\ \hline
\end{tabular}
\end{table}

\begin{table}[h]
\centering
\caption{Advantage of SGCL over GCC, with GCC performance in brackets, when the train set is restricted to particular graphs. The rows represent rank quintiles of the standard deviation of the degree.}
\begin{tabular}{|l|l|l|l|l|}

\hline
Advantage   & IMDB-M & COLLAB & RDT-B & RDT-M \\ \hline
Q1         & 0.5 (48.4)   & 0.8 (74.3)    & 0.8 (87.2)   & 1.2 (52.9)   \\ \hline
Q2       & 0.6 (49.5)    & 0.9 (74.6)   & 1.1 (87.6)   & 1.3 (52.3)   \\ \hline
Q3      & 0.6 (49.6)    & 1.0 (75.0)    & 1.2 (87.6)   & 1.3 (52.4)   \\ \hline
Q4        & 0.6 (50.0)   & 1.1 (75.2)    & 1.2 (87.7)  & 1.3 (52.2)   \\ \hline
Q5        & 0.3 (49.0)   & 0.6 (74.2)    & 0.7 (87.5)  & 1.0 (52.2)   \\ \hline
\end{tabular}
\end{table}

\begin{table}[h]
\centering
\caption{The behavior of SGCL, E2E, Frozen on spectrally splitting a train dataset(DBLP) into 5 quintiles along the rows, according to the value of $\lambda_2$, while testing on a similar split across the columns on COLLAB. Both diagonal and middle quintiles show elevated values.}
\begin{tabular}{|l|l|l|l|l|l|}
\hline
SGCL Accuracy   & Q1 & Q2 & Q3 & Q4 & Q5 \\ \hline
Q1         & 73.8 & 73.5 & 73.6 & 73.4 & 73.2   \\ \hline
Q2       & 73.4 & 73.7 & 73.8  & 73.6 & 73.7   \\ \hline
Q3      & 73.8 & 74.2 & 74.2 & 74.1 & 74.0  \\ \hline
Q4        & 73.5 & 73.9 & 74.2 & 74.3 & 73.8    \\ \hline
Q5        & 73.3 & 73.2 & 73.7 & 73.6 &  73.6  \\ \hline
\end{tabular}
\label{eigenvariation}
\end{table}

\clearpage

\section{Theorems underlying the augmentations}

\subsection{Crop augmentation}

Let us derive a few key claims that will help us put the crop augment on a surer footing.

Denote by $P_n$ the path-graph on $n$ vertices, which has $n{-}1$ edges of form $(i,i+1)$ for $i=1,\dots,n{-}1$. This corresponds to the line graph.

Define also $R_n, n \geq 3$, the ring graph on $n$ vertices defined as $P_n$ with an extra edge between $1$ and $n$. 

Recall the {\em product graph} : A product of two graphs $A,B$ with vertex sets $(v_A,v_B)$ and edge sets $(e_A,e_B)$ is a graph $A.B$ where each $v \in A.B$ can be identified with an ordered pair $(i,j), i \in v_A, j \in v_B$. Two nodes corresponding to $(i,j),(i',j')$ in $A.B$ have an edge between them if and only if either $i'=i, (j,j') \in v_B$ or $(i,i') \in v_A, j=j'$. The product of two line graphs of length $M,N$ respectively can be represented as a planar rectangular grid of lengths $M,N$. Denote by $G_{a,b}$ the rectangular grid graph formed by the product $P_a.P_b$. Structurally, this graph represents an image with dimensions $a \times b$. 

For simplicity, we will prove our work for unweighted and undirected graphs, but the properties are general and do not require these conditions.

\begin{theorem}
Let $A$ be a graph with eigenvalues of the Laplacian as $\lambda_1, \lambda_2, \dots, \lambda_N$ and corresponding eigenvectors $\boldsymbol{v}_1, \dots, \boldsymbol{v}_N$. Similarly consider $B$ another graph with eigenvalues $\mu_1, \dots \mu_M$ and eigenvectors $\boldsymbol{u}_1, \dots \boldsymbol{u}_M$. Let the product of graphs $A,B$ be $C$. Then, identifying each node in $C$ with an ordered pair $(x,y)$, the Laplacian of $C$ has an eigenvector $\boldsymbol{w}_{ij}$ with eigenvalue $\lambda_i + \mu_j$, such that 

$$\boldsymbol{w_{ij}}(x,y) = \boldsymbol{v_i}(x) \times \boldsymbol{u_j}(y) $$
\end{theorem}

\textbf{Proof} : let the laplacian of $C$ be $\boldsymbol{L_C}$. We need only compute the term

$$ \boldsymbol{L_C(w_{ij}}(x,y)) $$

This is equivalent to (with $e_A,e_B$ being the edge set of $A,B$ respectively) :

$$ \sum_{(x,x') \in e_A} (\boldsymbol{w_{ij}}(x,y) - \boldsymbol{w_{ij}}(x',y)) + \sum_{(y,y') \in e_B} (\boldsymbol{w_{ij}}(x,y) - \boldsymbol{w_{ij}}(x,y'))  $$

However, taking $ \sum_{(x,x') \in e_A} (\boldsymbol{w_{ij}}(x,y) - \boldsymbol{w_{ij}}(x',y))$, we observe that :

$$ \sum_{(x,x') \in e_A} (\boldsymbol{w_{ij}}(x,y) - \boldsymbol{w_{ij}}(x',y))$$

becomes, applying the hypothesized $\boldsymbol{w_{ij}} = \boldsymbol{v_i}(x) \times \boldsymbol{u_j}(y)$

$$ \sum_{(x,x') \in e_A}  (\boldsymbol{v_{i}}(x) \boldsymbol{u_j}(y) - \boldsymbol{v_{i}}(x') \boldsymbol{u_j}(y))$$

Taking $\boldsymbol{u_j}(y)$ in common, we recognize that $\sum_{(x,x') \in e_A} \boldsymbol{v_i}(x) - \boldsymbol{v_i}(x')$ will yield just $\boldsymbol{v_i}$ scaled by $\lambda_i$ as $\boldsymbol{v_i}$ is the eigenvector of the Laplacian.

Therefore this term becomes

$$ \sum_{(x,x') \in e_A}  \boldsymbol{u_j}(y) \times \boldsymbol{v_{i}}(x) \times \lambda_i$$

While the other term, i.e. $$ \sum_{(y,y') \in e_B} (\boldsymbol{w_{ij}}(x,y) - \boldsymbol{w_{ij}}(x,y')) $$ yields similarly

$$ \sum_{(y,y') \in e_B}  \boldsymbol{u_j}(y) \times \boldsymbol{v_{i}}(x) \times \mu_j$$

Adding the two, we see that the final matrix-vector product is parallel to the original vector (thus an eigenvector) with eigenvalue $\lambda_i + \mu_j$.

\begin{theorem}
The eigenvectors of the (un-normalized) Laplacian of $P_n$, for $n > k \geq 0$, are  of the form:

$$\boldsymbol{x_k}(u) = \cos (\pi k u / n - \pi k / 2n)$$

with eigenvalues $\lambda_k$

$$2 - 2\cos(\pi k/n)$$.
\end{theorem}

\textbf{Proof} : We will use the ring graph defined above. Let $P_n$ be the path graph. $R_{2n+2}$ is clearly the ring graph obtained by having two copies of $P_n$ with $2$ additional links.

Now, $R_n$ can be drawn on the plane with the vertex $i$ located at $(\cos(\alpha i), \sin(\alpha i))$ where $\alpha = \frac{2 \pi} {n}$. Observe that each vertex $i$ has a position in the plane which is parallel to the sum of the position vectors of $i+1$ and $i-1$. From this, it naturally follows (by the definition of the Laplacian operator which subtracts the value of the neighbour vectors from that at the node) that the valid eigenvectors for $R_n$ are :

$$ \boldsymbol{x_k}(i) = \cos(\alpha k i),  \boldsymbol{y_k}(i) = \sin(\alpha k i) $$

Regarding the eigenvalue, the node itself contributes $2$ (as it appears in the sum twice) and each neighbour contributes $-\cos(\alpha k)$ with $2$ neighbours, leading to an eigenvalue of $2 - 2 \cos(\alpha k)$.

Now it is trivial to find the eigenvectors of $P_n$ from $R_{2n}$. Simply take any eigenvector of $R_{2n}$ which has the same value for $i,i+n$ for $i \leq n$. Then the restriction of this eigenvector to $1 \leq i \leq n$ defines a valid eigenvector for $P_n$ with the same eigenvalue. This is why the terms of the angles in the theorem are the same as path graphs with $\pi$ taking the place of $2 \pi$ as, for example 

$$2 - 2 \cos(\frac{2 \pi} {2n} k) = 2 - 2 \cos(\frac{\pi} {n} k) $$

Which is the sought result.

\clearpage

\subsection{Reordering augmentation}

\begin{theorem}
Let $\boldsymbol{A}$ be the adjacency matrix of an undirected unweighted graph, $\boldsymbol{D}$ the degree matrix and $\boldsymbol{P} = \boldsymbol{D}^{-1/2} \boldsymbol{A} \boldsymbol{D}^{-1/2}$ the normalized adjacency matrix. Let $\boldsymbol{D}_k$ be the $k$-th order normalized diffusion matrix :

$$ \boldsymbol{D}_k = \sum_{i=1}^{k} \boldsymbol{P}^k $$

Then, the $j$-th eigenvector (sorted in order of eigenvalues and ties broken consistently) is the same for all $j$ for any odd $k$. That is, for any odd $k$, $\boldsymbol{D}_k$ and $\boldsymbol{P}$ have eigenvectors ordered in the same sequence.  

\end{theorem}

\textbf{Proof} : First, since the normalized Laplacian matrix $\boldsymbol{L}$ is related to $\boldsymbol{P}$ as $\boldsymbol{L} = \boldsymbol{I} - \boldsymbol{P}$, and it has eigenvalues in the range $[0,2]$, $\boldsymbol{P}$ has eigenvalues lying in the range $[-1,1]$.  

Now, observe that $\boldsymbol{P}$ shares the same eigenvectors with $\boldsymbol{P}^k$ for any $k$, however, eigenvalue $\lambda$ changes to $\lambda^k$. It can be seen that since the permutation is on the basis of sorting eigenvalues, the view for $\boldsymbol{A}+\boldsymbol{A}^2+\dots+\boldsymbol{A}^k$ will coincide with $\boldsymbol{A}$ if $f_k(x) = x+x^2+\dots+x^k$ is monotonic in the range $[-1,1]$, which is the range of allowed eigenvalues of the normalized adjacency matrix. It is trivial to note that $f_k(x)$ is monotonically increasing for $x \in [0,1]$ for any $k$. Now, ignoring the case $|x| = 1$, observe that $1 + f_k(x) = \frac{1-x^{k+1}}{1-x}$ by the geometric progression formula. If $k$ is odd, $k+1$ is even, and thus $x^{k+1}$ is positive for $x \in [-1,0]$. As we move $x$ from $-1$ to $0$ the numerator monotonically rises, and the denominator monotonically falls from $2$ to $1$, meaning that overall the function is monotonic and the ordering will just mirror $x$. This is not true when the sum terminates at an even power, for instance, $x+x^2$ which is $0$ at $-1$ and $0$ but negative at $-1/2$, indicating that it cannot be monotonic. The case where $x$ is $1$ or $-1$ is trivially true.

\subsection{Alignment closed forms}

Given a function of the following nature where $\boldsymbol{Q}$ is orthogonal

$$ ||\boldsymbol{XQ} - \boldsymbol{Y}||^2 $$

Minimization of the above function can be done by noting that this is equivalent to working with :

$$ ||\boldsymbol{XQ}||^2 + ||\boldsymbol{Y}||^2 - 2 \langle \boldsymbol{XQ}, \boldsymbol{Y} \rangle $$

$ ||\boldsymbol{XQ}|| = ||\boldsymbol{X}|| $, and we only have $\langle \boldsymbol{XQ}, \boldsymbol{Y} \rangle$ to maximize. This is $\boldsymbol{Y}^T \boldsymbol{X} Q$, which is equal to $\langle \boldsymbol{Q}, \boldsymbol{X}^T \boldsymbol{Y} \rangle$. Maximizing this boils down to the projection of the matrix $\boldsymbol{X}^T \boldsymbol{Y}$ on the set of orthogonal matrices under the square Frobenius norm. Let the SVD of $\boldsymbol{X}^T \boldsymbol{Y}$ be $\boldsymbol{USV}^T$, then we have $\langle \boldsymbol{Q}, \boldsymbol{USV}^T \rangle$ being minimized. This becomes $\langle \boldsymbol{U}^T \boldsymbol{Q V}, \boldsymbol{S} \rangle$, with $\boldsymbol{U}^T \boldsymbol{Q V}$ orthogonal and made to maximize inner product with diagonal $\boldsymbol{S}$, implying that $\boldsymbol{Q} = \boldsymbol{U V}^T$ is the solution.

\clearpage

\section{Proofs on the stochastic block model}

Usually, one divides graph contrastive learning and more generally all of contrastive learning into two categories :

\begin{itemize}
\item Supervised contrastive learning : There are unseen labels for each node. Our proofs will center on showing that with high probability, augmentations either preserve these unseen labels or some straightforward function of them that are made to suit the situation. For example, in a graph classification setting formed from the ego-graph of the nodes, the graph can be given the label of the node it is formed as an ego-graph from or the majority label of the nodes it contains. Our proofs in this case deal with the graph label and not the node label.
\item Unsupervised graph contrastive learning : Each node is its own class (the classic setting of graph contrastive learning). In this scenario, it is not possible to work with the node label. Since in our setting, the nodes also possess no node-specific attributes beyond the structural information, our work here must focus on the structures obtained under spectral augmentation only.
\end{itemize}

In both cases we assume contrastive learning in general to work. That is, we show that the process of generating positive pairs etc. continues properly, but not anything about whether contrastive learning as a whole can generalize better or learn better representations. We view such a proof as outside the scope of this paper.

In the paper, we have worked with six distinct augmentations , of which two modify the structures chosen : Crop and Similar/Diverse. Three of them modify the attributes alone : Mask, reorder, and align. In general, nothing can be proven about the latter three without assuming an easy function class such as linear classifiers, which we view as unrealistic. Hence, our work focuses on the first two.

Secondly, we work with the stochastic block model and the two-cluster case where differing label of a node indicates a different propensity to create edges. We only focus on the case where there are seen or unseen labels which are related to the structure of graphs. This can be seen as a scenario intermediate between the supervised and unsupervised contrastive learning case, and the block model a natural reification to study it, for it is well known~\citep{rohe2011spectral} that conditional on the label of a node being known, the degree structure and density etc. strongly concentrate around their fixed values for the stochastic block model. Indeed, no other parameter except the label which directly determines the edge density even exists to provide information about the structure. Proving that nodes of similar (seen or unseen) labels are brought together by our augmentations carries over to the unsupervised case fully as these parameters are the only ones directly determining the structure.

By assuming that (unseen) labels exist, our proof is quite relevant to the actual use case of the paper. This is because in the downstream usage, the classifier is used, zero-shot, to provide representations that are used to predict the label. In other words, hidden latent labels are assumed to be predictable from the structure. Our case should be understood as a special case of unsupervised representation learning that shares some conditions with the supervised scenario.

\subsection{Proof sketches and overall meaning}

Our proofs center around the stochastic block model. In this setting the spectrum is well known and analyzed. We show that in this case, the ``crop" operation around a node $v$ extracts a sub-graph of nodes which largely possess the same label as $v$ itself, where the label is considered to coincide with the cluster(block). Under the contrastive learning assumption, then, ``crop" recovers positive pairs.

We also show that common embedding methods such as LINE, DeepWalk etc. are meaningful in terms of establishing ``similar" and ``diverse" views in the stochastic block model and that ``similar" filtering would indeed yield a pathway to setting positive pairs apart. This is done by re-using our analysis for the supervised case which looks at the spectrum, and re-using the results from NETMF~\citep{qiu2018netfm} which connects the spectral results to embeddings obtained by random walks. In short, random walks and corresponding embeddings on stochastic block models can be seen, in the limit, as spectral decompositions of a block model with parameters that depend on the original block model. After this, we can recognize that the analysis for ``crop", which essentially shows that the spectral embeddings form a meaningful metric of closeness in terms of label, cluster etc. on the original model, fully carries over with transformed parameters.

\subsection{Supervised contrastive learning derivation}

We define our stochastic block model~\citep{rohe2011spectral} as follows in keeping with conventions of the field. We will consider one with two components.

\begin{itemize}
\item There are $N_0$ nodes generated with label $0$, and $N_1$ with label $1$. Denote the group generated with label $0$ as $C_0$ and the other as $C_1$. For simplicity, set $N_0 = N_1 = N$
\item An edge occurs with probability $p$ among two nodes of label $0$, with probability $q$ between two nodes of label $1$, and with $z$ among two nodes of different labels. $z < min (p,q)$ is a common assumption. Without loss of generality we can take $p > q > z$. We also consider the self-edges to be valid.
\end{itemize}

Note that in the setting of GCC, different local structures encode different labels. Hence $p \neq q$, as if they were equal it would imply the same structural generation process gives rise to two different labels.

Let $A$ be the adjacency matrix of the stochastic block model and $L$ the laplacian. Let $\lambda_n(v)$ be the function that assigns to a node $v$ its value under the $n$-th eigenvector of the Laplacian. Let $C_{\epsilon}(v)$ be the cropped local neighbourhood around any node $v$ defined as $\{ v' : ||\lambda(v')-\lambda(v)|| \leq \epsilon \}$ where $\lambda(v) = [\lambda_2(v), \lambda_3(v)]$.

\subsection{Factorizing the adjacency matrix}

The overall matrix $A$ is of shape $2N \times 2N$. Recall that we have allowed self-edges and diagonal entries of $A$ are not zero. Consider the matrix :

$$\begin{bmatrix}
p & z\\
z & q
\end{bmatrix}$$

Let $W$ be a $2N \times 2$ matrix formed by repeating the row vector $\begin{bmatrix}
1 & 0
\end{bmatrix}$ $N$ times and then the row vector $\begin{bmatrix}
0 & 1
\end{bmatrix}$ $N$ times. $W$ denotes the membership matrix. The first $N$ rows of $W$ denote that first $N$ nodes belong to label $0$ (and hence their zero-th entry is $1$) and the next $N$ rows likewise have a $1$ on their $2$-nd column signifying that they have label $1$.

Now, we can see that if we multiply $W \begin{bmatrix}
p & z\\
z & q
\end{bmatrix}$, the first $N$ rows of the resulting $2N \times 2$ matrix will be $\begin{bmatrix}
p & z
\end{bmatrix}$ and the next $N$ will be $\begin{bmatrix}
z & q
\end{bmatrix}$. Consider now multiplying from the right side with $W^T$ i.e. forming, overall,$W \begin{bmatrix}
p & z\\
z & q
\end{bmatrix} W^T$. This matrix will be of shape $2N \times 2N$ and it can be seen that it has a block structure of form :

$$ \begin{bmatrix}
p(1_N1_N^T) & z(1_N 1_N^T) \\
z (1_N 1_N^T) & q(1_N 1_N^T)
\end{bmatrix} $$

Where, $1_N$ is the $N \times 1$ vector of all $1$-s, and $1_N 1_N^T$ the $N \times N$ matrix of all $1$-s. So, it can be seen that the above matrix is nothing but the expectation of the stochastic block model's adjacency matrix.

Now, can we avoid analyzing this matrix and instead settle for analyzing the comparatively simpler $\begin{bmatrix}
p & z\\
z & q
\end{bmatrix}$ ? Let $v$ be an eigenvector of $\begin{bmatrix}
p & z\\
z & q
\end{bmatrix}$ with eigenvalue $\lambda$. Let $v$ have entries $\begin{bmatrix}
x\\
y
\end{bmatrix}$. By hypothesis,  $\begin{bmatrix}
p & z\\
z & q
\end{bmatrix} v = \lambda v$. Then, if we have the vector $v'$ of shape $2N \times 1$ with first $N$ entries as $x$, and the next $N$ as $y$

$$ \begin{bmatrix}
p(1_N1_N^T) & z(1_N 1_N^T) \\
z (1_N 1_N^T) & q(1_N 1_N^T)
\end{bmatrix}v' = (\lambda N)v' $$.

It can be seen that $v'$ is parallel to an eigenvector of the expectation of the adjacency matrix and the corresponding eigenvalue is $\lambda N$. However, we always assume eigenvectors are of unit norm, i.e. $x^2 + y^2 = 1$. So, $v'$ is going to be not $x$ repeated $N$ times, but $\frac{x}{\sqrt{N}}$ $N$ times and then $\frac{y}{\sqrt{N}}$ $N$ times. This makes $\|v'\| = 1 = \|v\|$. Therefore, for every pair $\lambda,v$ in the spectra of the $2 \times 2$ matrix, there is a corresponding $\lambda N, v'$ in the spectra of the expectation of the adjacency matrix. Next, note that the rank of the expectation of the adjacency is $\leq Rank(\begin{bmatrix}
p & z\\
z & q
\end{bmatrix}) \leq 2$. So if the $2 \times 2$ matrix has a full rank, there can be no extra eigenvalue-eigenvector pairs for the corresponding expectation of the adjacency matrix. All the nonzero eigenvalues and corresponding eigenvectors of the expectation of the adjacency are derivable from the $2 \times 2$ matrix's spectra. \textbf{In short, to understand the spectrum of the expectation of adjacency, we can just study the $2 \times 2$ matrix, as there is a one to one relationship between the nonzero eigenvalues and corresponding eigenvectors.}

\subsection{Crop augmentation}

\textbf{Notation of probability.} In proofs involving convergence, it is customary to provide a guarantee that a statement holds with probability $\geq 1 - F(N)$ where $F(N)$ tends to zero as $N$ goes to infinity. We will somewhat abuse the notation and say the probability $\rightarrow 1$ as $N \rightarrow \infty$ to denote this. We do not distinguish between things such as convergence in probability, convergence in distribution, almost surely convergence etc. and provide a largely concentration-inequality based proof overview.

We can state our proof for the \textbf{crop augmentation} as follows. All of the following statements hold with high probability (i.e. hold with a probability that $\rightarrow 1$ as $N \rightarrow \infty$.)

\begin{theorem}
Let the number of samples $N \rightarrow \infty$. Let $v$ be chosen uniformly at random from the nodes. The following statements hold with a probability that $\rightarrow 1$ as $N \rightarrow \infty$ :
\begin{itemize}
    \item \textbf{Proposition 1} : The majority label in $C_{\epsilon}(v)$ is the label of $v$.
    \item \textbf{Proposition 2} : Two nodes $v,v'$ of different labels generate non-isomorphic cropped subgraphs $C_{\epsilon}(v),C_{\epsilon}(v')$, if $v'$ is chosen uniformly at random as well.
    \item \textbf{Proposition 3} : For any $v$ and $C_{\epsilon}(v)$, there is no differently labeled $v'$ and $C_{\epsilon}(v')$ which is isomorphic to $C_{\epsilon}(v)$ for a high enough $\epsilon$, if both are chosen uniformly at random.
\end{itemize}

\end{theorem}

Note that in the main text, we state a slightly different version of the theorem (Theorem $1$ of the main text) involving ego networks as well, which we restate here :

\begin{theorem}
Let node $v$ be chosen uniformly at random from $G$, a $2N$-node graph generated according to the SBM described above.  With probability $\geq 1 - f(N)$ for a function $f(N) \rightarrow 0$ as $N \rightarrow \infty$, $\exists \epsilon \in \mathbb{R+}, k_{max} \in \mathbb{N}$ such that :
\begin{equation} 
\forall k \in \mathbb{N} \leq k_{max}, Y(E_{k,v}(G)) = Y(C_{\epsilon}(v)) = Y(v)
\end{equation}
\end{theorem}

Terming $k_{max}$ used above as $k_{crit}$, we can see that this is adding an extra part that agrees with the node label, namely the majority label of an ego network. However, the majority label for the $k$ ego network when $k \leq k_{crit}$ is trivially equal to the node's label for at least some values of $k_{crit}$ allowing $k$ beyond $0$ (i.e. the node itself is the ego network). To see this, take $k = k_{crit} = 1$. The expected number of nodes of the same label - considering label $0$ for simplicity - in its ego network is $p(N-1) + 1$, and the ones of a different label are of expected number $zN$. By Hoeffding's inequality, both quantities with high probability i.e. with probability $\geq 1 - O(\exp (-\delta^2))$ have deviations of only $O(\delta \sqrt{N})$ from their expectations which are terms of $O(N)$. Therefore, the majority label agrees with the node's own label with high probability (note that this requires $p,q > z$). We assume this step to hold with high probability and focus now on proving the equality of the cropped subgraph's majority label and the node label. We discuss ego networks other than the $1$-ego network at the end of the section. It can be easily checked that at least for the $1$-ego network, our proof involving mostly the vertex label case requires no changes.

To prove this, consider the random matrix $A$. We can denote the expectation of $A$ as $A*$. We can see the rank of $A*$ is $2$ as it has exactly 2 possible types of columns in it. It remains to find the corresponding two eigenvectors and eigenvalues. Clearly, by the structure of the matrix, the eigenvectors are of the form of repeating one value $c$ $N$ times and then another value $c'$ $N$ times, and we can WLOG set $c=1$ and replace $c'$ with $c$. Then it remains to solve

$$ \begin{bmatrix}
p & z\\
z & q
\end{bmatrix} \begin{bmatrix}
1\\
c
\end{bmatrix} $$

Which becomes, by the definition of eigenvector,

$$ \frac{z+ qc}{p +zc} = c $$

or, simplifying :

$$ zc^2 + (p-q)c - z = 0  $$

Therefore, by the quadratic formula :

$$ c = \frac{ (q-p) \pm \sqrt{ (p-q)^2 + 4z^2 } }{2z} $$

The eigenvalue for a corresponding $c$ is $p + zc$. Since $p,z \geq 0$ the larger eigenvalue $\lambda_1$ corresponds always to the larger value of $c$. Therefore, $\lambda_2$ takes the value gained by plugging in the negative root above, for c. This yields the \textbf{unnormalized} eigenvector, the actual entries assigned under the eigendecomposition are respectively $\frac{1}{\sqrt{1+c^2}},\frac{c}{\sqrt{1+c^2}}$ when a vertex $v$ is assigned its value under the eigenvector.



We re-use previous results in random matrix theory~\citep{vu2007spectral} (theorem $1.4$) that imply that with a probability $\rightarrow 1$, $||A-A*||_{op} \leq \sqrt{18pN}$, when $p$ is $\Omega(\log N)^4 / N$. Since the lower bound on $p \rightarrow 0$ as $N \rightarrow \infty$ we can assume it to hold for large $N$. The $\| \cdot \|_{op}$ notation denotes operator norm.

\textbf{Explanation of the order through Hoeffding and RIP property.} To intuitively understand the above result, we can consider each entry of $A-A*$. This is a random variable (blockwise) that takes one among the following set of values : $-p, 1-p$ (among the $N \times N$ entries of label $0$), $-z, 1-z$ (among the $2 \times N \times N$ entries between labels $0,1$) and $-q, 1-q$ (among label $1$). Now, only the upper triangle and diagonal are independent as the edges are symmetric, and the matrix is symmetric about the diagonal. We can write that :

$$ ||A-A*||^2_F = 2 ||A-A*||^2_{F,UT} + ||A-A*||^2_{F,D} $$

Where, $F$ denotes Frobenius norm, and $UT,D$ respectively denote summing over upper triangular and diagonal indices.

Therefore, if each entry of $A-A*$ be denoted as $\Delta_{i}$, enumerated in any order over the diagonal and upper triangle of the $2N \times 2N$ matrix ($1 \leq i \leq N(2N+1)$), $\Delta_{i},\Delta_{j}$ are independent r.v.s for any $i \neq j$. Further, $-1 \leq \Delta_i \leq 1$. Therefore, $0 \leq \Delta^2_i \leq 1$. Further, $||A-A*||_F \leq \sqrt {\sum 2 \Delta^2_i}$. 

$\sum \Delta^2_i$ is a sum of independent random variables in a fixed, finite range of size $1$. Therefore, Hoeffding's inequality applies, which yields that with probability $\geq 1 - O(1/N)$ :

$$ E(\sum \Delta^2_i) - \sqrt{ N (N + 1/2) \log (N)} \leq \sum \Delta^2_i \leq E(\sum \Delta^2_i) + \sqrt{N (N + 1/2) \log (N)} $$

$E(\sum \Delta^2_i)$ is the sum of the variances of random variables $\Delta_{i}$ over $N(2N+1)$ entries, bounded above by $1$. Hence, it follows that $||A-A*||^2_F$ is $O(N^2)$ with probability $\geq 1 - O(1/N)$, therefore, $||A-A*||_F$ is $O(N)$.

Further, we can see that $A-A*$ has the following structure :

$$ \begin{bmatrix}
J & K\\
K & L
\end{bmatrix} $$

Where, $J,L$ are symmetric matrices with each upper triangular and diagonal entry as i.i.d random variables $Z$ satisfying :

$$ Z = a \;\; \textrm{with probability p else} \; Z = b, E(Z) = 0 $$

In addition, $K$ is a matrix (not necessarily symmetric) which has every entry as i.i.d realizations of such $Z$. Then, such a matrix $\begin{bmatrix}
J & K\\
K & L
\end{bmatrix}$ by Restricted Isometry Property~\citep{vu2014modern} has the property that, with high probability, there is a constant $K'$ independent of $N$ :

$$ \max_i \lambda_i(A-A*) \approx \sqrt {\frac{K'}{N} \sum_{i=1}^{N} \lambda^2_i [(A-A*)]}  $$

The left hand side is the maximum eigenvalue, which we recognize as $\|A-A*\|_{op}$ i.e. the operator norm.

Now, using the relation between eigenvalues and the Frobenius norm :

$$ \sum_{i=1}^{N} \lambda^2_i [(A-A*)] = ||A-A*||^2_F =  O(N^2) $$

The RHS comes from plugging in the Frobenius norm bound from the Hoeffding's inequality step. Finally, this yields :

$$ ||A-A*||_{op} =  O(\sqrt{N}) $$

The result of Vu's we state above is merely a formalization of this sketch with constants, the order is the same i.e. $\sqrt{N}$.

Now consider the second eigenvector i.e. the $\lambda_2$ function from $A$ against the calculated $\lambda_2$ above for $A*$. We need to use the Davis Kahan theorem~\citep{demmel1997applied} (theorem $5.4$) which states that if the angle between these is $\theta$, then

$$ \sin 2 \theta \leq \frac{2 ||A - A*||_{op} } { 2N * \min (|\mu_1 - \mu_2|, \mu_2) } $$

As both eigenvectors are unit vectors, we can use the property that if two unit vectors have angle $\theta$ between them, the norm of their difference is bounded above by $\sqrt {2} sin 2 \theta$. Ignoring constants, we end up with the result that for some constant $c_0$, and denoting $v_{M,i}$ the $i$-th eigenvector of the Laplacian formed from some adjacency matrix $M$

$$ ||v_{A,2} - v_{A*,2} || \leq c_0 \frac{\sqrt{pN}}{N * \min (|\mu_1 - \mu_2|, \mu_2)} $$


The LHS however works with the eigenvector of the overall adjacency matrix formed by the multiplication by $W$ we have discussed above. Recall that we have already noted the adjacency matrix and the $2 \times 2$ matrix share eigenvalues upto scaling in $N$. Their eigenvectors are also likewise related, and since eigenvectors are always of unit norm, an eigenvector of the $2 \times 2$ matrix is first repeated in its entries and then normalized by a factor of $\frac{1}{\sqrt{N}}$ by virtue of being an eigenvector, to become an eigenvector of the overall adjacency matrix. 

By substituting the eigenvectors we found earlier, i.e. $\frac{1}{\sqrt{1+c^2}} [1,c]^T$ into the LHS, scaling by this $\sqrt{N}$ factor cancels the extra $\sqrt{N}$ on the RHS. Recalling that

$$c = q_1 =  \frac{ (q-p) - \sqrt{ (p-q)^2 + 4z^2 } }{2z}$$

Let :

$$ S_1 : \{ x : v_{A*,2}(x) = \frac{1}{\sqrt{1+q_1^2}},  v_{A,2}(x) \leq \frac{1+q_1}{2\sqrt{1+q_1^2}} + \frac{\epsilon}{2}  \}  $$

$$ S_2 : \{ x : v_{A*,2}(x) = \frac{q_1}{\sqrt{1+q_1^2}},  v_{A,2}(x) \geq \frac{1+q_1}{2\sqrt{1+q_1^2}} - \frac{\epsilon}{2} \}  $$

Let $v_1,v_2$ be any pair of nodes that satisfy the conditions of :

\begin{itemize}
\item Labels of $v_1,v_2$ are different. WLOG, let $v_1$ have label $0$, $v_2$ label $1$.
\item $C_\epsilon(v_1)$ contains $v_2$ (and by symmetry, $C_\epsilon(v_2)$ contains $v_1$)
\end{itemize}

Clearly, we can see that either $v_1 \in S_1$ or $v_2 \in S_2$. Let $K_1 = |S_1|, K_2 = |S_2|$. Summing only over $S_1$ :

$$ K (\frac{1-q_1}{2\sqrt{1+q_1^2}}-\epsilon/2)^2 \leq \frac{p(c_0)^2}{ \min ((\mu_1 - \mu_2)^2, \mu^2_2) }  $$

This yields a bound on $K_1$, which we can term $K_{1,max} = \frac{p(c_0)^2}{ (\frac{1-q_1}{2\sqrt{1+q_1^2}}-\epsilon/2)^2 \times \min ((\mu_1 - \mu_2)^2, \mu^2_2) }$. Similarly, we can consider $K_2$ to get $K_{2,max}$ as : $\frac{p(c_0)^2}{ (\frac{1-q_1}{2\sqrt{1+q_1^2}}-\epsilon/2)^2 \times \min ((\mu_1 - \mu_2)^2, \mu^2_2) }$. Set $\epsilon = \frac{1-q_1}{2\sqrt{1+q_1^2}}$ to get :

$$ K_{1,max} + K_{2,max} \leq  \frac{2p(c_0)^2}{ (\frac{1-q_1}{4\sqrt{1+q_1^2}})^2 \times \min ((\mu_1 - \mu_2)^2, \mu^2_2) }$$

Since $K_{1,max} = K_{2,max}$ we can term it $K_{max}$. Simultaneously, consider :

$$ S_3 : \{ x : v_{A*,2}(x) = \frac{1}{\sqrt{1+q_1^2}},  v_{A,2}(x) \geq \frac{1+q_1}{2\sqrt{1+q_1^2}} + \frac{3\epsilon}{2}  \} , \epsilon = \frac{1-q_1}{2\sqrt{1+q_1^2}} $$

$$ \frac{1+q_1}{2\sqrt{1+q_1^2}} + \frac{3\epsilon}{2} = \frac{\frac{5}{4}-\frac{q_1}{4}}{\sqrt{1+q_1^2}} \geq \frac{1}{\sqrt{1+q_1^2}} + \frac{1}{4\sqrt{1+q_1^2}} $$

Last inequality is by the property $q_1 < 0$. By a similar argument as for $S_1,S_2$, $S_3$ is of constant size and $\frac{S_3}{N} \rightarrow 0$ as $N \rightarrow \infty$. Let the maximum size of $S_3$ be $K_{3,max}$. Now, if we pick a vertex $v$ of label $0$ at random, with probability $\geq 1 - \frac{K_{3,max}+K_{max}}{N}$, $v \notin S_1, v \notin S_3$. If both these conditions hold, in $C_\epsilon(v)$ any $v'$ which does not have the same label must have $v' \in S_2$. (Because any such pair must have at least one element in $S_1,S_2$ and $v \notin S_1$). Simultaneously, $C_\epsilon(v)$ contains at least all vertices of label $0$ not in $S_1 \bigcup S_3$ i.e. has vertices of label $0$ $\geq N - K_{3,max} - K_{max}$. Noting that $K$ values are all constants and $N \rightarrow \infty$ implies that majority label in $C_{\epsilon}(v)$ will agree with $v$ as $K_{max}/N \rightarrow 0$ completes the proof. The only aspect of the proof which required high probability was the norm of $||A-A*||_{op}$ varying as $\sqrt{N}$, this step may be assumed to be true with probability $\geq 1 - O(1/N^3)$ (tighter bounds are possible but this suffices). \textbf{This concludes the proof of proposition 1.}

\textbf{Tightness of operator norm.} Consider the statement that :

$$ ||A-A*||^2_F = O(N^2) $$

Recall that we showed :

$$ E(\sum \Delta^2_i) - \sqrt{ N (N + 1/2) \log (N)} \leq \sum \Delta^2_i \leq E(\sum \Delta^2_i) + \sqrt{N (N + 1/2) \log (N)} $$

With high probability. We used the bound of the RHS, but the bound of the LHS is also true. Hence, $ ||A-A*||^2_F = \Theta(N^2) $.

Next, we used :

$$ \max_i \lambda_i(A-A*) \approx \sqrt {\frac{K'}{N} \sum_{i=1}^{N} \lambda^2_i [(A-A*)]}  $$

However : 

$$ \max_i \lambda_i(A-A*) \geq \sqrt {\frac{1}{N} \sum_{i=1}^{N} \lambda^2_i [(A-A*)]}  $$

Therefore, $\max_i \lambda_i(A-A*) = ||A-A*||_{op} = O(\sqrt{N})$.

In short, every step till we apply the Davis-Kahan bound, i.e. :

$$ ||v_{A,2} - v_{A*,2} || \leq c_0 \frac{\sqrt{pN}}{N * \min (|\mu_1 - \mu_2|, \mu_2)} $$

Is as tight as possible.

\textbf{Tightness of Davis-Kahan bound.} The Davis-Kahan upper bound is sharp. That is, $\exists S, H$, $S = S^T, H = H^T$, with $\mu_1 \geq \mu_2 \geq\dots.. \mu_N $ the eigenvalues of $S$, $v_1,v_2, \dots, v_N$ the corresponding eigenvectors of $S$, $v'_1,v'_2,\dots, v_N$ the eigenvectors of $S+H$, $\theta_i$ the angle between $v_i,v'_i$ satisfying :

$$ \sin (2 \theta_i) = c'\frac{2||H||_{op}}{\min_{j \neq i} |\mu_i - \mu_j| } $$

Where $c'$ is a constant $\leq 1$ that does not depend on N. And at the same time, $\forall S,H,\mu_i,v_i,v'_i$ :

$$ \sin (2 \theta_i) \leq \frac{2||H||_{op}}{\min_{j \neq i} |\mu_i - \mu_j| } $$

However, in our case, $S$ is not arbitrary, but $S=A*$.  When we take $\exists S$, it allows taking e.g. $S = \begin{bmatrix}
0.6 & 0.8\\
0.8 & 0.7
\end{bmatrix}$. But this cannot be $A*$ with $N=1$, as it violates all our assumptions for $A*$, here $p=0.6, q=0.7, z=0.8$ violating $p>q>z$ assumptions. We need to show that $\exists S,N,H$ such that $H=H^T$ and :

$$ S = WS'W^T \;, S' = \begin{bmatrix}
p & z\\
z & q
\end{bmatrix}, 0 \leq z \leq q \leq p \leq 1 $$

with $W$ of shape  $2N \times 2$ such that first $N$ rows of $W$ are $[1,0]$, next $N$ are $[1,0]$. This makes $S=S^T$ and we already constrain $H=H^T$. With $\mu_1 \geq \mu_2 \geq\dots.. \mu_{2N} $ as eigenvalues of $S$, $v_1,v_2, \dots, v_{2N}$,$v'_1,v'_2,\dots, v_{2N}$ corresponding eigenvectors of $S,S+H$ we must show $\exists i$

$$ \sin (2 \theta_i) = c" \frac{2||H||_{op}}{\min_{j \neq i} |\mu_i - \mu_j| } $$

Where $c"$ is constant, not a function of $N$. This is reached at :

$$ S' = \begin{bmatrix}
0.6 & 0\\
0 & 0.4
\end{bmatrix}, H' = \begin{bmatrix}
-0.1 & 0.1\\
0.1 & 0.1
\end{bmatrix} $$

$$ S = WS'W^T, H = WH'W^T $$

Where $W$ is as specified and of shape $2N \times 2$.

\textbf{Proof of proposition two.}

Recall that by the proof of proposition one, $C_\epsilon(v)$ contains $M$ nodes of label $0$, where if $v$ is selected randomly over all nodes with label $0$, with probability $\rightarrow 1$, $\frac{M}{N} \rightarrow 1$. This step is with probability $\geq 1 - O(1/N^3)$.

Let $E_0$ be the set of all edges $(v_i,v_j)$ with $v_i,v_j$ having label both labels $0$. Let $M_(0,v)$ be the set of all edges $(v_i,v_j)$ s.t. $(v_i,v_j) \in E_0, v_i,v_j \in C_\epsilon(v)$. Clearly, $M_{0,v} \subseteq E_0$, and since $\frac{M}{N} \rightarrow 1$, $\frac{|M_{0,v}|}{|E_0|} \rightarrow 1$. 

By a similar argument, let $M_{1,v'}, E_1$ be the corresponding edge sets for $C_\epsilon(v')$ with label of $v'$ being $1$,  $\frac{|M_{1,v'}|}{|E_1|} \rightarrow 1$.

$E_0,E_1$ are, denoting $B_w(p)$ as an independent Bernoulli random variable of bias p :

$$ E_0 = \sum_{w=1}^{N(N+1)/2} B_w(p), E_0 = \sum_{w=1}^{N(N+1)/2} B_w(q) $$

Via Hoeffding's inequality, $E_0 = (pN(N+1)/2) + O(N\sqrt{\log N}), E_1 = (qN(N+1)/2) + O(N\sqrt{\log N})$ with probability $\geq 1 - O(1/N^3)$. Therefore, with probability $\geq 1 - O(1/N^3)$, $C_\epsilon(v),C_\epsilon(v')$ are not isomorphic. \textbf{This proves proposition two}. Applying the union bound over all choices of $(v,v')$ \textbf{proves proposition three}, because number of possible pairs is $O(N^2)$ and the property holds with $\geq 1 - O(1/N^3)$, leading to overall probability $\geq 1 - O(1/N)$. 

Note that the step of the $1$ ego-network's majority label agreeing with the node's label was derived by Hoeffding's inequality and does not affect any order used so far. \textbf{Hence, this completes the proof of propositions $1,2,3$.}

\subsection{Embedding-based similarity}

First, we remind the reader that usually, each node embedding method such as LINE~\citep{tang2015line} always normalizes its embedding per node. That is, each node $v$ receives some vector $e_v$ with $\| e_v \| = 1$. But that, in turn implies that given two distinct embeddings $e_v, e_{v'}$,

$$ \frac{\langle e_v, e_{v'} \rangle }{ \| e_v \| \|e_{v'} \| } = \langle e_v, e_{v'} \rangle $$

$$ \| e_v - e_{v'} \|^2 = \|e_v \|^2 + \|e_{v'} \|^2 - 2 \langle e_v, e_{v'} \rangle = 2 -  2 \langle e_v, e_{v'} \rangle$$

That is, selecting on the more similar cosine distance (similar filtering) is equivalent to selecting on lower values of $\| e_v - e_{v'} \|$ - the type of proximity analyzed in crop. This simplifies our analysis, allowing re-use of crop results.

In the context of embeddings, let us analyze two in particular : DeepWalk~\citep{perozzi2014deepwalk} and LINE~\citep{tang2015line}. It is known previously from the analysis of NETMF~\citep{qiu2018netfm} that these methods replicate matrix factorization. Specifically, let $A,D$ be the adjacency matrices and degree matrices, then :

$$ P_r = \frac{1}{T} (\sum_{r=1}^{T} (D^{-1} A)^r) D^{-1} $$

Let the volume of a graph $G$ $V(G)$ be the sum of the number of edges, then $\log (V(G) P_r) - \log b$ where $b$ is the negative sampling rate is a matrix $Z_r$. Under the framing above, DeepWalk factors any $Z_r$, while LINE factors $Z_1$, i.e. LINE is a special case of DeepWalk. This $\log$ is taken per element.

Now, we are ready to state our theorems for similarity. Let $E(v)$ be the embedding assigned to a node $v$. Let $p,q,z$ be as before.

\begin{theorem}

Let the number of samples $N \rightarrow \infty, \frac{p}{q} \neq \infty$, $\exists \epsilon_{crit}$ such that $\forall v$, let $\{S_v : v', \textrm{label}(v') = \textrm{label}(v), \|E(v) - E(v')\| \leq \epsilon_{crit}$ \}, then, $\frac{|S_v|}{N} \rightarrow 1$ and $C_{\epsilon_{crit}}(v)$ satisfies the three propositions of theorem $1$, when $v$ is chosen uniformly at random.

\end{theorem}

Our proof for this will first translate the graph adjacency to familiar matrix forms. Note that $V(G)$ is equal to, in expectation and allowing self-loops, as :

$$ 2N^2(p/2 + q/2 + z) $$

Further, we can set $b=1$ to remove it from consideration. Set $r=1$ to recreate the case of LINE. Now, by a similar argument as for the cropping analysis for SBM where the matrix $W$ generates the $A*$ matrix using a $2 \times 2$ matrix, we can examine the $2 \times 2$ matrix which is :

$$ \begin{bmatrix}
\frac{p}{(p+z)^2} & \frac{z}{(p+z)(q+z)}\\
 \frac{z}{(p+z)(q+z)} & \frac{q}{(q+z)^2}
\end{bmatrix} $$

Now, let us apply the logarithm to get our new values of $p',q',z'$ which will be fed back to crop analysis and behave equivalently to the original parameters (as before in the crop case, the $N^2$ term in $V(G)$ can be ignored while reducing to the $2 \times 2$ case) :

$$ p' = \log p - 2 \log(p+z)  + \log(p+q+2z) $$

$$ q' = \log q - 2 \log(q+z)  + \log(p+q+2z) $$

$$ z' = \log z - \log(p+z) - \log(q+z)  + \log(p+q+2z) $$

Note that $p > q$ implies $p' < q'$, as :

$$ p' - q' = \log p - \log q + 2 \log (q+z) - 2 \log (p+z) $$

$$ p(q+z)^2 - q(p+z)^2 = pq^2 + pz^2 - qp^2 - qz^2 = (p-q)(z^2 - pq) < 0 $$

And, $p > z$ implies $p' > z'$, as :

$$ p' - z' = \log p - \log (p+z) + \log (q+z) - \log z $$

$$ p(q+z) - z(p+z) = pq - z^2 > 0 $$

Therefore, the analysis from the crop case carries over, except we swap the order of $p,q$. This is equivalent to swapping the labels of the nodes, and makes no difference. Recall that the error rate for the spectral analysis in the crop section depends on $p'$ (numerator) and $\min (\mu_2, | \mu_1 - \mu_2 |)$ (denominator) $\implies$ : the error will remain bounded above iff : $|p'|, |q'|$ are bounded above and $\min (\mu_2, |\mu_1 - \mu_2|)$ are bounded below.

\textbf{Cases of issues in $p', q'$.} Can be ruled out as follows :

\begin{itemize}
\item $p' = \log p - 2 \log (p+z) + \log (p+q+2z) \leq \log p - 2 \log p + \log 4p = \log 4$.

\item $q' \leq \log q - 2 \log q + \log (4p) = \log 4 + \log (p/q) < \infty$ by hypothesis $p/q \neq \infty$.

\end{itemize}

\textbf{Case of eigenvalue issues, i.e.} : $\min (\mu_2, |\mu_1 - \mu_2|) \rightarrow 0 \implies \mu_2 \rightarrow 0$, $\mu_1 - \mu_2 \rightarrow 0$. From the analysis of crop, the eigenvalues are of the form $p' + z'c'$ where :

$$ c' = \frac{ (q'-p') \pm \sqrt{ (p'-q')^2 + 4z'^2 } }{2z'} $$

Note that $z'$ is $ < 0$. To see this, observe that :

$$ (p+q+2z)z = 2z^2 + pz + qz = (p+z)(q+z) - pq + z^2 < (p+z)(q+z) $$

Simultaneously, $p' > 0$, as :

$$ p(p+q+2z) - (p+z)^2 = pq - z^2  $$

So the eigenvalues are : 

$$ \mu_1 = \frac{(q' + p') + \sqrt{ (p'-q')^2 + 4z'^2 }}{2}, \mu_2 = \frac{(q' + p') - \sqrt{ (p'-q')^2 + 4z'^2 }}{2}  $$

$$ \min {\mu_2, |\mu_2 - \mu_1| } = 0 \iff z' \rightarrow 0 $$

$$ 0 > z' = \log (pz + qz +2z^2) - \log (pq + pz + qz + z^2) = \log (1 - \frac{pq -z^2 }{pq + z^2 + pz + qz})  $$

Since $p> q>z, pq -z^2 > 0, z' \rightarrow 0$ is not possible, and we are done. We simply need to adjust for the final step of vertex-wise normalization. Since in this case, $\mu_1$ and its corresponding eigenvector is also utilized, we have that all elements of cluster $0$ receive embeddings of form (recall $q_1,q_2$ are the roots of the quadratic for $c$) $[\frac{1}{\sqrt{N(1+q_1^2)}}, \frac{1}{\sqrt{N(1+q_2^2)}}]$ pre-normalization, and upon vertex-wise normalization this becomes $[\frac{\sqrt{1+q_2^2}}{\sqrt{2 + q_1^2 +q_2^2}}, \frac{\sqrt{1+q_1^2}}{\sqrt{2 + q_1^2 +q_2^2}}] $. For cluster $2$, the corresponding post-normalization embeddings are $[\frac{q_1\sqrt{1+q_2^2}}{\sqrt{q_1^2 + q_2^2 + 2q_1^2 q_2^2}}, \frac{q_2\sqrt{1+q_1^2}}{\sqrt{q_1^2 + q_2^2 + 2q_1^2 q_2^2}}] $. These are both on expectation and all the gaps in expectation remain $O(N)$ with deviations of $O(\sqrt{N})$. Re-applying the results from the crop section, we get that there is $\epsilon_{crit}$ such that with probability $\geq 1 - O(1/N^3)$ with a fixed $v$, a fraction $v' \rightarrow 1$ sharing the label of $v$ lies within $\epsilon_{crit}$, while at most $O(1)$ (the $K_{max}$ terms derived earlier) do not fall within this $\epsilon_{crit}$ or are $v''$ not sharing the label of $v$ but lying within $\epsilon_{crit}$. Most importantly, $q_1,q_2$ differ in sign, making the inner product consist of two positive terms between nodes of same label with high probability and one positive and one negative term between nodes of differing labels - the inner product is now a meaningful similarity metric ! For an easy example, we can see that setting $p=q$ leads to (in expectation) embeddings $[\frac{1}{\sqrt{2}},\frac{1}{\sqrt{2}}]$ for one label/cluster and $[\frac{1}{\sqrt{2}},\frac{-1}{\sqrt{2}}]$ for the other. This means nodes of two differing labels have $\approx 0$ inner product and two of the same have inner product $\approx 1$ with high probability.

Clearly, in the SBM case, \textbf{similar filtering} is the correct course of action with the threshold being this $\epsilon_{crit}$. It is plausible that in other generative graph cases, this would not be the case. However, in our experiments, \textbf{similar} was always superior to diverse filtering, possibly reflecting that real life graphs are well-modeled by SBMs in this aspect. Note also that our graphs such as LiveJournal, FaceBook etc. arrive from communities in social networks which may be expected to display a clustered / stochastic block model type of pattern. Note that the factorization under the transformed parameters is not necessarily a descending algebraic decomposition, but one where we perform the decomposition in order of magnitude.

\subsection{Ignorable perturbation effect from third eigenvector}

In both the proofs, we have examined only $2$-component SBMs. In these cases, $\mu_3$ and its corresponding eigenvector plays no role and indeed we have proven everything in terms of $\mu_2,\lambda_2$ alone. This is because in the expected adjacency matrix, $\mu_3 = 0$. 

The proof fully extends to the case where $\lambda_3$ is added. For simplicity, we did not add it, and the only change required is that for every node, we instead use $\mu_2 \lambda_2$ instead of $\lambda_2$, and $ \mu_3 \lambda_3 $ instead of $\lambda_3 $. Since $\mu_3$ in the original expected adjacency ($A*$) is zero, it is solely from $A-A*$ that $\mu_3$ arises. By Weyl's eigenvalue perturbation theorem~\citep{weyl1912asymptotische}, 

$$\lambda_3(A-A*) \leq \mu_3 \leq \lambda_1(A-A*) $$

We know, however, that $A-A*$ is a RIP matrix~\citep{vu2014modern} and thus $\mu_3 \approx \sqrt{N}$. This is with respect to the full matrix, i.e. $W \begin{bmatrix}
p & z\\
z & q
\end{bmatrix} W^T$ being $A$. This has $\mu_2$ as $O(N)$, making $\mu_3$ and thus $\mu_3 v_3$ only $\frac{1}{\sqrt{N}}$ relative to the other terms, and thus ignorable in the final embedding.

\subsection{The case of DeepWalk}

We sketch here the proof extension of the LINE case to DeepWalk. In DeepWalk, the matrix $D^{-1} A$ is replaced with the average over the first $T$ powers, that is :

$$ \frac{1}{T} \sum_{r=1}^{T} (D^{-1} A)^r $$

First note a few things. $A$ is the adjacency matrix which we know to be of form (in expectation) as $WBW^T$ where $B$ is $2 \times 2$. $A^2$ is $WBW^TWBW^T$. But, $W^TW$ is a scaled identity matrix. Thus $A^2$ in expectation is $WB^2W^T$ (times factors purely in $N$). Ergo, the analysis can still be carried out in terms of $B$, only this time using $B^2$. Next, note that replacing $A$ with any sum of powers of $K$ does not change the eigenvectors, it only changes the eigenvalues, because :

$$ M = USU^T \rightarrow M + M^2 \dots M^K = U(S+S^2 \dots + S^K)U^T $$

The final right multiplication with $D^{-1}$ does not affect this conclusion, since right multiplication with any diagonal matrix simply changes eigenvectors by inverse of the said matrix. Since the eigenvectors remain the same, all the steps crop onward to filtering continue to function, but the scaling factors might change due to eigenvalue changes. Since the eigenvalue change does not change finitely large quantities to infinitely small quantities and we only use this step to rule out noise from the third eigenvector, which after normalization contributes a term of order $O(\frac{1}{\sqrt{N}})$ relative to the other terms, all the steps continue to work.

\textbf{Caveats and extensions.} In the following subsections, we check some alternate scenarios of other ways to do NETMF, $\geq 3$ components, and most importantly how random walk augmentations and ego networks of distances $\geq 2$ may shift our analysis. These subsections should be considered as extensions of the main proof and mostly expository.

\subsection{NETMF rounding vs no rounding.} In the NETMF implementation, terms $< 1$ before taking the log can be rounded up to $1$ in an alternate usage. This case either results in no off diagonal entries after the log (making the analysis trivial as it is the adjacency matrix of two disconnected blocks) or a zero matrix making it nonsensical. Thus, we did not analyze this case, as in this case our augmentations either trivially work or no sensible embedding is produced at all due to a matrix of all zeroes.

\subsection{Extension to 3 and greater components}

In the cases of $\geq 3$ component SBMs, the eigenvectors are significantly more complicated than for a $2 \times 2$ case. However, the consistency of spectral clustering under the $L_2$ norm - which is, as one might recognize, what we have shown here with some modifications - is proven in alternate ways, and convergence is guaranteed so long as the number of eigenvectors used is $k$ and equal to the number of components. However, these proofs carry over to the case where the eigenvalues after $3$ satisfy magnitude constraints~\citep{rohe2011spectral,sarkar2015role,von2007tutorial}, and also for product graphs such as the grid graph (shown in main text).  Therefore, using eigenvectors upto $\lambda_3$ would suffice in these low rank cases even if the overall number of components was high.

\subsection{Notes on the random walk augmentation in the SBM scenario}

The random walk step in GCC~\citep{qiu2020gcc} is essential for the purposes of scaling Graph Contrastive methods. This is because in most cases, the ego-networks obtained by taking the simple neighbours within $d$ steps of a node $v$ are too large (over $1000$) whereas a random walk on these ego networks, and then collecting the nodes visited, yields much smaller graphs ($\leq 256$ for our implementation, with averages much lower, $<100$). This naturally leads us to ask if this step is only required for scalability - does it also have other desirable properties ?

In the stochastic block model, at least, it does. Consider the adjacency matrix of a $2$-component SBM as $A$, with $N$ nodes each of two classes. Let a node be $v$, and keep inter-connection probabilities as $p,q,z$.

For any $p,q,z$ which are not arbitrarily low, i.e. do not $\rightarrow 0$ as $N \rightarrow 1$, it can be seen that any $k$-ego network of $v$ for $k \geq 2$ covers a fraction of nodes $\rightarrow 1$ of the entire SBM. This can be understood by considering any node $v' \neq v$. There will be no path $v' \rightarrow z \rightarrow v$ (and in the other direction - we are dealing with undirected graphs) with $z \neq v, v'$ iff $\forall z$ there is either no edge $(z,v)$ or no edge $(z, v')$.

The probability of such a path existing for a particular $z$ is $ \geq p_{min} = \min \{ p^2, q^2, z^2, pz, qz, pq \}$. Therefore, any $z$ will not have such a path with a probability $ \leq 1-p_{min}$. And, since there are $O(N)$ independent choices of $z$, such a path will \textbf{not} exist between $v,v'$ through such a $z$ with a probability $\prod_{i=1}^{O(N)} (1-p_{min})$. Thus, with probability $\rightarrow 1$ as $N \rightarrow \infty$, all such $v,v'$ have a path between them of length $2$. This implies all but $1$-ego networks are unsuitable as they will include such paths and cover almost the entire graph.

The probability of a random walk, on the other hand, should be analyzed as follows. While it is tempting to consider the transition matrix (the adjacency matrix normalized by degree) for analysis and take its limit (as it enjoys well-understood properties), the random walks utilized in practice have a very high return probability to the origin ($0.8$). This implies, in turn, that we should only consider the random walk lengths of low step number, as the probability of visiting a node even at distance $3$ is $0.2 \times 0.2 = 0.04$. Over a typical random walk of transition length $256$, only $10$ nodes at distance $\geq 3$ occur.

With this in mind, consider a random walk starting WLOG from a node of class $0$. Now :

\begin{itemize}
    \item At walk length $1$, the random walk has as neighbours, on expectation, $p(N-1)$ nodes of class $0$ and $zN$ nodes of class $1$.
    \item At the beginning of walk step $2$, there is a roughly $\approx \frac{p}{p+z}$ probability the random walk is at label $0$, and $\frac{z}{p+z}$ that it is at label $1$. The corresponding probabilities at the end are : $\frac{p^2}{(p+z)^2} + \frac{z^2}{(z+q)(p+z)}$ for class $0$ and $\frac{zq}{(z+q)(p+z)} + \frac{pz}{(p+z)^2}$ for class $1$. 
\end{itemize}

Compared to blindly taking the $2$-ego network, this can be seen to notably bias the the network in favor of the first class, by a ratio equal to :

$$ \frac{p^2(z+q) + z^2(p+z)} { zq(p+z) + pz(z+q) } = \frac{ p^2z + p^2q + z^2p + z^3 } { pqz + z^2q + pz^2 + pqz } $$

To see the numerator is greater, observe that $p^2z + z^3 \geq 2pz^2 \geq pz^2 + qz^2$ (AM-GM). This leaves us with proving that $p^2q + z^2p \geq 2pqz$. However, $p^2q + z^2p \geq pq^2 + z^2p \geq 2pqz$ (AM-GM). Therefore, unlike the $2$-ego network case which virtually has the same number of nodes of either class with high probability, the random walk slants the node label ratio, as desired, in the $2$-nd step (it trivially does so in the first step simply due to the $1$-ego network case and this case has no interesting differences between the random walk and directly taking the ego network).

Ergo, the random walk may help offset nonsense nodes included in the ego-network, at least in the block model setting. The fact it is run with a high return probability aids this - were it allowed to run longer, it would approach its mixing time and be closer to uniform in its probability of being found over nodes of either class.

\subsection{Larger ego networks}

In this section, we have implicitly considered $1$-ego networks when taking the full ego network into consideration. It is clear from our analysis in the random walk section that $2$-ego networks or higher can only become feasible as at least one of $p,q,z$ go to zero as $N$ goes to infinity. Clearly, we cannot have $z$ remain finitely large while $p,q$ go to zero (as this violates our assumptions) and so either :

\begin{itemize}
    \item all of $p,q,z$ go to zero
    \item $p,q$ stay finite, and $z$ goes to zero.
\end{itemize}

Case $1$ is much harder to tackle. For instance, our argument re : the Frobenius deviation in the adjacency matrix assumes that the expected adjacency matrix has a Frobenius norm of order $O(N^2)$. This may not be true when $p,q,z$ are allowed to be infinitely small.

Instead, let $p,q$ remain finite and $z \rightarrow 0$ as $N \rightarrow \infty$. Assume that the graph remains connected with high probability. This is true when, for instance, $z = 1/N$. The number of edges on expectation cross cluster is still $O(N)$. Observe that the crop proof we have used continues to work assuming no numerical problems. This is because the eigenvalues use $p+zc$, and $c$ inversely varies as $z$. There are no infinities incurred as a result of $z$ except for $c$ itself. The value of $c \rightarrow \infty$, implying that post-normalization of eigenvectors, the cluster of label $0$ receives embeddings which approach $[0, 0]$, and the cluster of label $1$ receives embeddings which approach $[1,-1]$.  The result also thus carries over to the similar embedding proof, which re-uses the crop result. In this particular case, $2$ and higher ego networks are viable, and depend on the value of $z$. For example, if $z = O(1/N)$, each node $v$ in the cluster of label $0$ has $O(1)$ neighbours in cluster $2$, and $qN * O(1)$ neighbours at distance $2$ of label $1$, allowing $2$-ego networks ($3$-ego networks fail in this case). We did not analyze such scenarios in depth, but we put the $k_{max}$ in our theorem to allow such cases.

\clearpage

\section{Time complexity analysis}

First, we present the ego graph sizes as a function of average degree of the graphs and also as functions of the overall graph size. Then, we try to plot the time taken to process each graph instance as a function of these parameters.

\begin{minipage}[b]{0.5\textwidth}

    \centering
    \includegraphics[width=7.5cm]{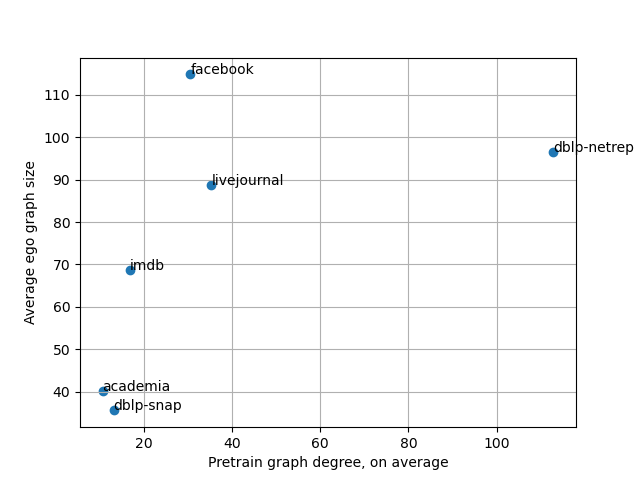}
    \captionof{figure}{Average ego graph size vs pretrain graph degree.}
    \label{fig:egoanddegree}

\end{minipage}
\hfill
  \begin{minipage}[b]{0.5\textwidth}

    \centering
    \includegraphics[width=7.5cm]{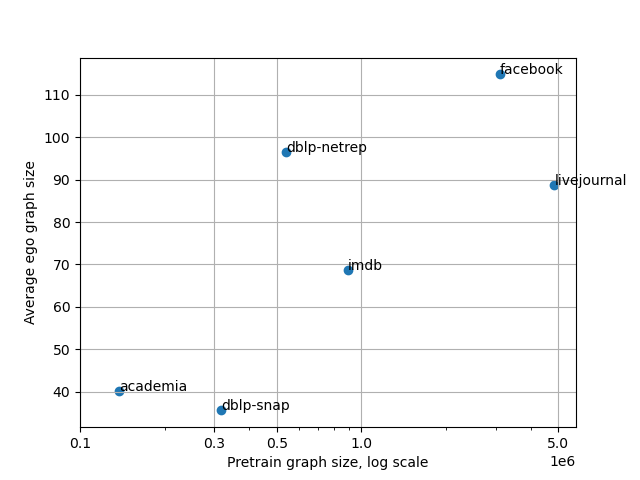}
    \captionof{figure}{Ego graph size vs graph size}
    \label{fig:egoandsize}
\end{minipage}
\hfill

\begin{minipage}[b]{0.5\textwidth}

    \centering
    \includegraphics[width=7cm]{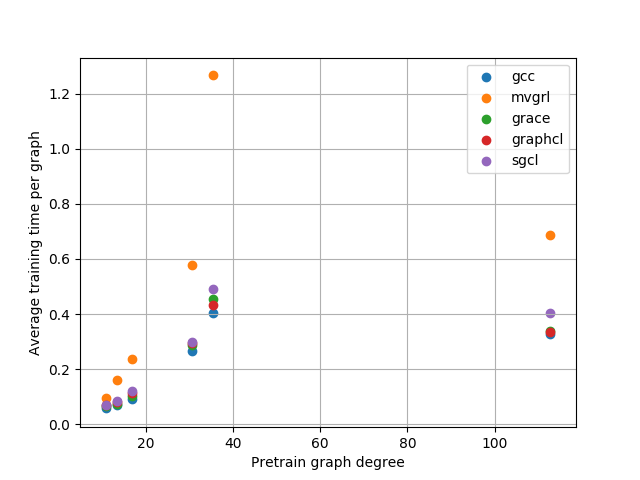}
    \captionof{figure}{Time per graph vs average degree}
    \label{fig:timeanddegree}

\end{minipage}
\hfill
  \begin{minipage}[b]{0.5\textwidth}

    \centering
    \includegraphics[width=8cm]{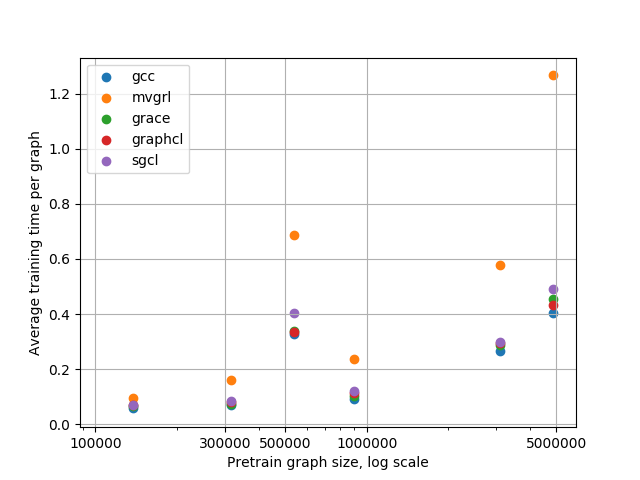}
    \captionof{figure}{Time per graph vs graph size}
    \label{fig:timeandsize}
\end{minipage}
\hfill

\clearpage

\section{Negative transfer effects}

In some of our datasets, there is a noted \textbf{negative transfer effect}. What we mean by this is that further training actually decreases the performance on the dataset. One should be wary of this when looking at result pairs where, for instance, the pre-trained model performs worse than a non pre-trained model or a fresh model. 

We repeat that the goal of pre-training is to come up with a general model that is trying to excel at all tasks, simultaneously - it is optimizing an average over all tasks. Optimizing such an average may come at the cost of a particular task. We illustrate this effect with IMDB-BINARY. We show $3$ consecutive results on this dataset, E2E Frozen, at $5,10,20$ epochs of training on DBLP. The results actually progressively worsen. 

\begin{figure}[h]
    \centering
    \includegraphics[width=6cm]{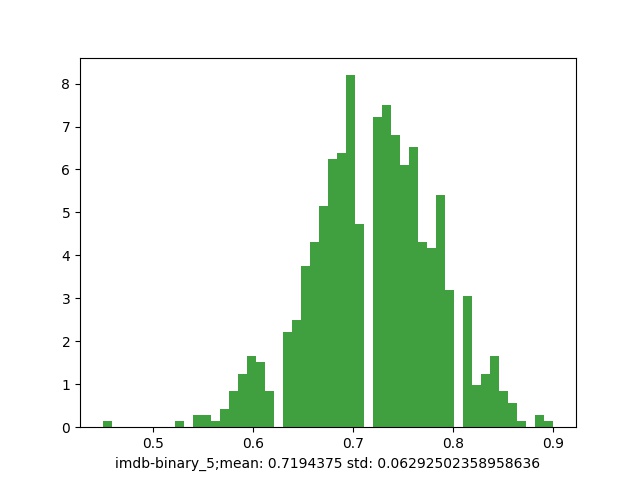}
    \caption{Performance on IMDB-Binary, 5 epochs}
    \label{fig:imdb5}
\end{figure}

\begin{figure}[h]
    \centering
    \includegraphics[width=6cm]{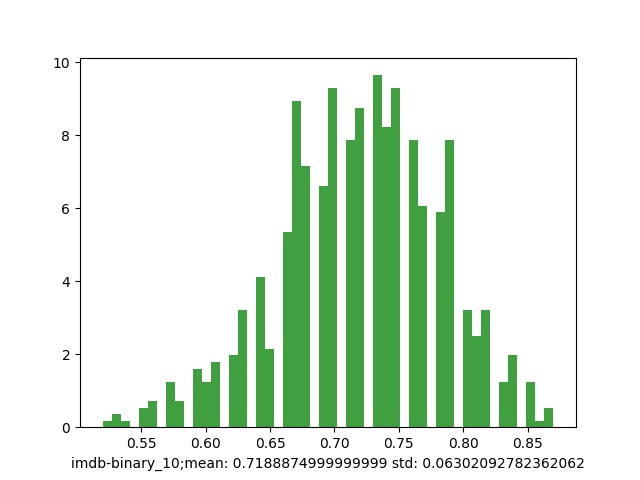}
    \caption{Performance on IMDB-Binary, 10 epochs}
    \label{fig:imdb10}
\end{figure}

\begin{figure}[h]
    \centering
    \includegraphics[width=6cm]{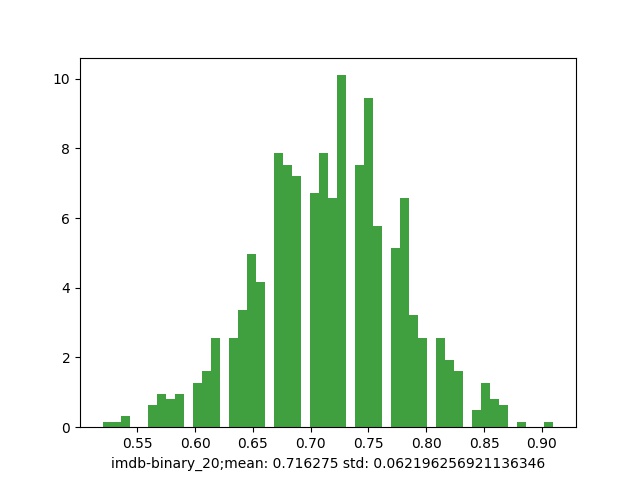}
    \caption{Performance on IMDB-Binary, 20 epochs}
    \label{fig:imdb20}
\end{figure}

\clearpage

\section{Visualizations of need for alignment}

\begin{figure}
    \centering
    \includegraphics[width=\textwidth]{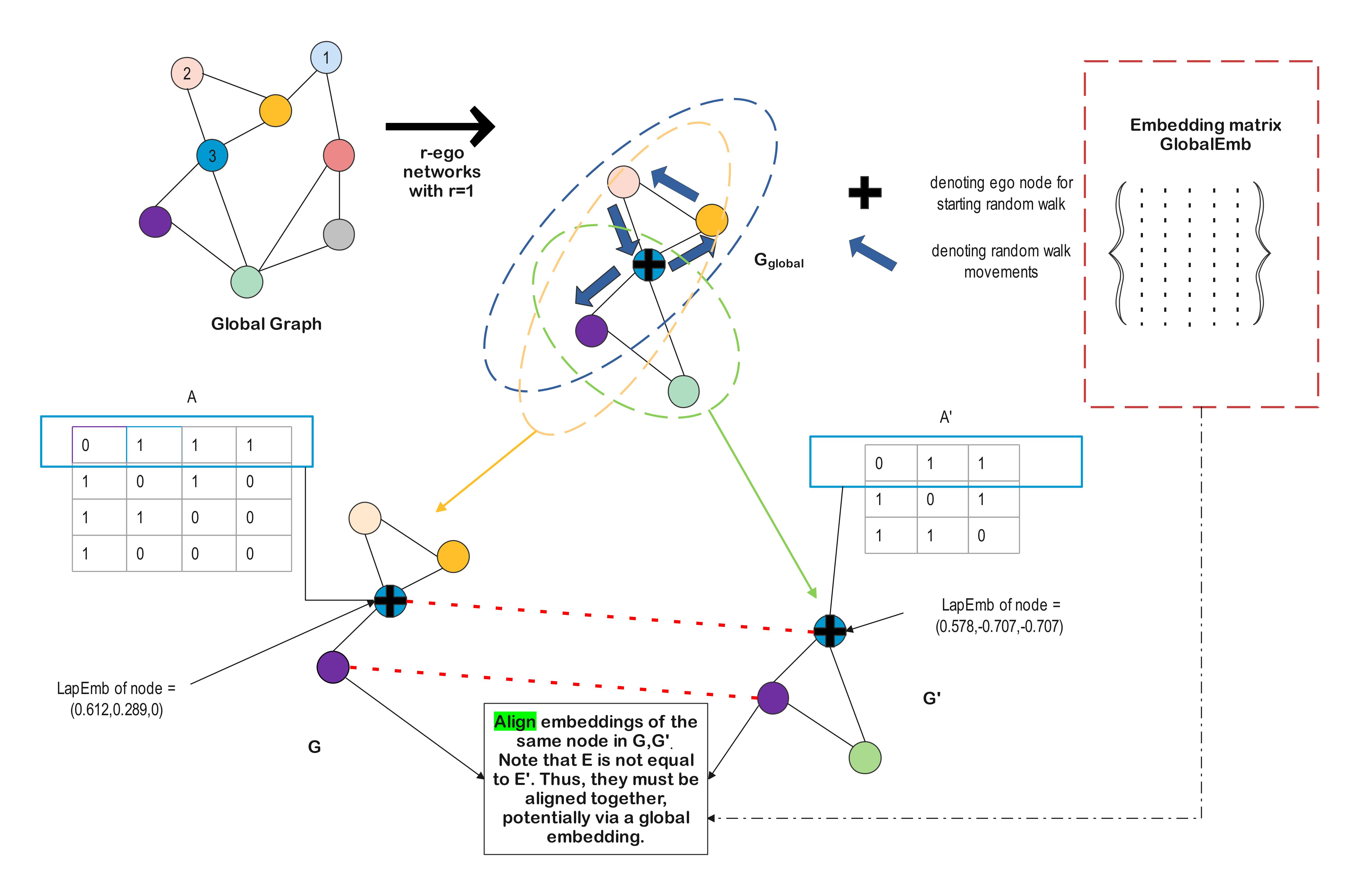}
    \caption{Alignment process overall}
    \label{fig:alignsketch}
\end{figure}

\begin{figure}
    \centering
    \includegraphics[width=\textwidth]{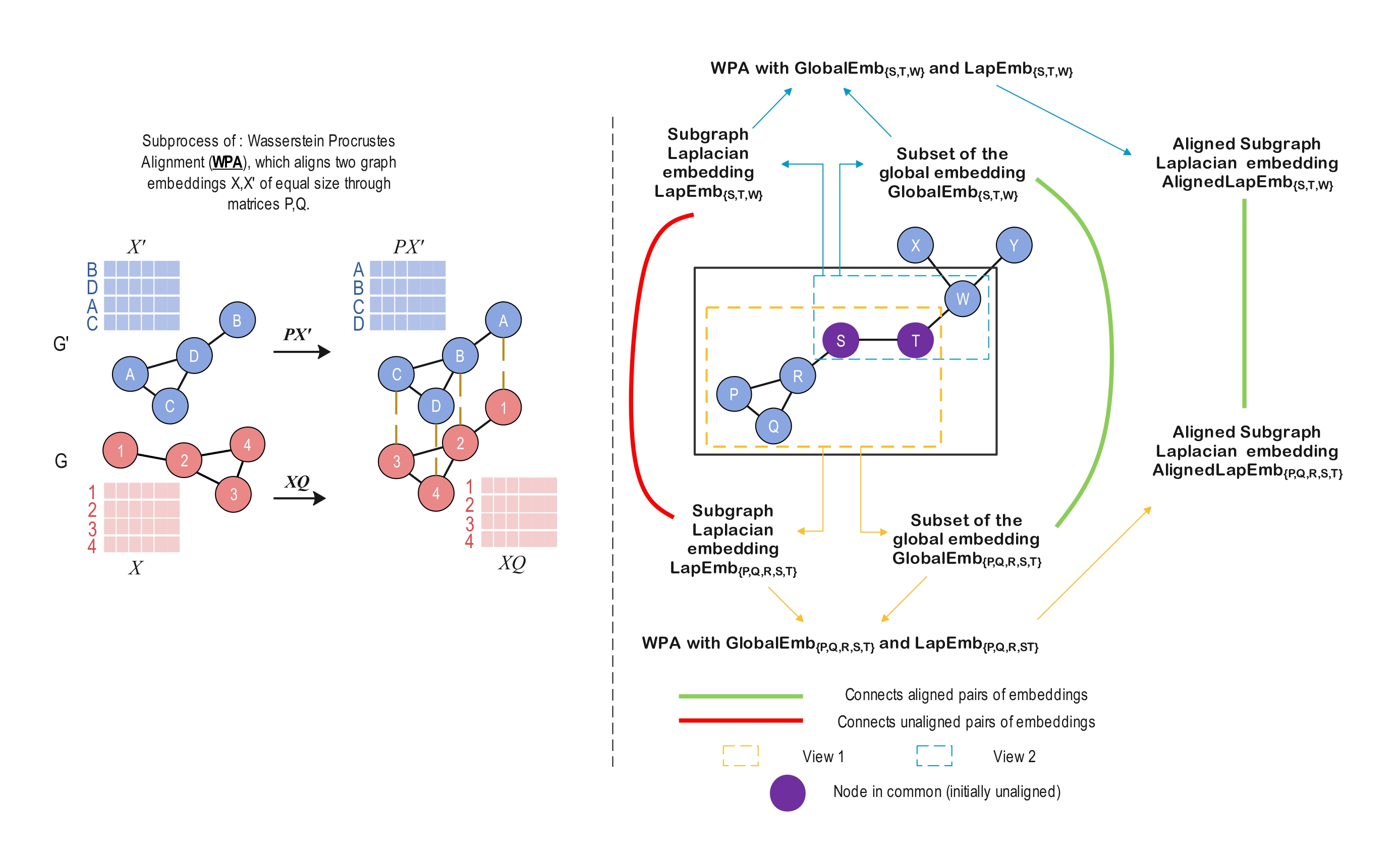}
    \caption{Alignment process detailed, with subprocess}
    \label{fig:aligndetail}
\end{figure}

Here, we provide two illustrative figures \ref{fig:alignsketch},~\ref{fig:aligndetail} we make that respectively demonstrate:

\begin{itemize}
\item The case where the global graph, after a random walk, can yield two views, which after Laplacian eigendecomposition end up with inconsistent embeddings for the same node, and thus requires alignment.
\item The Wasserstein-Procrustes alignment process which is used as a subprocess to correct the inconsistent embeddings. Representative papers that explain the Wasserstein Procrustes method include CONE-ALIGN~\citep{chen2020conealign}- especially in figures 1 and 2 and section 4.2 of the main text of the CONE-ALIGN paper, as well as REGAL~\citep{heimann2018regal} and G-CREWE~\citep{qin2020g}.
\end{itemize}

\clearpage

\end{document}